\documentclass[10pt,twocolumn,letterpaper]{article}
\pdfoutput=1

\usepackage{style/cvpr}
\usepackage{times}
\usepackage{epsfig}
\usepackage{graphicx}
\usepackage{amsmath}
\usepackage{amssymb}
\usepackage{booktabs}  %mine

\usepackage[breaklinks=true,bookmarks=false]{hyperref}

\cvprfinalcopy % *** Uncomment this line for the final submission

\begin{document}

%%%%%%%%% TITLE
\title{Improving Multi-Person Pose Estimation using Label Correction}

\author{\hspace{1cm}
Naoki Kato${}^\text{{\rm 1,2}}$\\
\and
Tianqi Li${}^\text{{\rm 2}}$\\
\and
Kohei Nishino${}^\text{{\rm 2}}$\\
\and
Yusuke Uchida${}^\text{{\rm 2}}$
\hspace{1cm}\\
\and
${}^\text{1}$Keio University
\hspace{1cm}
${}^\text{2}$DeNA Co., Ltd.
}

\maketitle

\setlength{\leftmargini}{10pt}

% -------------------------------------------------------------

\begin{abstract}
Significant attention is being paid to multi-person pose estimation methods recently, as there has been rapid progress in the field owing to convolutional neural networks. Especially, recent method which exploits part confidence maps and Part Affinity Fields (PAFs) has achieved accurate real-time prediction of multi-person keypoints. However, human annotated labels are sometimes inappropriate for learning models. For example, if there is a limb that extends outside an image, a keypoint for the limb may not have annotations because it exists outside of the image, and thus the labels for the limb can not be generated. If a model is trained with data including such missing labels, the output of the model for the location, even though it is correct, is penalized as a false positive, which is likely to cause negative effects on the performance of the model. In this paper, we point out the existence of some patterns of inappropriate labels, and propose a novel method for correcting such labels with a teacher model trained on such incomplete data. Experiments on the COCO dataset show that training with the corrected labels improves the performance of the model and also speeds up training.
\end{abstract}

% -------------------------------------------------------------

\section{Introduction}
Multi-person pose estimation aims at detecting and localizing keypoints for all persons in an image, which is useful for many applications like human action recognition and in-vehicle video recognition. It is a challenging task since it requires accurate localization of the keypoints of an unknown number of persons in situations where there may be a variety of lighting, clothing, human poses and occlusion due to crowds.
In addition, algorithms are required to be both accurate and computationally efficient for practical applications.

\begin{figure}[t]
\begin{center}
{\setlength\tabcolsep{.4mm}
\begin{tabular}{ccc}
  \includegraphics[width=.32\hsize]{data/00001296_annotation} &
  \includegraphics[width=.32\hsize]{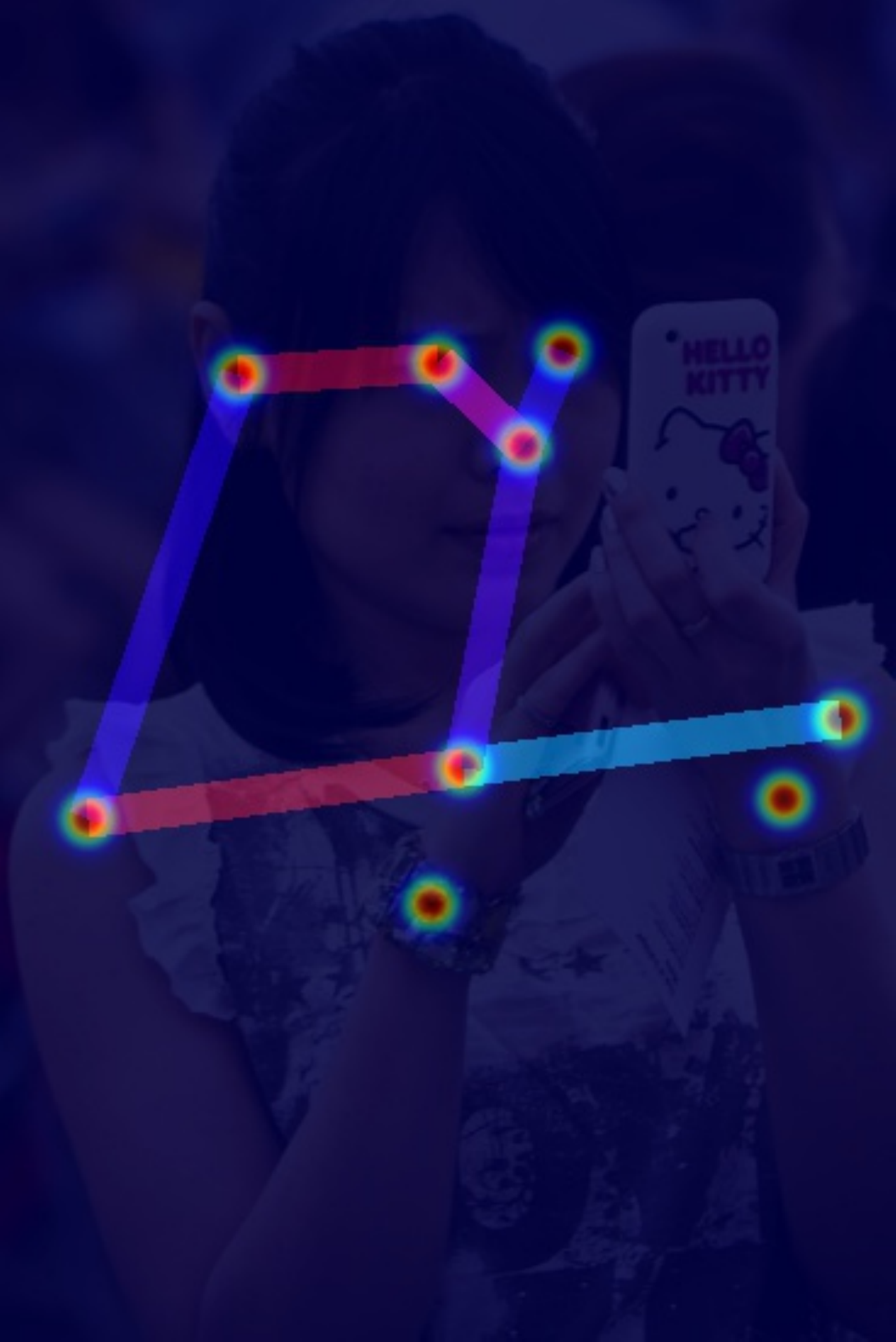} &
  \includegraphics[width=.32\hsize]{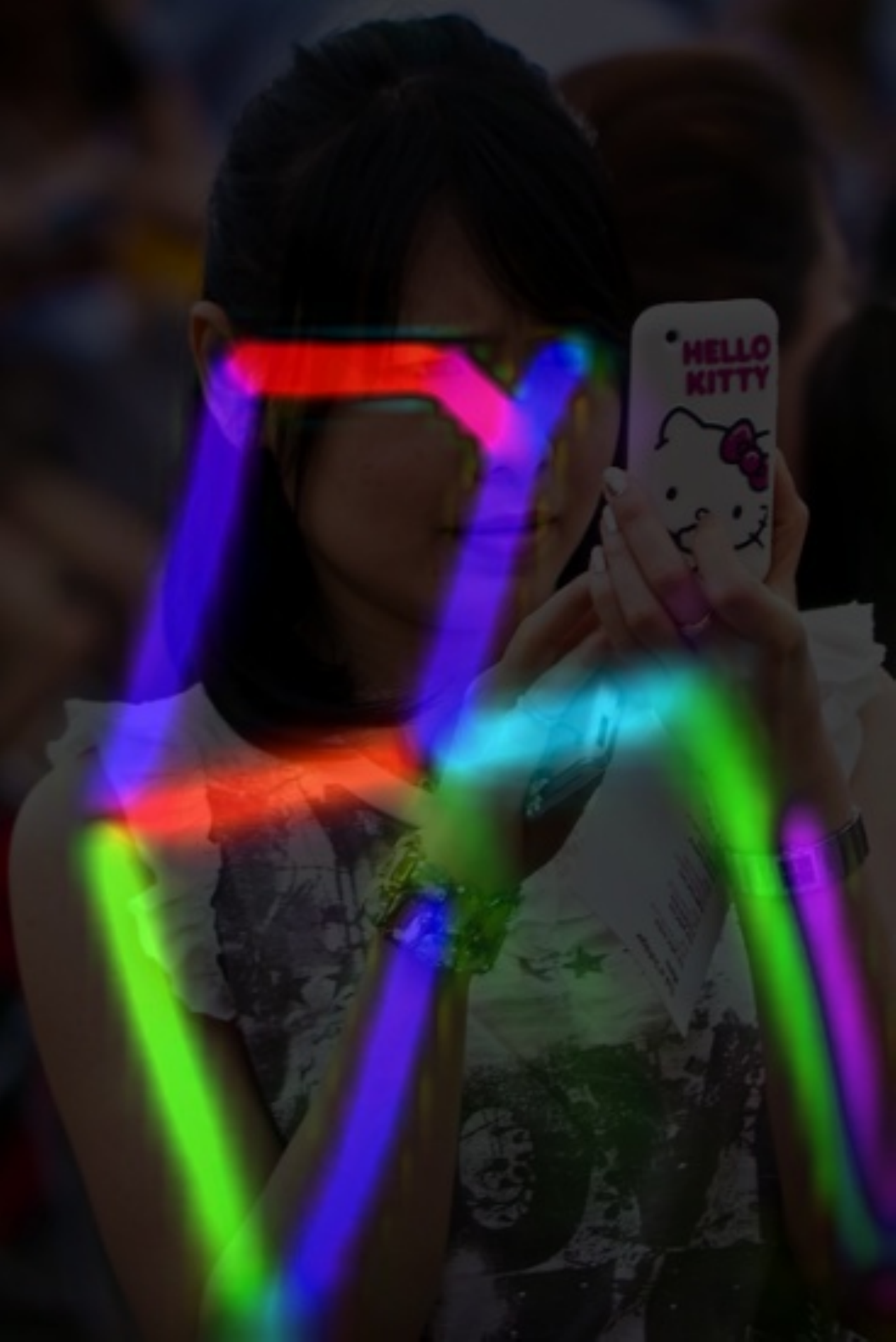} \\
  (a) & (b) & (c)
\end{tabular}}
\end{center}
\caption{(a) Keypoint annotations. (b) Confidence maps and PAFs generated from the annotations. Since the keypoints for the left and right elbows exist outside the image, they are not annotated. Therefore the PAFs for the upper arm and forearm can not be generated. (c) The estimated PAFs by a trained model. The model estimates the existence of the limbs properly.}
\label{fig:compare}
\end{figure}

Recently, along with the significant progress of convolutional neural networks (CNNs) \cite{lecun1998gradient,ussakovsky2015imagenet,simonyan2014very},
the performance of pose estimation algorithms has also greatly improved.
Cao et al. proposed a pose estimation algorithm using a CNN to estimate confidence maps of human keypoints and Part Affinity Fields (PAFs) representing the existence of
'limbs' (in this paper, meaning keypoint pairs) \cite{cao2017realtime}.
Final pose estimation results can be obtained by parsing process that picks keypoint candidates from confidence maps and associate them using PAFs. The process requires low computational cost even if the number of people in an image increases.
While conventional approaches associate keypoints based mainly on their geometric relationships such as distance and angle \cite{pishchulin2016deepcut,insafutdinov2016deepercut}, Cao's approach defines PAFs and propose a way to learn them directly with a CNN, which enables the computation of the confidence for keypoint association, considering the context in the image.

While many algorithms for accurate multi-person pose estimation have been proposed, there is few researches to explore training datasets for this task. Human annotated labels are sometimes inappropriate for training models, which is a fundamental problem in many vision tasks. Although some works on learning with noisy labels exist \cite{reed2014training,jindal2016learning,tanaka2018joint}, they mainly focus on classification. To the best of our knowledge, learning with noisy labels for the pose estimation task has not been discussed. For instance, Cao's approach have issues regarding PAFs.
Since the PAF is defined as connections of keypoint pairs, if one side of the keypoint pair exists outside an image like Figure \ref{fig:compare}a, the PAF for the limb can not be generated (Figure \ref{fig:compare}b). Predicted results of limbs without labels are penalized as false positives, which becomes a factor that decreases the performance of the model.

In this paper, we point out the existence of inappropriate labels generated from human annotations mainly targeting Cao's pose estimation algorithm and propose a novel method to correct such incomplete labels.
A trained model can generate complemental labels which does not exist in human annotations like Figure \ref{fig:compare}c, because the model is generalized to the features of keypoints and limbs. Thus, we propose to correct incomplete labels with the teacher model which can predict complemental labels.
Specifically, we correct such labels by performing a element-wise max operation on the label and the output of the teacher model to supplement the missing part in the label. Since the proposed method uses the soft output of the teacher model to correct labels, it also speeds up training, like what is observed in training with knowledge distillation \cite{hinton2015distilling}. In our experiment using the COCO dataset \cite{lin2014microsoft}, we show that
the performance of the model trained with the corrected labels improves by our method.

Our contribution can be summarized as follows:
\begin{itemize}
  \item We point out multiple types of inappropriate labels generated from human annotations using Cao's pose estimation algorithm \cite{cao2017realtime} as an example.
  \item We propose a novel method for correcting such incomplete labels using the output of the teacher model. To the best of our knowledge, this is the first attempt to use a trained model to correct noisy labels in a pose estimation task.
  \item In experiments, we compare the performance of the models trained with knowledge distillation \cite{hinton2015distilling} and proposed label correction to show the superiority of our approach.
\end{itemize}

% -------------------------------------------------------------

\begin{figure*}[t]
\begin{center}
{\setlength\tabcolsep{.6mm}
\begin{tabular}{cccc}
  \includegraphics[height=31.4mm]{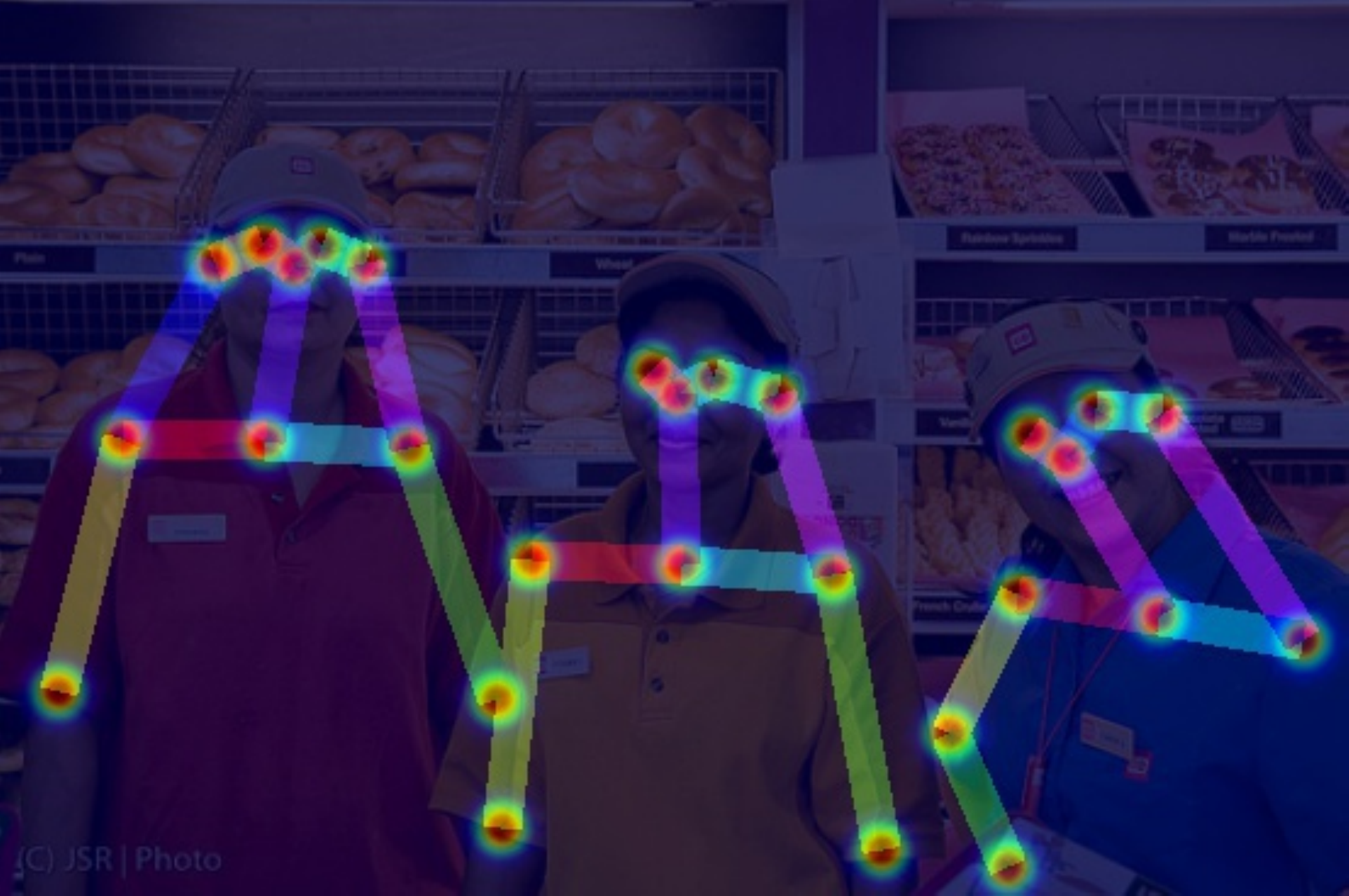} &
  \includegraphics[height=31.4mm]{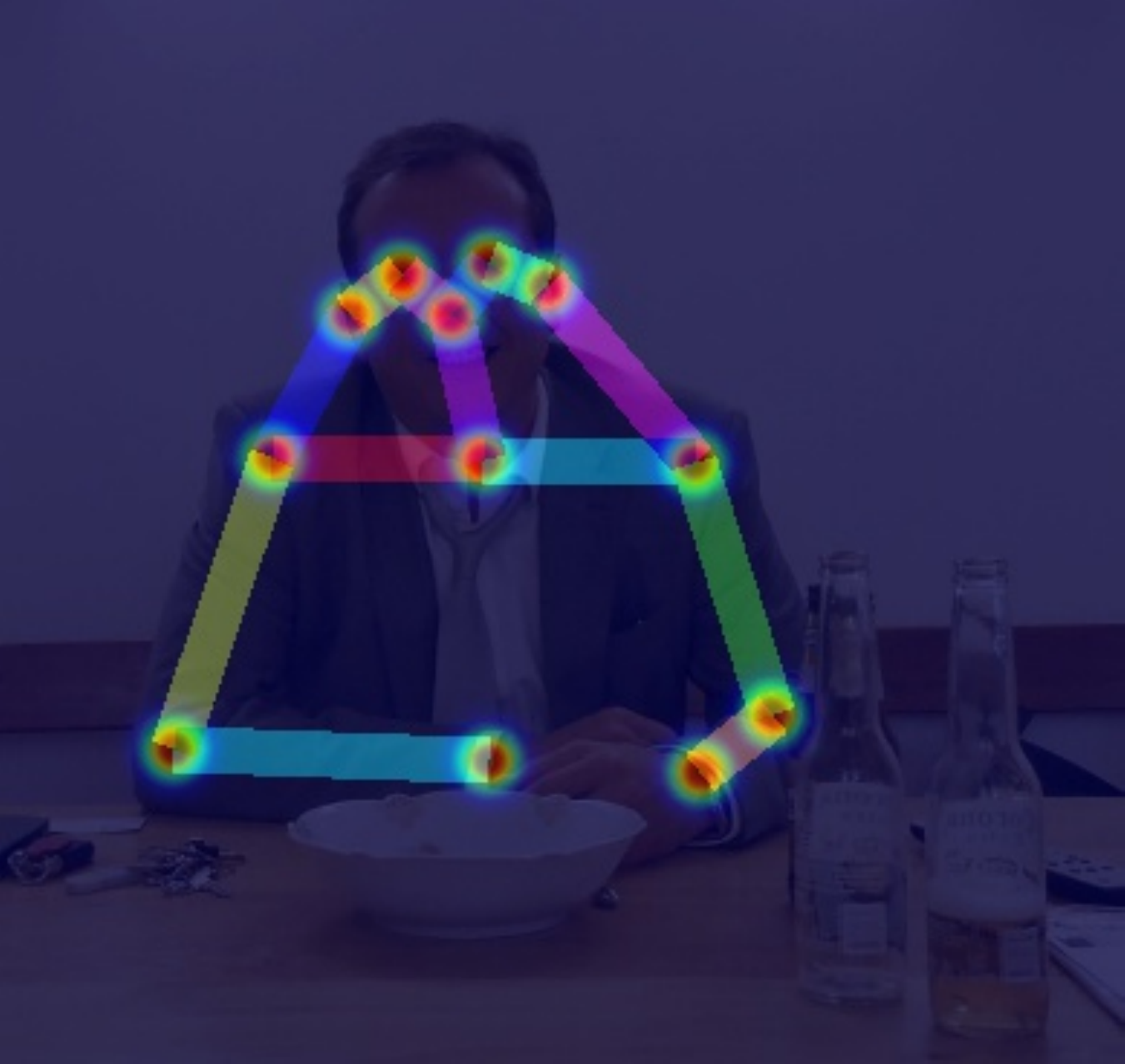} &
  \includegraphics[height=31.4mm]{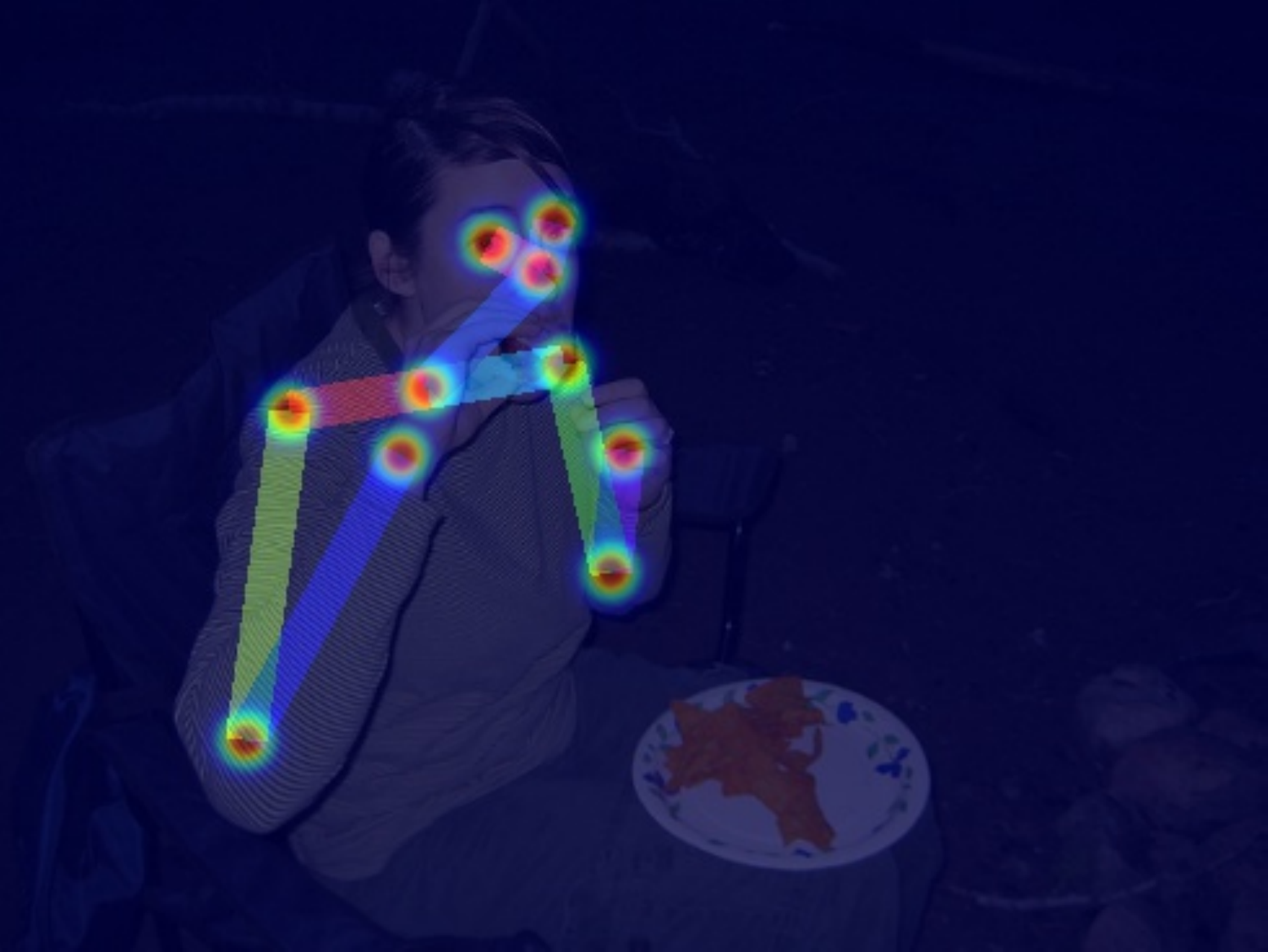} &
  \includegraphics[height=31.4mm]{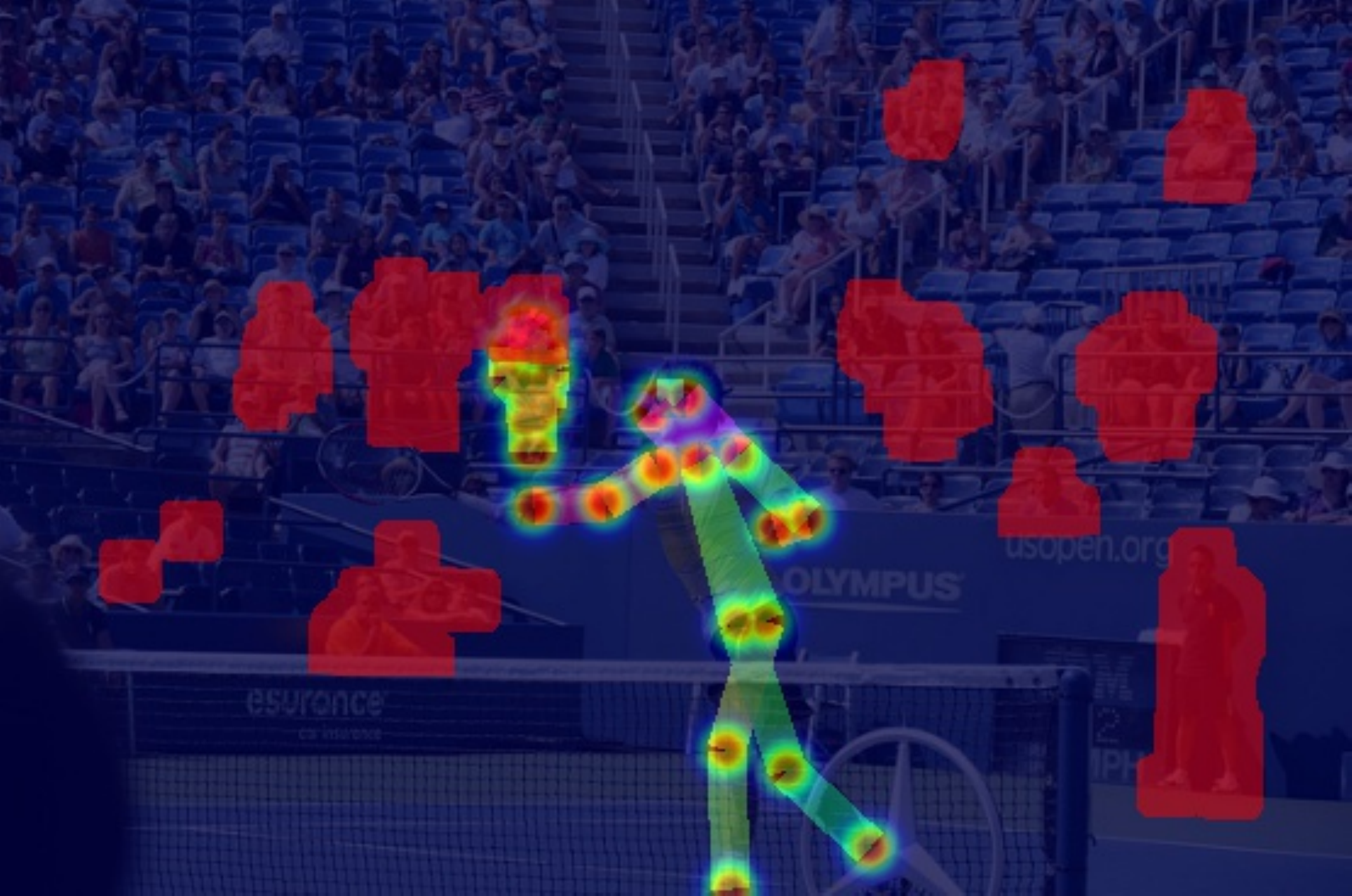} \\
  (a) Keypoint protrusions & (b) Keypoint occlusion & (c) Missed keypoints & (d) Missed masks
\end{tabular}}
\end{center}
\caption{Labels generated from keypoint annotations of the COCO dataset. (a) Due to the lack of annotations for keypoints connecting limbs that extend outside the image, PAFs for the limbs can not be generated. (b) Because of missing keypoint annotations due to occlusion, PAFs for limbs can not be generated even though they exist in the image. (c) There are missed keypoint annotations. (d) The red regions are masks indicating there is no annotation, but there are persons not given keypoint annotations and not covered by the masks either. These data are likely to cause negative effects on training a model.}
\label{fig:bad_label}
\end{figure*}

% -------------------------------------------------------------

\section{Related Work}
\subsection{Multi-Person Pose Estimation}
Human pose estimation is the task of estimating the keypoint coordinates of persons in an image, which is very challenging due to the wide variation in human appearance present in images \cite{modec13,Johnson10}. Due to its high applicability, the research topic has attracted significant attention in recent years.
Pose estimation is categorized into single person pose estimation and multi-person pose estimation. While single person pose estimation makes the assumption that there is a single person in an image and locates his/her keypoints, multi-person pose estimation requires the detection and localization of all the keypoints of an unspecified number of people individually.
The latter becomes highly challenging especially in crowded scenes.
Both single-person and multi-person pose estimation algorithms depend on CNNs for their performance.
In this paper, we focus on multi-person pose estimation, which is closer to real world situations.
Multi-person pose estimation methods are roughly divided into top-down approaches and bottom-up approaches, and the difference between them will be described below.

% \subsubsection{Top-down approaches.}
\noindent{\bf Top-down approaches.}
Top-down approaches firstly detect bounding boxes of people in an image and locate all the keypoints of the persons afterwards.
% Major errors rarely occur if the bounding box estimation is accurate [].
These approaches need to estimate each person's keypoints separately, so
% there is the disadvantage that
the computational cost increases in proportion to the number of people in the image.
Many algorithms of this type first predict bounding boxes with a human detector and then apply a single person pose estimator for each person, so
the approaches rely heavily on human detector for their performances \cite{cao2017realtime}.
Many single person pose estimation algorithms based on CNNs predict heatmaps representing the existence of persons' keypoints.
The Convolutional Pose Machine \cite{wei2016convolutional} enforces intermediate supervision of the outputs of each stage of the network to address the problem of vanishing gradients. The stacked hourglass network \cite{newell2016stacked} is based on the successive steps of pooling and upsampling that enlarge receptive field.
The Cascaded Pyramid Network \cite{chen2017cascaded} adopts an architecture like the Feature Pyramid Network \cite{lin2017feature} to produce feature maps of different resolutions which are integrated together to achieve multi-scale detection.
Papandreou et al. predicts both heatmaps and offsets for the ground-truth keypiont location, and combine them to obtain the final keypoint predictions \cite{papandreou2017towards}.
% Mask R-CNN \cite{he2017mask} extends Faster R-CNN \cite{ren2015faster} to learn bounding box regression and human keypoint estimation in an end-to-end fashion.
Mask R-CNN \cite{he2017mask} extends Faster R-CNN \cite{ren2015faster} to multi-task detection -  bounding box regression and human keypoint estimation - in an end-to-end fashion.

% \subsubsection{Bottom-up approaches.}
\noindent{\bf Bottom-up approaches.}
Bottom-up approaches firstly detect all the keypoints of multiple persons in an image without distinguishing the individuals, and assemble the keypoints of each person to get final pose estimation results.
DeepCut \cite{pishchulin2016deepcut} interprets the problem of distinguishing different persons as an Integer Linear Program (ILP) and partitions part detection candidates into person clusters. Then the final pose estimation results are obtained when person clusters are combined with labeled body parts.
DeeperCut \cite{insafutdinov2016deepercut} improves DeepCut using ResNet \cite{he2016deep} and employs image-conditioned pairwise terms.
% to get better performance.
The above two methods need to solve ILP for partitioning part candidates, but since it has a high computational cost, these methods are not suitable for realtime applications.
Newell et al. simultaneously produces score maps and pixel-wise embedding to group the candidate keypoints into different people to get final pose estimation results \cite{newell2017associative}.
The method proposed by Cao et al. has a better trade-off between performance and speed \cite{cao2017realtime}, since the increase in computational cost is small even if the number of people in an image increases.
Due to its high applicability, we use this algorithm as our basic pose estimation approach. We will briefly describe this algorithm in the next section.

\subsection{Pose Estimation with Part Affinity Fields}
This pose estimation method takes an image of size $w \times h$ as an input into a CNN to predict part confidence maps $H$ of body part locations and Part Affinity Fields (PAFs) $L$ which represent the existence of limbs. Confidence maps $H=(H_1...H_J)$ have $J$ confidence maps, one per part, where $H_j\in \mathbb{R}^{w\times h}, j\in\{1...J\}$. The ground-truth label for each part $H_j^{GT}$ is generated by the gaussian distributions whose mean positions are annotated keypoint locations for all the persons in an image regardless of human identity. PAFs $L=(L_1...L_c)$ have $C$ vector fields, one per limb, where $L_c\in \mathbb{R}^{w\times h\times2}, c\in\{1...C\}$.
The ground-truth label for each limb $L_c^{GT}$ is generated by a unit vector that points from $j_1$ to $j_2$ in a rectangular area defined between the corresponding keypoint pair $j_1,j_2$ and the zero vector for all other points.
Given model predictions $P=(H,L)$ and ground-truth labels $y_{GT}=(H^{GT},L^{GT})$, the model is trained using mean squared error $E_{L2}(P,y_{GT})$ defined as follows:
\begin{equation}
\label{eq:normal_loss}
\scalebox{0.96}{$\displaystyle
\begin{split}
  E_{L2}(P,y_{GT})=& \sum_{j=1}^{J}\sum_{p}W(p)\|H_j(p)-H_j^{GT}(p)\|_2^2 \\
                   &+\sum_{c=1}^{C}\sum_{p}W(p)\|L_c(p)-L_c^{GT}(p)\|_2^2,
\end{split}
$}
\end{equation}
where $p$ is 2D coordinates in an image, $W$ is a binary mask with $W(p)=0$ for regions that should be ignored, such as crowds.

At the parsing step, a set of bipartite matchings to associate body part candidates is performed.
First, we detect keypoint candidates from maximum position of the confidence maps. Then, we get limb confidence scores by computing line integral between all the corresponding keypoint pairs. Final pose estimation results are obtained by associating keypoint pairs greedily using these scores.

% -------------------------------------------------------------

\subsection{Knowledge Distillation}
Hinton et al. proposed a model compression method called knowledge distillation \cite{hinton2015distilling}. They use the final output of a trained teacher network as a soft target for training a small student network to transfer the information acquired by the teacher. Since soft targets contain information obtained by the teacher network that is not involved in hard targets generated from human annotations, they function as strong regularizers for training. Romero et al. and Yim et al. extend knowledge distillation by using intermediate outputs of a teacher network for training signals to improve the performance of the student network \cite{romero2014fitnets,yim2017gift}.
Gao et al. distilled knowledge obtained by multiple teacher networks into a single student network to enable it to classify 100K object categories \cite{gao2017knowledge}. % Todo: 見てもらう
Although knowledge distillation is first proposed as a model compression method that aims at improving the performance of small student network, Furlanello et al. showed that student networks that are parameterized identically to the teacher networks outperform them \cite{furlanello2017born}.
For the tasks of object detection and human pose estimation, Radosavovic et al. proposed data distillation that ensembles predictions of a single model from multiple transformations of unlabeled data to increase training data \cite{radosavovic2017data}. Data distillation differs from other distillation methods in terms of aiming at generating annotations for unlabeled data,
which can be used for training as a hard target.

In this paper, as opposed to data distillation \cite{radosavovic2017data}, we aim to improve existing labels in training data for a pose estimation task, and propose an approach to distill knowledge obtained by a teacher network to correct labels. We train a student network that has identical architecture to the teacher network like Furlanello's approach \cite{furlanello2017born} and use corrected labels to train it to improve the performance of the model.

% -------------------------------------------------------------

\section{Proposed Method}
In this section, we first indicate the existence of inappropriate labels in the pose estimation method proposed by Cao et al. \cite{cao2017realtime}. Then, we describe the detailed method of our approach to improve those labels.

\subsection{Problems with Labels}
Datasets used for pose estimation such as the COCO dataset \cite{lin2014microsoft} and the MPII dataset \cite{andriluka20142d} have annotations of keypoint locations for each person in the images. The multi-person pose estimation method proposed by Cao et al. mentioned in the previous section generates labels based on these annotations. In this section, we indicate that there are inappropriate labels for training in their approach.

Figure \ref{fig:bad_label} shows some examples of inappropriate labels generated using the COCO dataset. PAFs are generated from the pairs of keypoint annotations. The keypoints existing outside of images (Figure 2a) and having severe occlusion (Figure 2b) don't have annotations. In that case, labels for corresponding limbs cannot be generated, even though they are inside of the images. The datasets have images with missing annotations like Figure 2c as well. Furthermore, the datasets have masks for ignoring regions where it is hard to annotate keypoints such as crowded areas.
We train a model ignoring the masked regions (Equation \ref{eq:normal_loss}),
but these masks are sometimes missing. The prediction of a model for regions that do not have appropriate labels are penalized as false positives, even if the keypoints and limbs are correctly predicted by the model. This lowers the performance of the trained model.

To summarize, the failure patterns of the labels are summarized as follows:
\begin{itemize}
  \item There are no annotations for keypoints outside images.
  \item There are no annotations for keypoints with severe occlusion.
  \item Annotations for visible keypoints are missing.
  \item Mask annotations for ignoring region are missing.
\end{itemize}

% -------------------------------------------------------------

\begin{figure}[t]
\begin{center}
{\setlength\tabcolsep{.3mm}
\begin{tabular}{ccc}
  \includegraphics[width=.326\hsize]{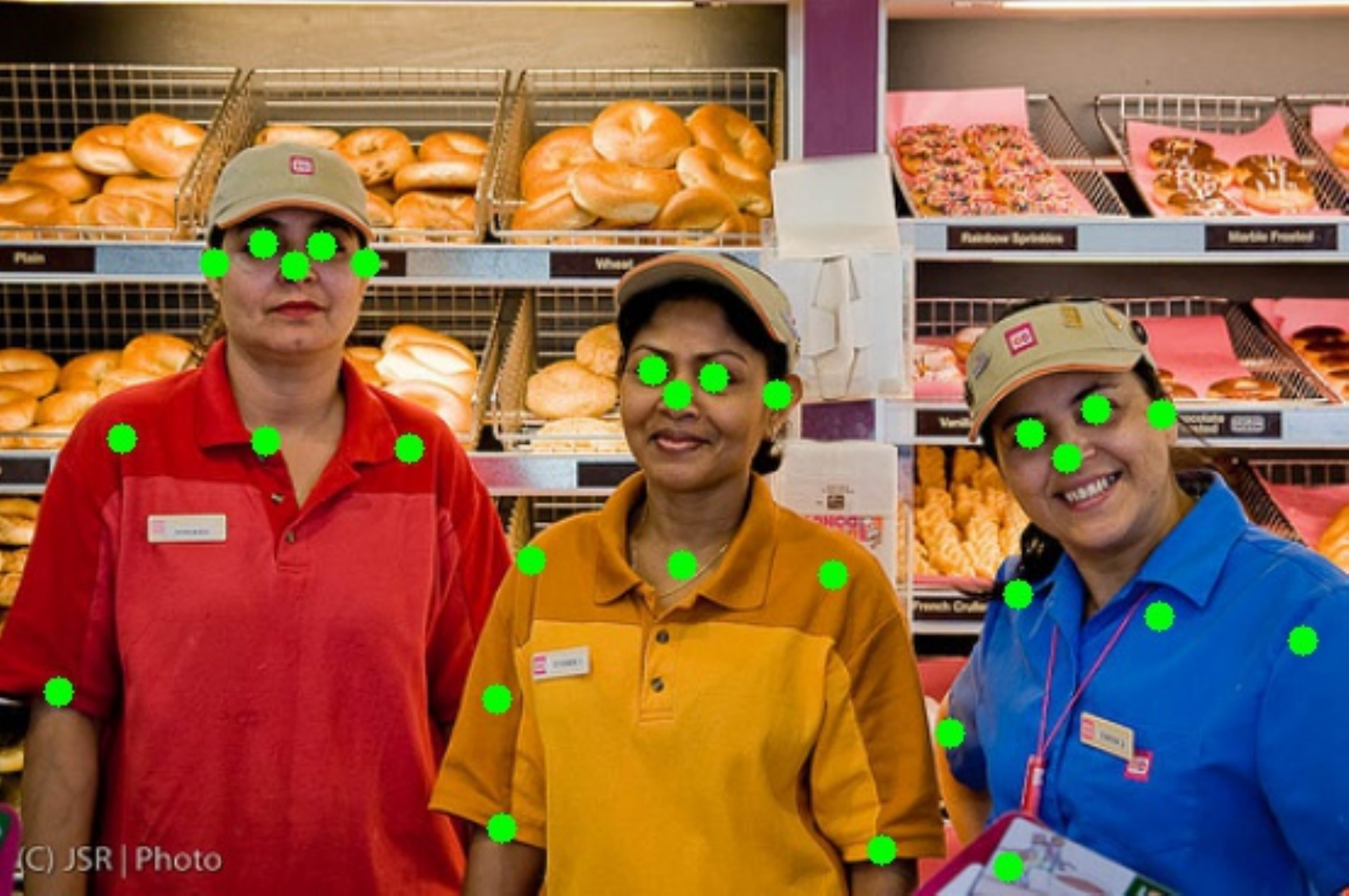} &
  \includegraphics[width=.326\hsize]{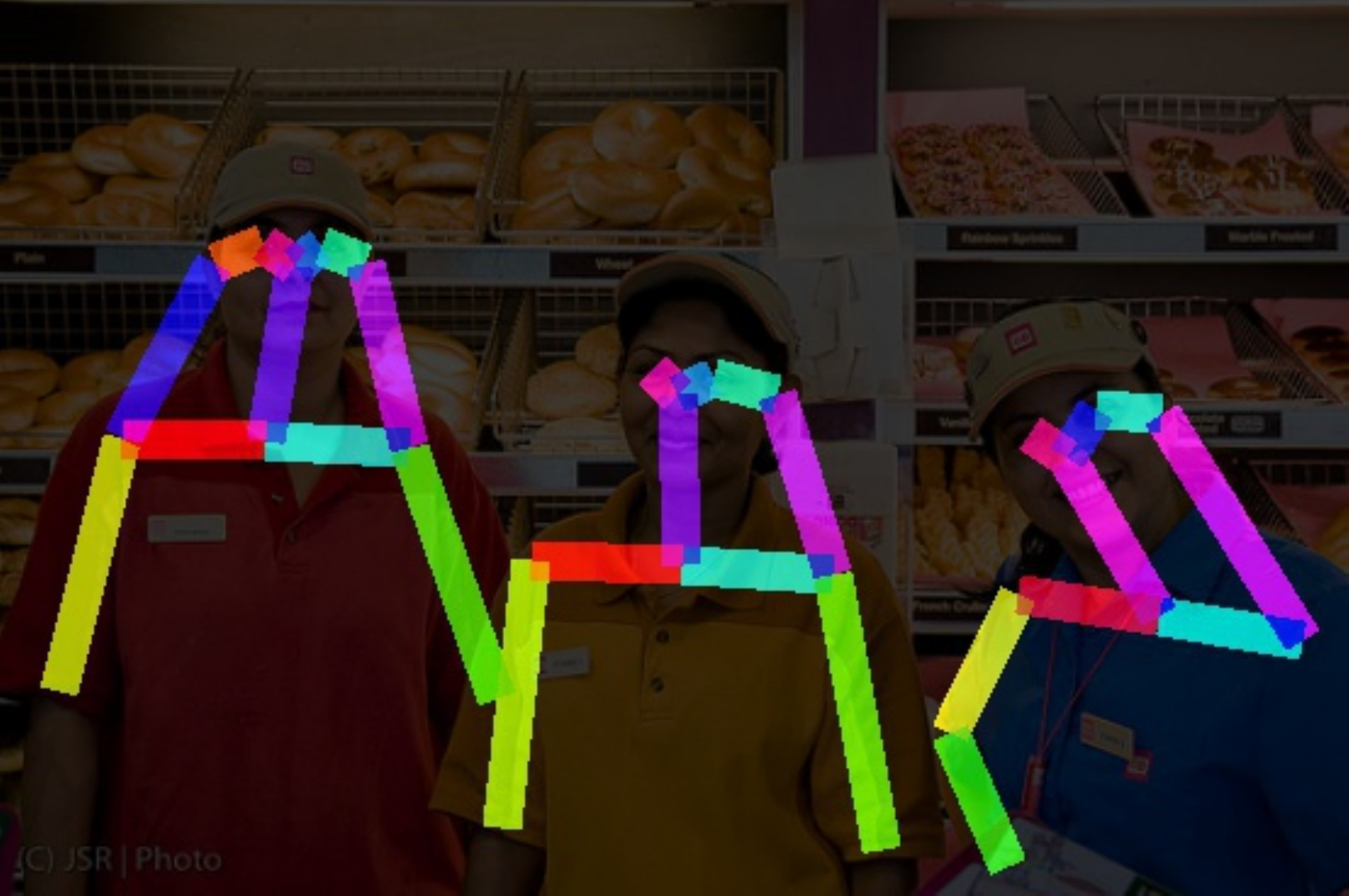} &
  \includegraphics[width=.326\hsize]{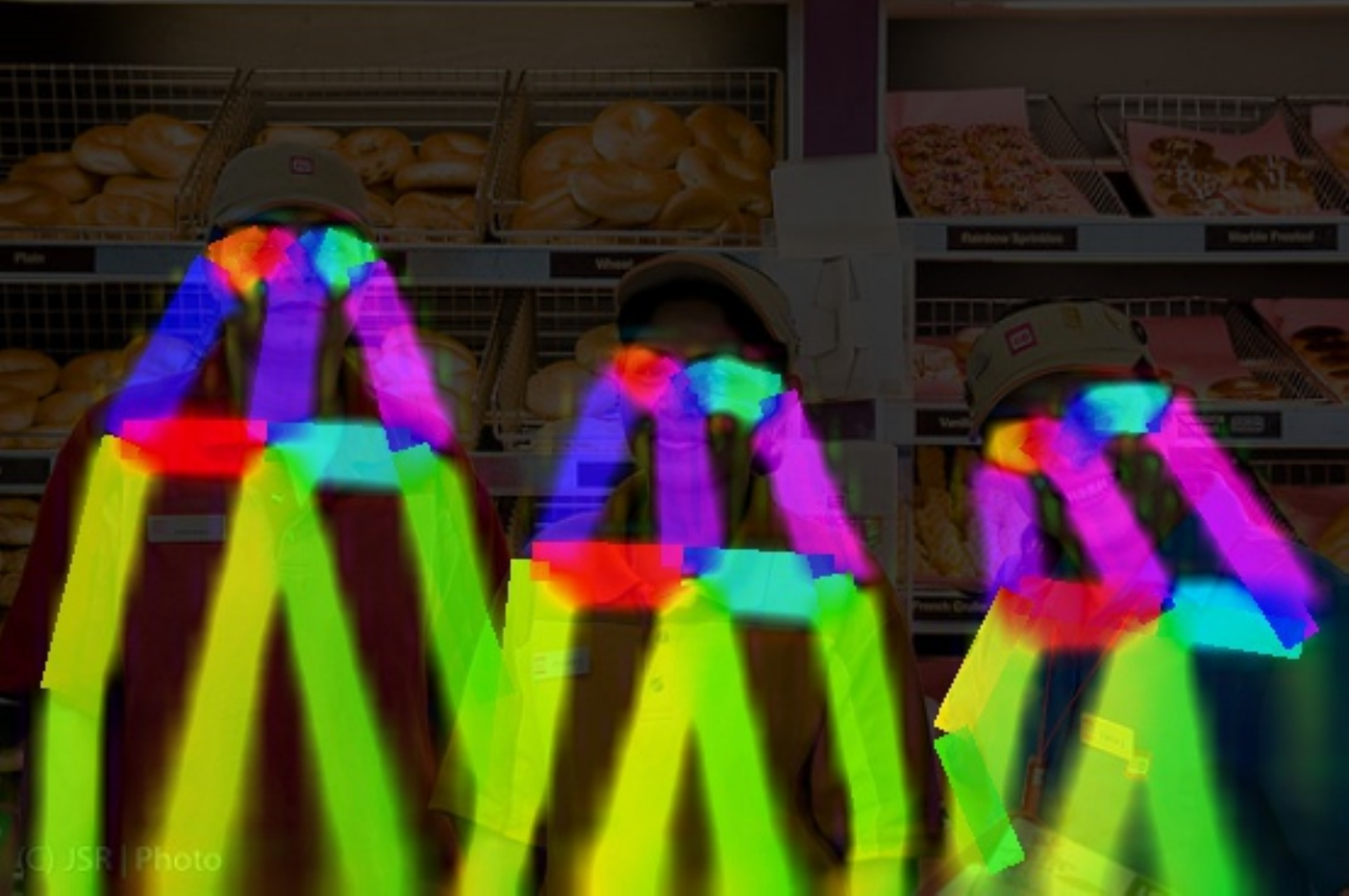} \\
  \includegraphics[width=.326\hsize]{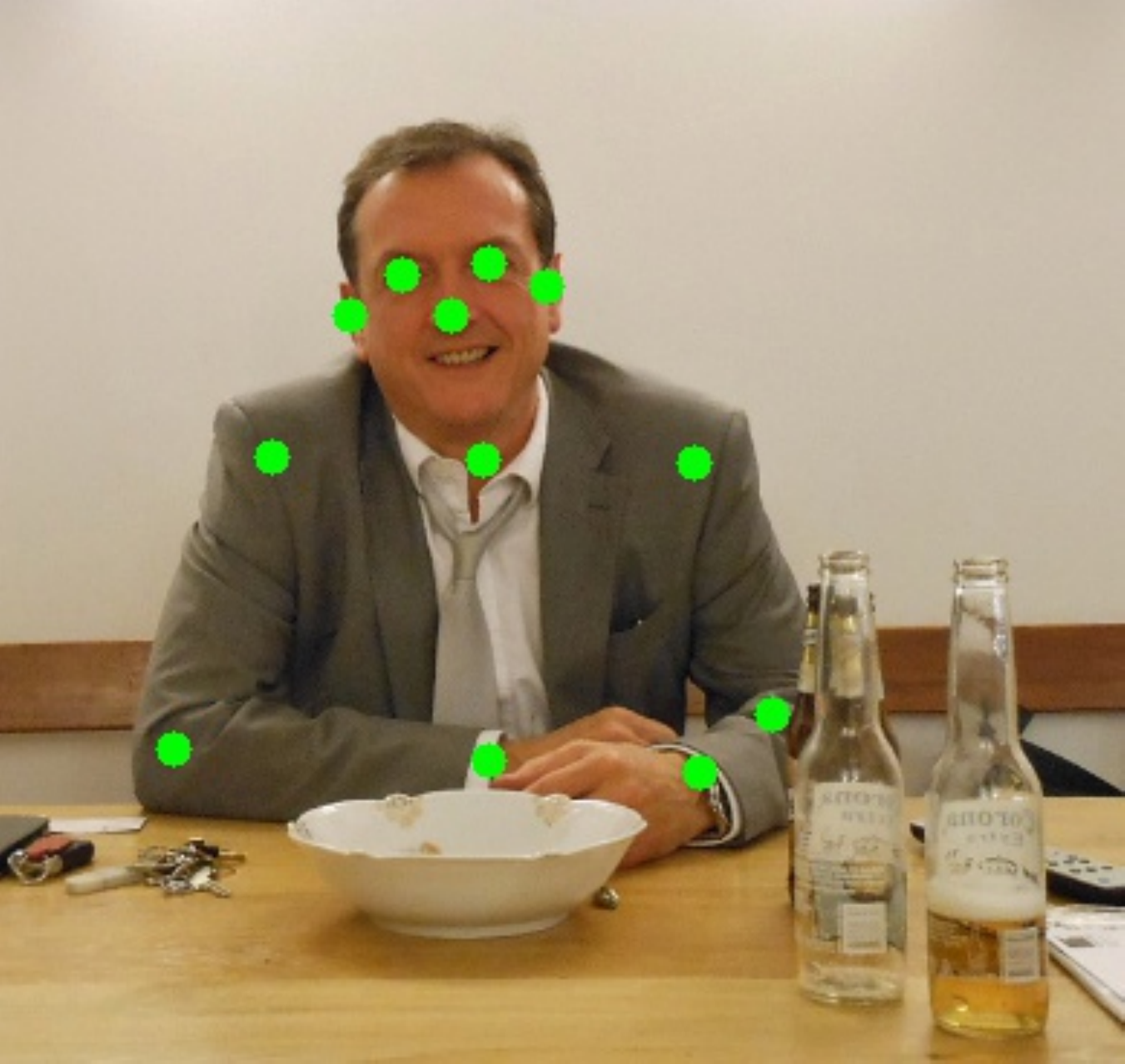} &
  \includegraphics[width=.326\hsize]{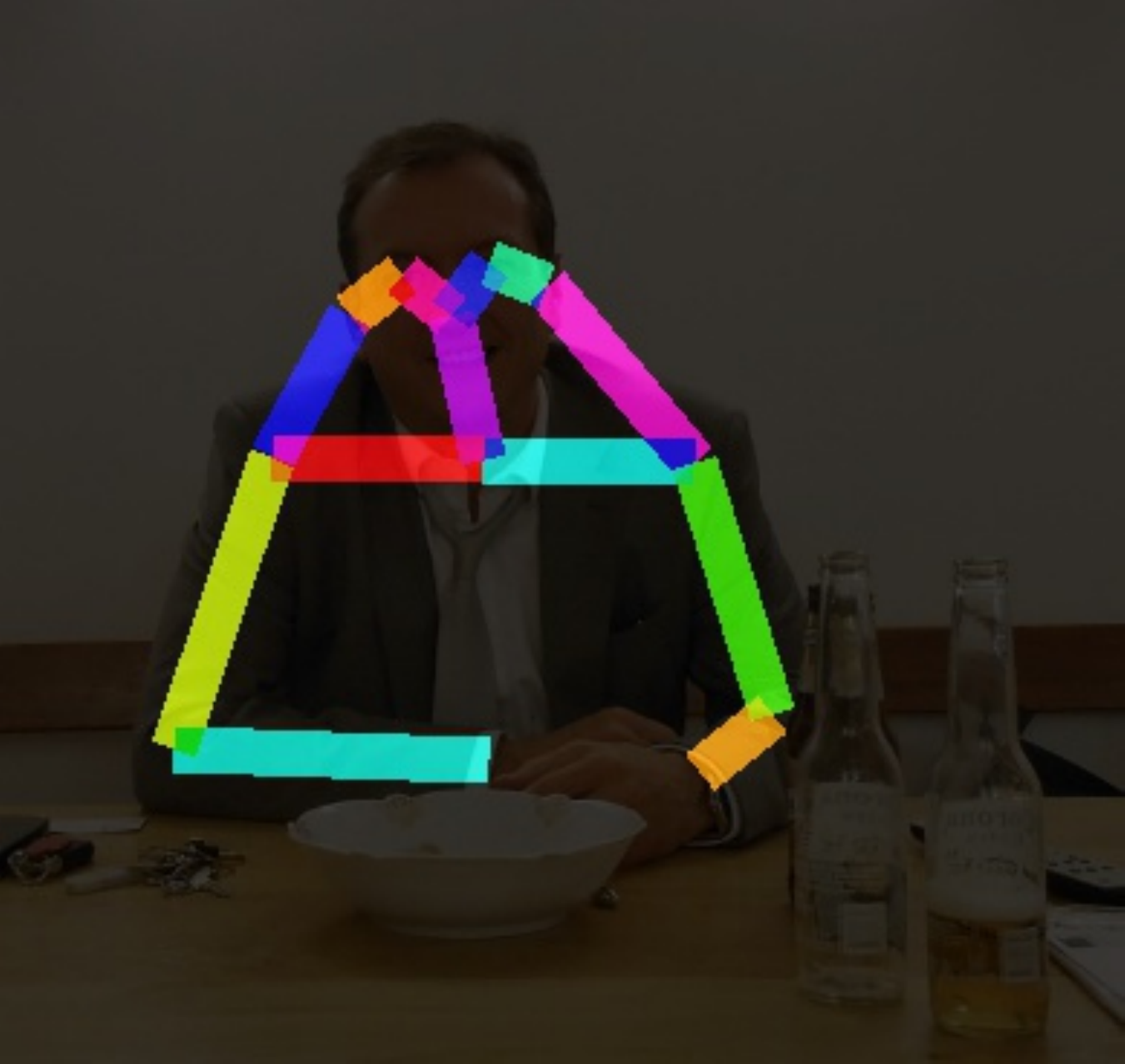} &
  \includegraphics[width=.326\hsize]{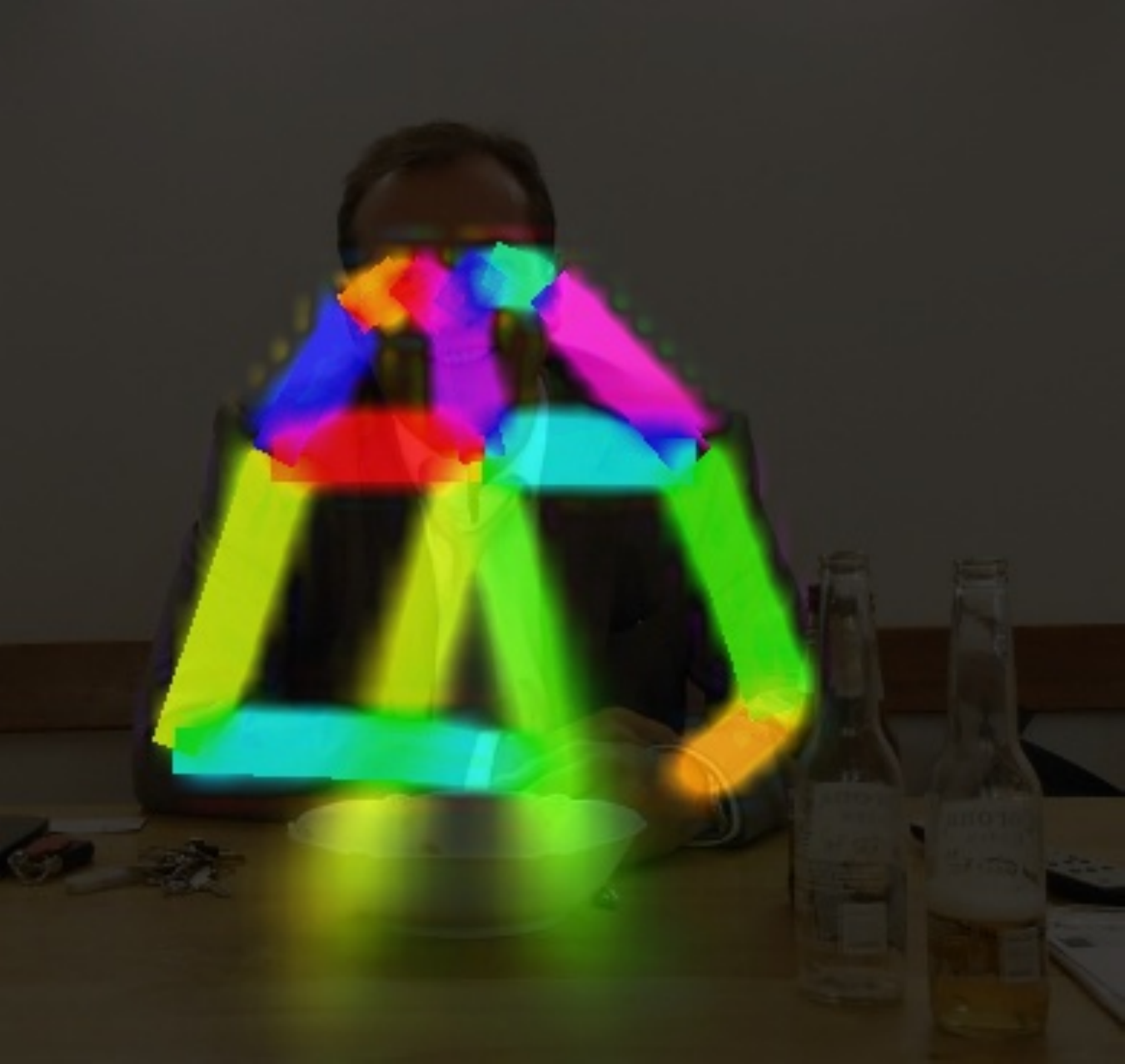} \\
  \includegraphics[width=.326\hsize]{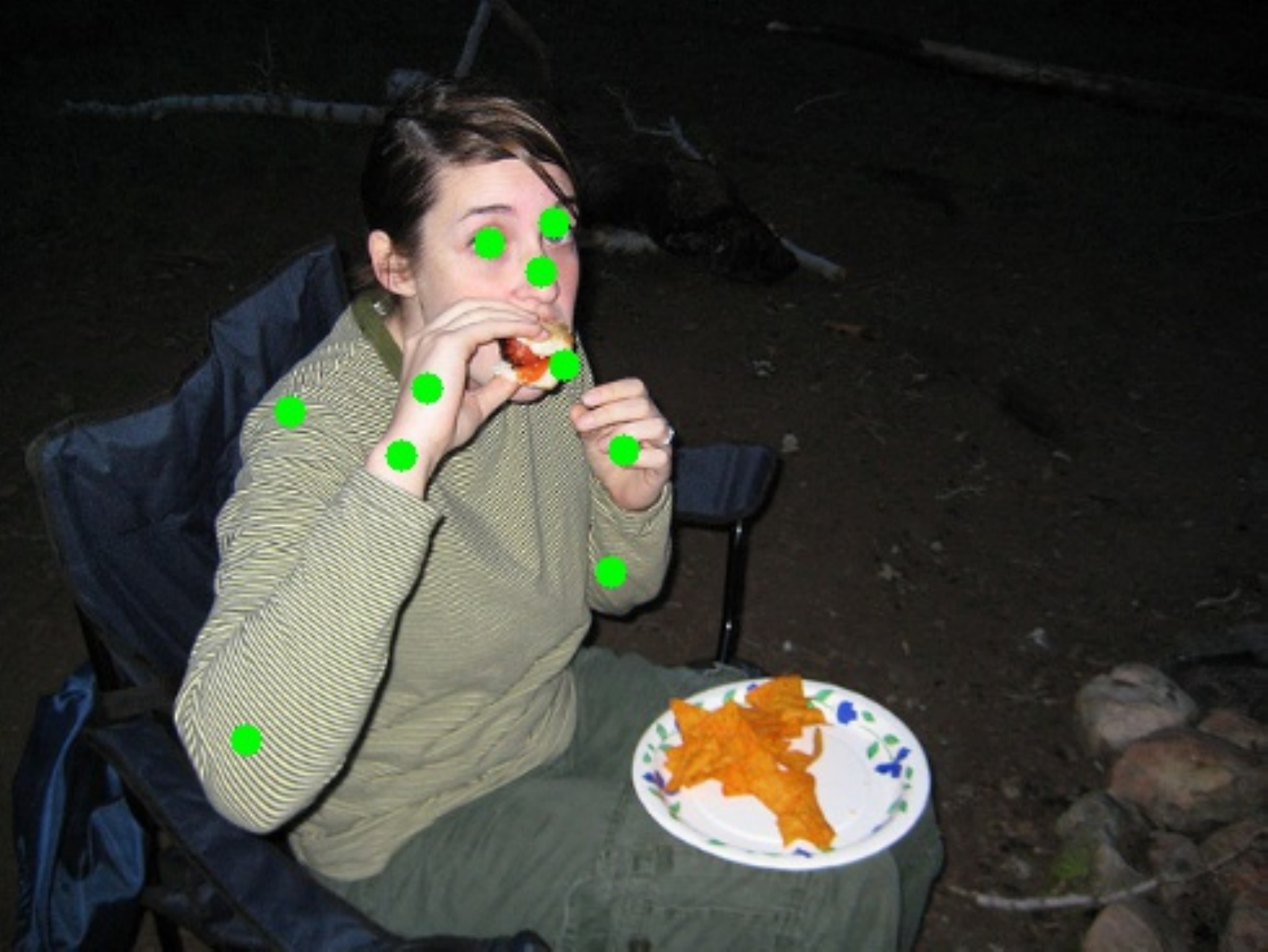} &
  \includegraphics[width=.326\hsize]{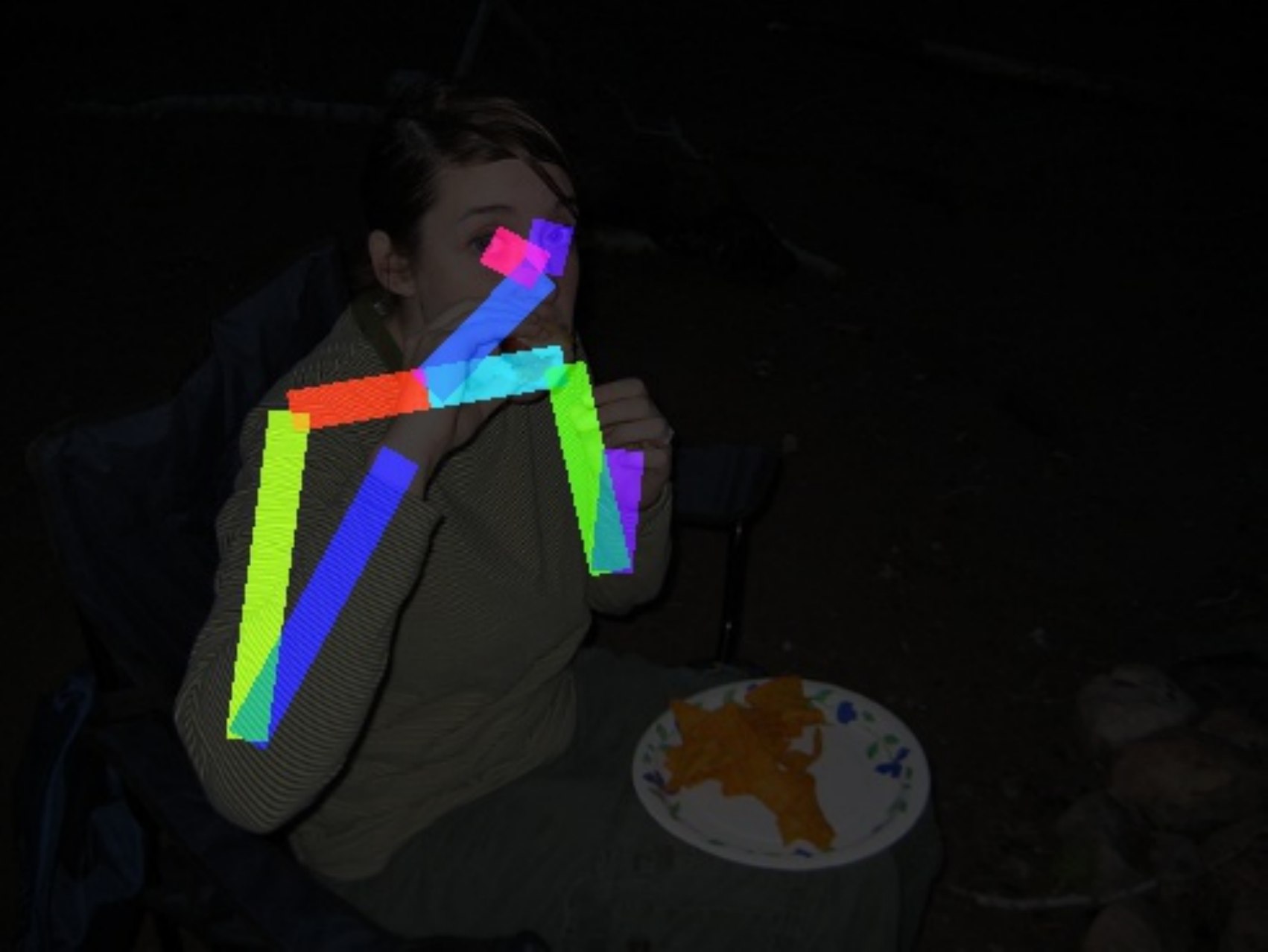} &
  \includegraphics[width=.326\hsize]{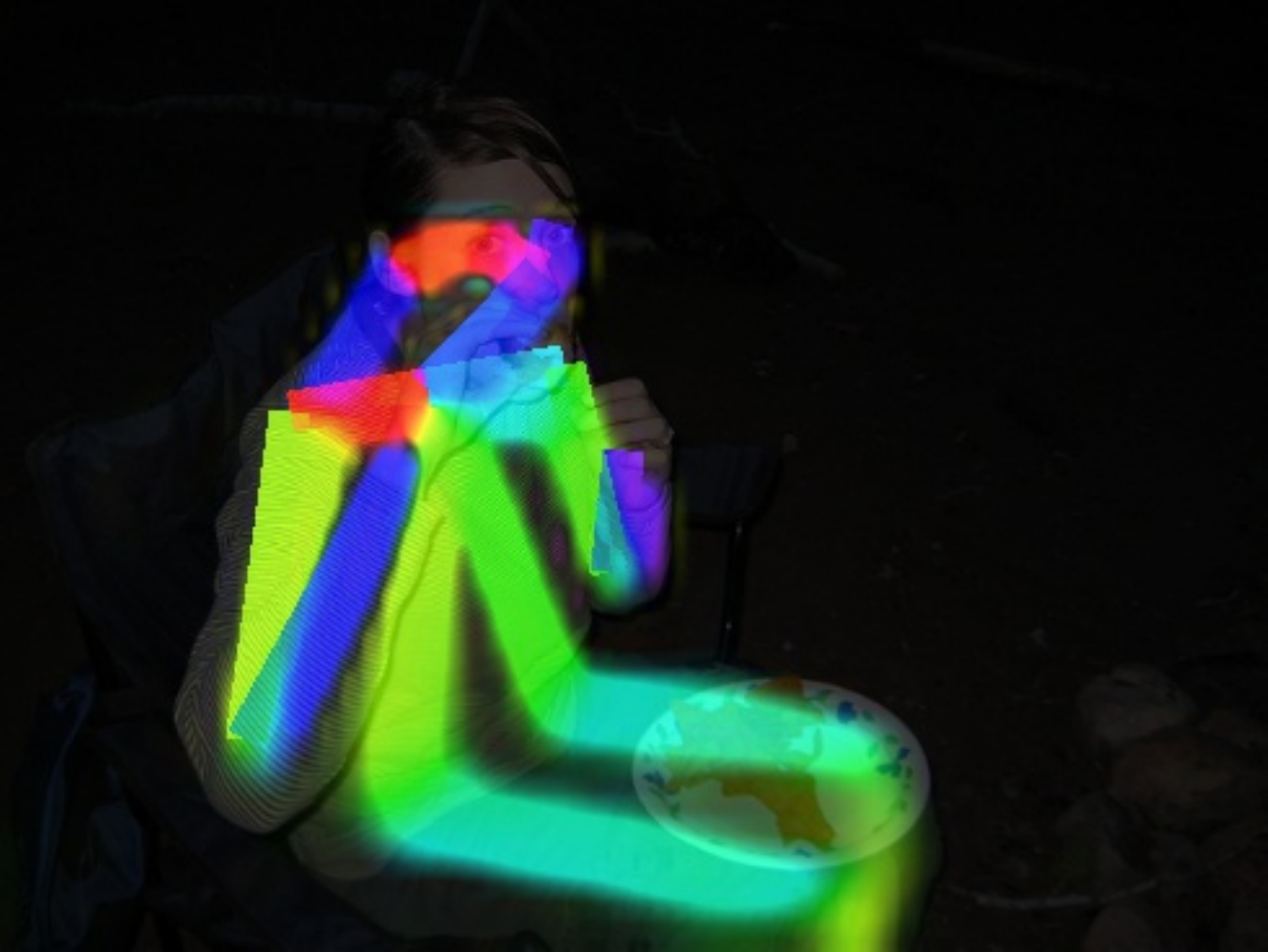} \\
  \scalebox{.92}{Annotated keypoints} &
  \scalebox{.92}{Ground-truth PAFs} &
  \scalebox{.92}{Corrected PAFs}
\end{tabular}}
\end{center}
\caption{Examples of label correction results for PAFs. Each row represents the label correction results for the data that have keypoint protrusions, occlusion, and missed annotations, respectively.}
\label{fig:compare_labels}
\end{figure}

\subsection{Label Correction}
As shown in Figure 1c, there are cases where the output of a learned model is more appropriate than labels generated from annotations even though the model is trained with data including inappropriate labels. In this example, the model estimates PAFs that cannot be generated from the annotations. In addition, the PAFs predicted by the model are located along the limbs smoothly although each of the ground-truth PAF is generated in the rectangular area defined between the pair of corresponding keypoints.
Like this, the learned model is generalized to the features of keypoints and limbs to adequately estimate the existence of keypoints and limbs. Based on such observations, we propose to improve such incomplete labels by correcting them using the output of the learned teacher model.

Let the predictions of a teacher model $P_{T}=(H^{T},L^{T})$ trained using Equation \ref{eq:normal_loss} and ground-truth labels $y_{GT}=(H^{GT},L^{GT})$ which include incomplete training data. We correct the labels by
applying the following functions for each pixel:
\begin{equation}
\label{eq:lc_cmaps}
 H_j^{LC}(p)={\rm max}(H_j^{GT}(p), H_j^{T}(p)),
\end{equation}
\begin{equation}
\label{eq:lc_pafs}
 L_c^{LC}(p) = \begin{cases}
   L_c^{GT}(p) & (\|L_c^{GT}(p)\|_2>\|L_c^{T}(p)\|_2) \\
   L_c^{T}(p) & (otherwise).
\end{cases}
\end{equation}

Figure \ref{fig:compare_labels} shows some examples of corrected PAFs. The labels for missing limbs are supplemented and have become better than the original. Like this, By adopting max operation for correcting labels, existing labels are not changed and missing labels are supplemented by the output of the teacher.
In addition, even if the teacher can not predict the existence of keypoints and limbs, this does not cause negative effect to the final labels.

When we train a new model using corrected labels $y_{LC}$, the mean squared error $E_{LC}$ is used as a loss function,
\begin{equation}
\label{eq:lc_loss}
  E_{LC}=E_{L2}(P_{S},y_{LC}),
\end{equation}
where $P_{S}=(H^{S},L^{S})$ are the outputs of a student model in training. We use a student model that has identical architecture as the teacher model, different from many existing works on knowledge distillation.

% \subsubsection{Label correction with knowledge distillation.}
\noindent{\bf Label correction with knowledge distillation.}
Proposed label correction can be used together with knowledge distillation \cite{hinton2015distilling}. Knowledge distillation is first proposed for classification tasks, but it is also applicable to the pose estimation task in the same way by using the weighted average of losses whose targets are the ground-truth label and the predictions of the teacher model,
\begin{equation}
\label{eq:kd_loss}
  \scalebox{0.94}{$\displaystyle
    E_{KD}=(1-\lambda)E_{L2}(P_{S},y_{GT})+\lambda E_{L2}(P_{S},P_{T}),
  $}
\end{equation}
where $\lambda$ is a parameter representing mixing ratio of the loss functions. In the case of using label correction and knowledge distillation at the same time, we use the loss function $E_{KD, LC}$ which replaces the former term in Equation \ref{eq:kd_loss} with corrected labels,
\begin{equation}
\label{eq:kd_lc_loss}
  \scalebox{0.94}{$\displaystyle
    E_{KD,LC}=(1-\lambda)E_{L2}(P_{S},y_{LC})+\lambda E_{L2}(P_{S},P_{T}).
  $}
\end{equation}
In our experiments, we evaluate the effectiveness of knowledge distillation, and label correction with knowledge distillation as comparison methods for label correction.

% \subsubsection{Iterative learning of label correction.}
\noindent{\bf Iterative learning of label correction.}
Since we can get labels appropriate for training a model via proposed label correction, it might be possible to improve the performance of a model by making the student model the next teacher model and repeating learning. We evaluate the effectiveness of this iterative learning in our experiment.

\begin{figure}[t]
\begin{center}
{\setlength\tabcolsep{1mm}
\begin{tabular}[t]{cc}
  \includegraphics[width=39mm]{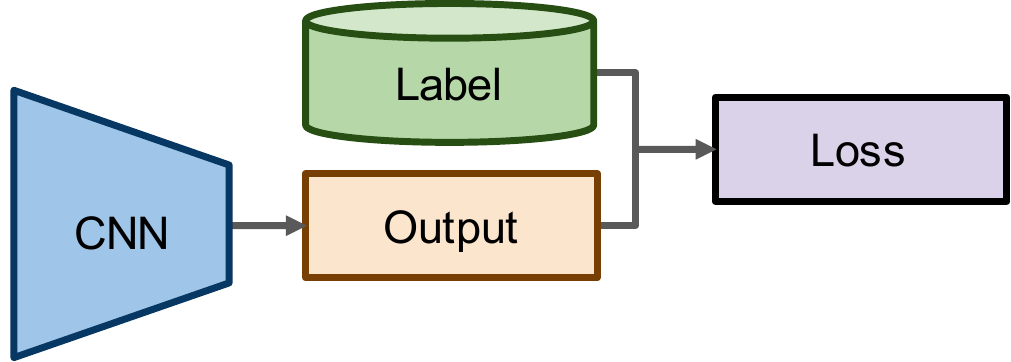} &
  \includegraphics[width=39mm]{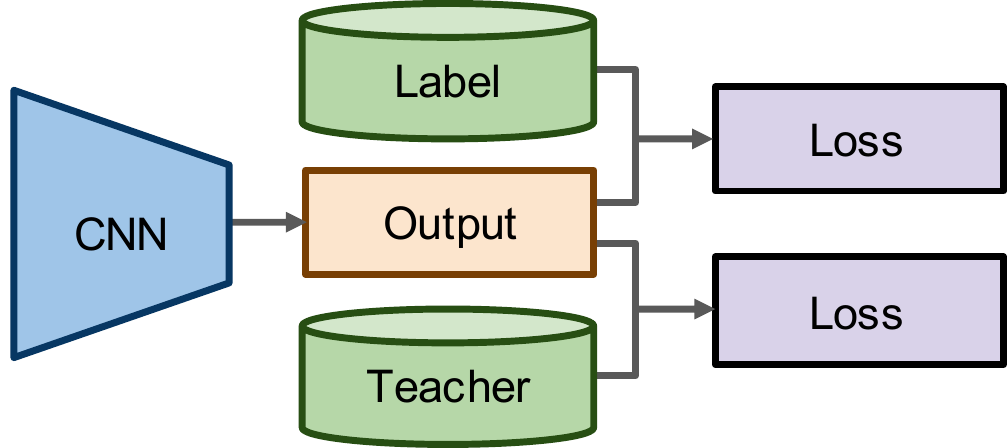} \\
  (a) Normal training & (b) Knowledge Distillation \\ \\
  \includegraphics[width=39mm]{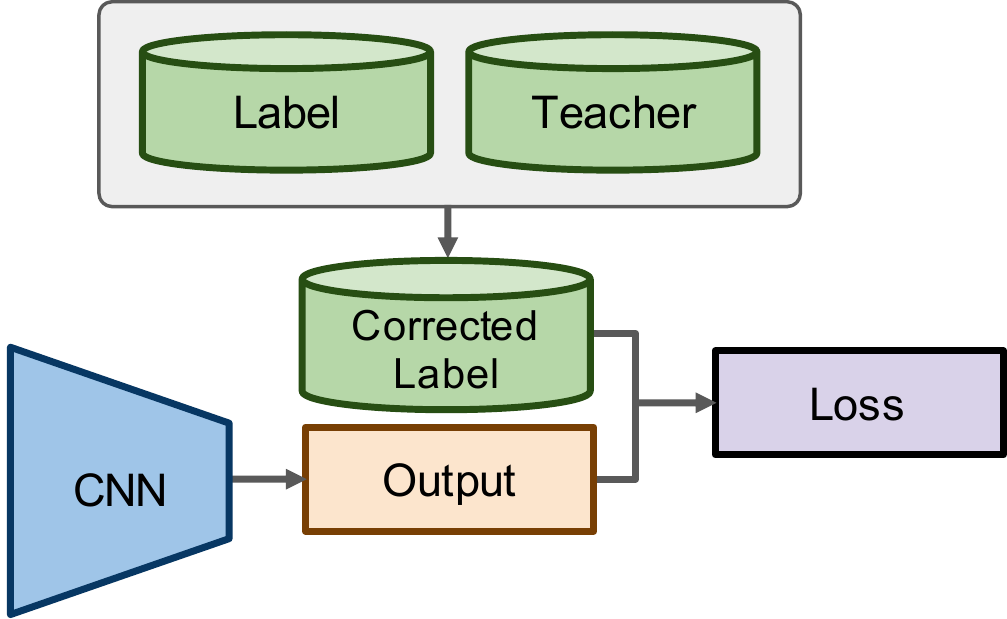} &
  \includegraphics[width=39mm]{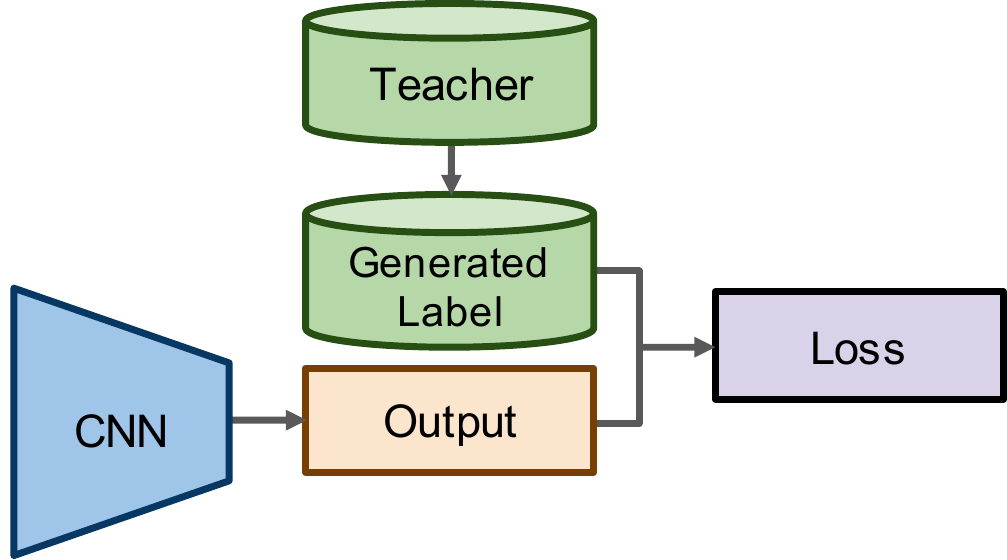} \\
  (c) Label Correction & (d) Data Distillation
\end{tabular}}
\end{center}
\caption{Training schema of each approach.}
\label{fig:comparison_methods}
\end{figure}

% \subsubsection{Relationship to other methods.}
\noindent{\bf Relationship to other methods.}
We now explain the relationship between our proposed label correction and related methods. Figures 4a $\sim$ 4d show the training schema of normal supervised learning with unchanged labels (Equation \ref{eq:normal_loss}), knowledge distillation (Equation \ref{eq:kd_loss}), label correction (Equation \ref{eq:lc_loss}) and data distillation, respectively. Whereas knowledge distillation learns a model with the weighted average of the losses calculated from labels generated from annotations and the outputs of the teacher model, label correction revises the labels directly with the teacher's outputs. Although data distillation seems equivalent to knowledge distillation with $\lambda=1.0$, it differs from knowledge distillation in terms of aiming to generate labels for unlabeled data to enlarge the number of training data and generate keypoint annotations other than making use of the soft outputs of a teacher model to directly generate labels.

% -------------------------------------------------------------

\section{Experiments}
This section describes experimental results that verify the effectiveness of our approach. After explaining the dataset used in the experiment and implementation details, we report quantitative results and qualitative results.

\subsection{Dataset and Evaluation Metric}
We use the COCO 2017 dataset \cite{lin2014microsoft} for evaluation. This dataset consists of real world images including the variety of lighting, clothing, and occlusion, making it highly challenging. We train models with the training set and report the evaluation results with the validation set. As evaluation metrics, we use Average Precision (AP), ${\rm AP}_{50}$, ${\rm AP}_{75}$, ${\rm AP}_{M}$, ${\rm AP}_{L}$ used in the COCO keypoint challenge. The primary challenge metric AP is calculated from the mean AP over 10 Object Keypoint Similarity (OKS) thresholds, where OKS is calculated from scale of the person and the distance between predicted keypoints and the ground-truth keypoints.
${\rm AP}_{50}$ and ${\rm AP}_{75}$ are AP at OKS 0.50, 0.75, respectively. ${\rm AP}_{M}$ and ${\rm AP}_{L}$ represent AP for the medium and large size of persons, respectively.

\subsection{Data Augmentation}
Data augmentation is critical for the learning of scale invariance and rotation invariance. % Todo: 表現変えたい
After applying random flip, random rotation ($-40^\circ\sim+40^\circ$) and random scale ($0.5\sim1.1$), we crop the image to the size of $368\times368$ at a random position which is the model input size.

\subsection{Implementation Details}
We report the evaluation results for two models, CMU-Pose \cite{cao2017realtime} (52.3 M parameters) proposed by Cao et al. and small NN (5.2 M parameters) which is a relatively small model consisting of a stack of 19 convolution layers. For label correction and knowledge distillation, CMU-Pose was used as the teacher network in all the experiments. Regardless of the teacher model and the student model, all the trained models have only the difference between the labels used for training and the loss functions accompanying them, and all the other training settings were set to be the same.

For optimization we use Adam \cite{kingma2014adam} with $\beta_1=0.9$. All the models were trained with $10^5$ update iterations at $\alpha=10^{-4}$ and another $10^5$ iterations at $\alpha=10^{-5}$. In all the experiments, the first 10 layers of VGG-19 in CMU-Pose were initialized with the weights of the ImageNet \cite{ussakovsky2015imagenet} pretrained model.

% -------------------------------------------------------------

\begin{table}
\caption{Evaluation results on the COCO validation set. $\times 2$ denotes the model trained with iterative learning that makes the trained student model the next teacher model.}
\label{tab:results}
\begin{center}
(a) CMU-Pose
\vspace{.8mm} \\
\scalebox{0.86}{
\begin{tabular}{l|c|cc|cc}
  Method                & AP         & ${\rm AP_{50}}$ & ${\rm AP_{75}}$ & ${\rm AP_M}$ & ${\rm AP_L}$ \\ \hline \hline
  Baseline                & 54.6       & 80.9       & 56.4       & 51.7       & 60.9       \\
  KD ($\lambda=0.5$)       & 55.6       & 82.0       & 59.3       & 51.7       & 63.1       \\
  % KD($\lambda=0.5$)*      & 56.4       & 82.1       & 60.1       & 52.5       & {\bf 63.6} \\
  KD ($\lambda=1$)         & 54.3       & 80.1       & 58.2       & 51.1       & 60.9       \\
  % KD($\lambda=1$)*        & 55.4       & 81.9       & 55.6       & 51.9       & 62.4       \\
  LC                      & {\bf 56.9} & 82.8       & 58.9       & {\bf 53.9} & 63.1       \\
  LC $\times$2                   & {\bf 56.9} & {\bf 83.8} & 57.9       & 53.5       & {\bf 63.5} \\
  % LC*                     & 56.2       & 82.3       & 57.6       & 52.5       & 63.1       \\
  KD ($\lambda=0.5$) + LC    & 56.8       & 81.9       & {\bf 59.8} & 53.7       & 63.0       \\
  KD ($\lambda=0.5$) + LC $\times$2 & 56.3       & 82.0       & 57.7       & 52.8       & 63.2       \\
  % KD($\lambda=0.5$)*+LC   & 56.5       & 82.2       & {\bf 60.6} & 53.8       & 62.6       \\
\end{tabular}}
\vspace{2mm} \\
(b) small NN
\vspace{.8mm} \\
\scalebox{0.87}{
\begin{tabular}{l|c|cc|cc}
  Method                & AP         & ${\rm AP_{50}}$ & ${\rm AP_{75}}$ & ${\rm AP_M}$ & ${\rm AP_L}$ \\ \hline \hline
  Baseline                & 44.0       & 76.0       & 44.5       & 41.8       & 49.3       \\
  KD ($\lambda=0.5$)       & 44.5       & 76.6       & 41.7       & 43.2       & 48.5       \\
  % KD($\lambda=0.5$)*      & 45.9       & {\bf 79.0} & 42.7       & 44.2       & 50.6       \\
  % KD($\lambda=1$)*        & 46.7       & 78.2       & 43.8       & 44.4       & 51.7       \\
  LC                      & 45.4       & 76.3       & 44.5       & 43.1       & 50.2       \\
  KD ($\lambda=0.5$) + LC    & {\bf 45.8} & {\bf 77.4} & {\bf 45.2} & {\bf 43.8} & {\bf 51.2} \\
  % KD($\lambda=0.5$)*+LC   & {\bf 46.9} & 78.1       & 45.0       & {\bf 44.8} & {\bf 51.9} \\
\end{tabular}}
\end{center}
\end{table}

% -------------------------------------------------------------

\subsection{Main Results}
In this section, we report evaluation results measured with CMU-Pose and small NN learned changing labels and loss functions. We compare the performance of a model trained with unchanged labels with Equation \ref{eq:normal_loss} (Baseline), knowledge distillation with Equation \ref{eq:kd_loss} (KD), label correction with Equation \ref{eq:lc_loss} (LC), trained simultaneously with label correction and knowledge distillation with Equation \ref{eq:kd_lc_loss} (KD + LC), and iterative learning that makes the learned student model the next teacher model. Note that in the cases of training with label correction or knowledge distillation, The baseline model is used as their teacher model.

% \subsubsection{Results.}
\noindent{\bf Results.}
Table \ref{tab:results}a. shows results with CMU-Pose.

When models are trained with knowledge distillation, the AP increases 1.0 point from the baseline model when $\lambda=0.5$, but the performance drops in the case of $\lambda=1.0$.
The effectiveness of using both of the loss term from unchanged labels and the output of the teacher model is observed.
This result is similar to the effectiveness of using both hard targets and soft targets in knowledge distillation for classification \cite{hinton2015distilling}.
In addition, this result is similar to the fact that the student model which has identical architecture to the teacher model exceeds the performance of the teacher model by
knowledge distillation in classification \cite{furlanello2017born}.

The model trained with label correction exceeds 2.3 points of AP from the baseline model, which is the largest performance gain obtained in our experiments. It seems that proposed label correction adequately improves the quality of training data as we thought would be the case.
In training with knowledge distillation, the loss of the former term of Equation \ref{eq:kd_loss} calculated using data which includes inappropriate labels still seems to have negative effect on training.
In contrast, in training with corrected labels, such labels are directly modified, which mitigates the negative effect for the training, as is thought to be the reason the model trained with label correction exceeds the performance of the model trained with knowledge distillation.

\begin{figure}[t]
  \begin{center}
  {\setlength\tabcolsep{.8mm}
  \begin{tabular}{cc}
    \includegraphics[width=.48\hsize]{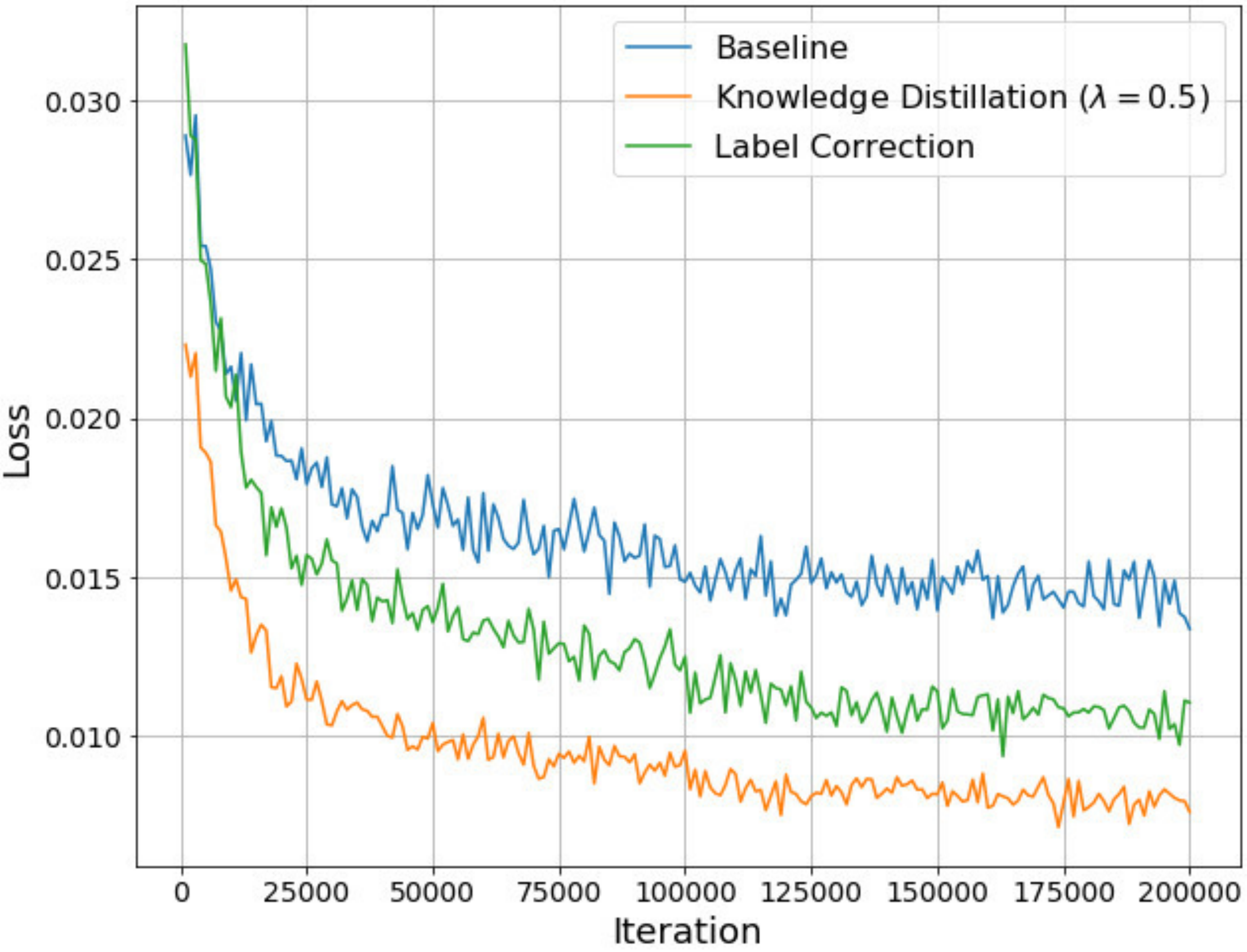} &
    \includegraphics[width=.48\hsize]{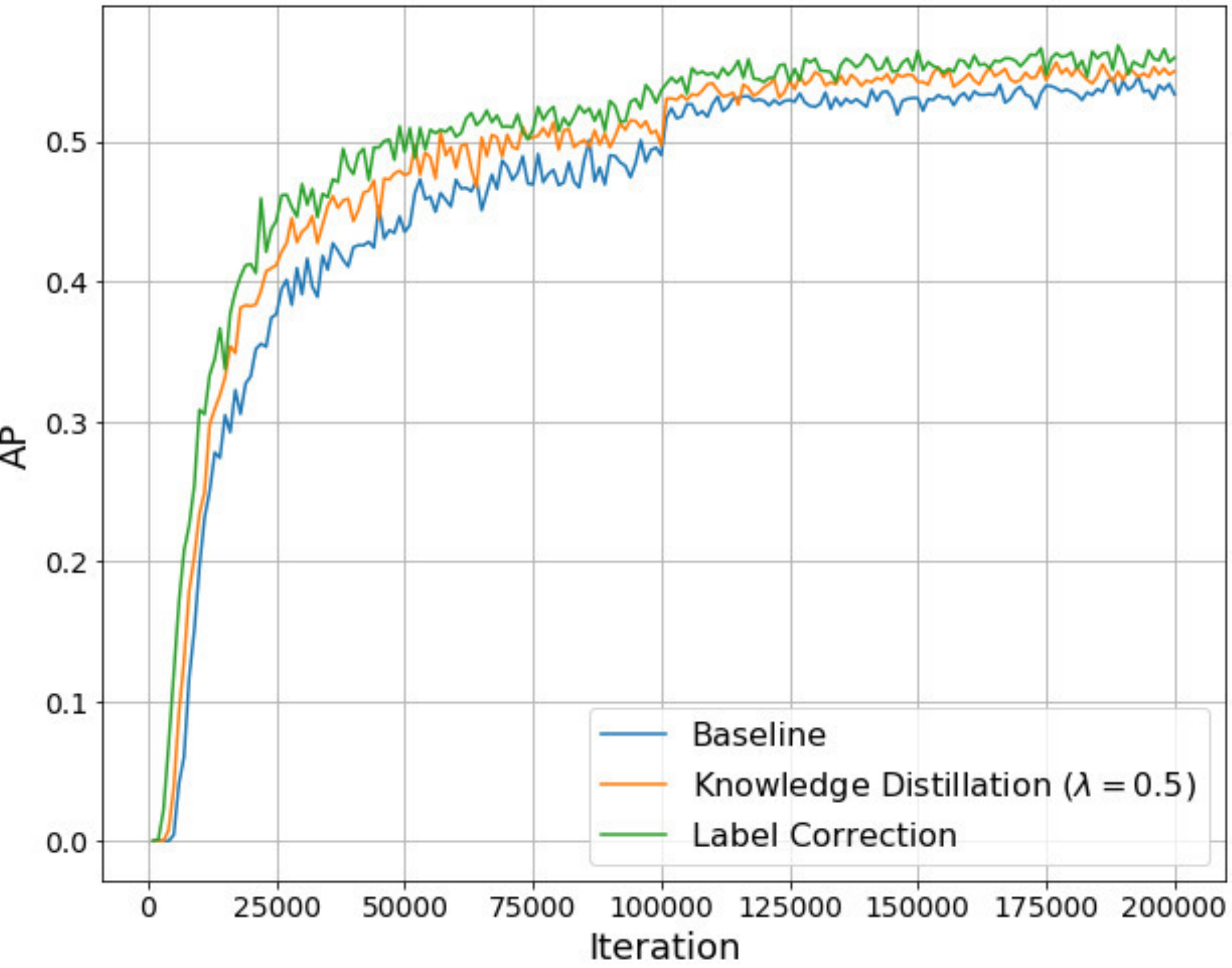} \\
    (a) Loss curve & (b) AP curve
  \end{tabular}}
  \end{center}
  \caption{Learning curves of the models.}
  \label{fig:learning_curve}
\end{figure}

Simultaneous learning of label correction and knowledge distillation results in almost the same performance as label correction. This result seems to be caused by the corrected labels partly including soft output of the teacher model, which makes label correction behave similarly to knowledge distillation. The performance gain was not observed by iterative learning of models, which imply that it is sufficient to use corrected labels once to get proper labels.

Table \ref{tab:results}b shows results evaluated with small NN. Although it is similar to CMU-Pose in that the performance gain obtained by label correction is larger than that of knowledge distillation, performance is best when we use label correction and knowledge distillation simultaneously in this case. Similar to the fact that knowledge distillation is effective for model compression \cite{hinton2015distilling}, it might be effective to directly use the soft output of a teacher model as a training signal for learning a small model with limited expression ability.

Figure \ref{fig:learning_curve} shows learning curve of each model. In the training using the corrected labels, the loss changes between baseline and knowledge distillation. Additionally, label correction not only improves the performance of the trained model, but also speeds up convergence of a model. These facts also imply that label correction has the effect of improving the performance of the model by correcting inappropriate labels while including the properties observed in knowledge distillation.

% -------------------------------------------------------------

\begin{table}[t]
\caption{Ablation study of label correction. This experiment uses CMU-Pose as model architecture.}
\label{tab:ablation_study}
\begin{center}
\scalebox{0.87}{
\begin{tabular}{l|c|cc|cc}
  Method                & AP   & ${\rm AP_{50}}$ & ${\rm AP_{75}}$ & ${\rm AP_M}$ & ${\rm AP_L}$ \\ \hline \hline
  Baseline              & 54.6 & 80.9 & 56.4 & 51.7 & 60.9 \\
  LC (Confidence maps)  & 55.5 & 80.6 & 56.9 & 52.4 & 62.4 \\
  LC (PAFs)             & 55.5 & 81.9 & 57.3 & 52.0 & 62.7 \\
  LC (Both)             & 56.9 & 82.8 & 58.9 & 53.9 & 63.1 \\
\end{tabular}}
\end{center}
\end{table}

\begin{figure*}
\begin{center}
{\setlength\tabcolsep{.3mm}
\begin{tabular}{ccccccccc}
  \raisebox{15mm}{(a)} &
  \includegraphics[width=.133\hsize]{result/00477689_gt_joints} &
  \includegraphics[width=.133\hsize]{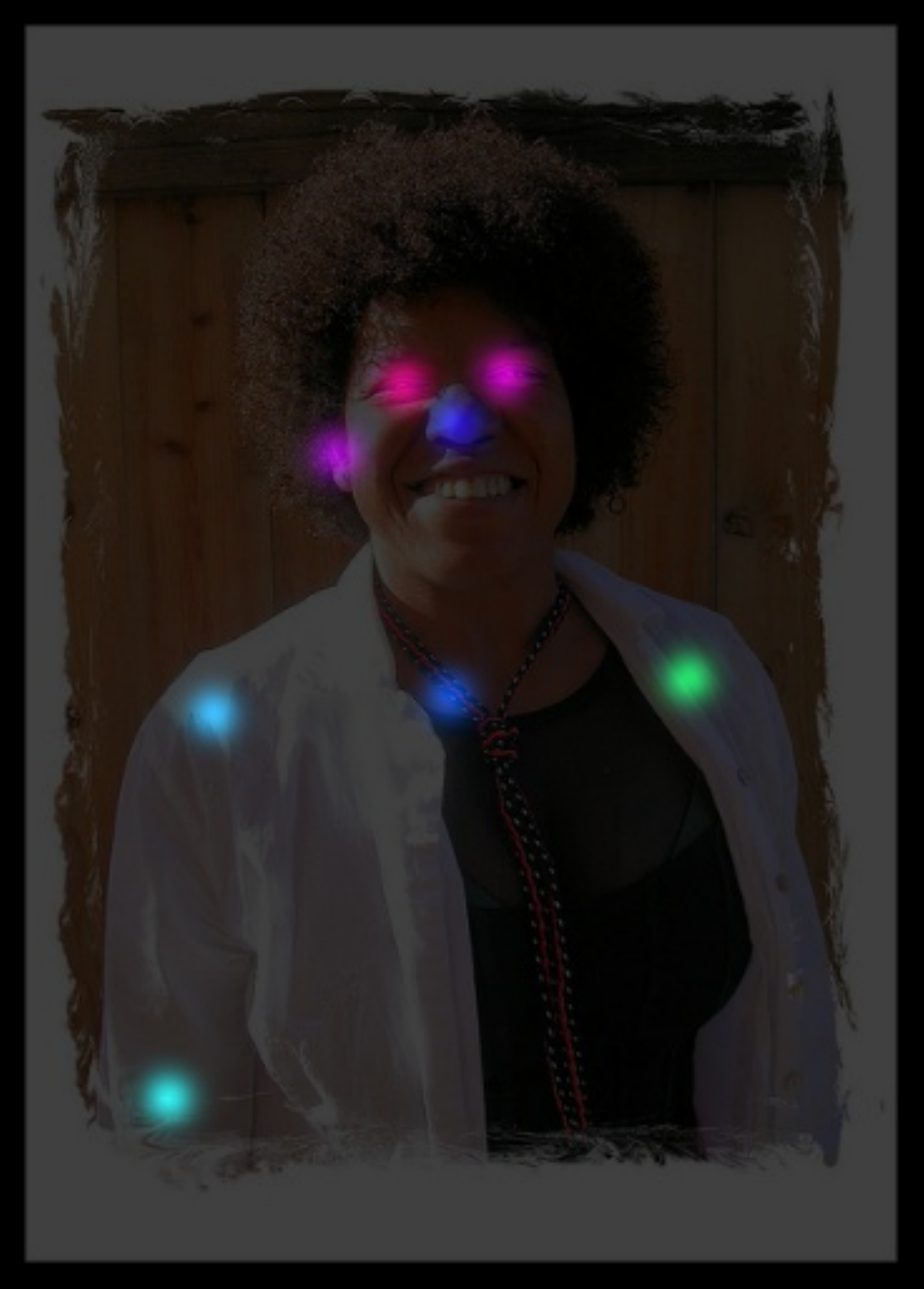} &
  \includegraphics[width=.133\hsize]{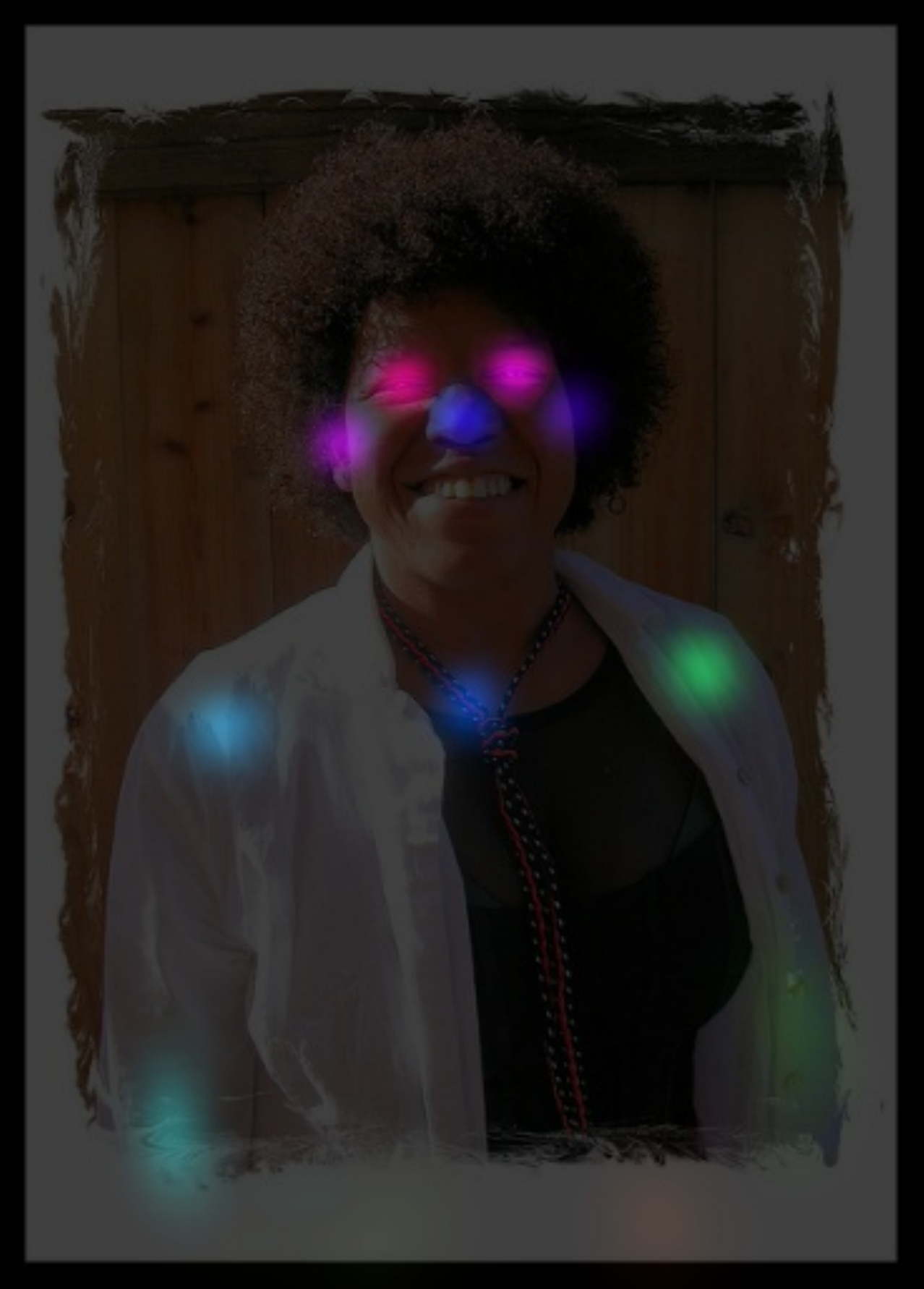} &
  \includegraphics[width=.133\hsize]{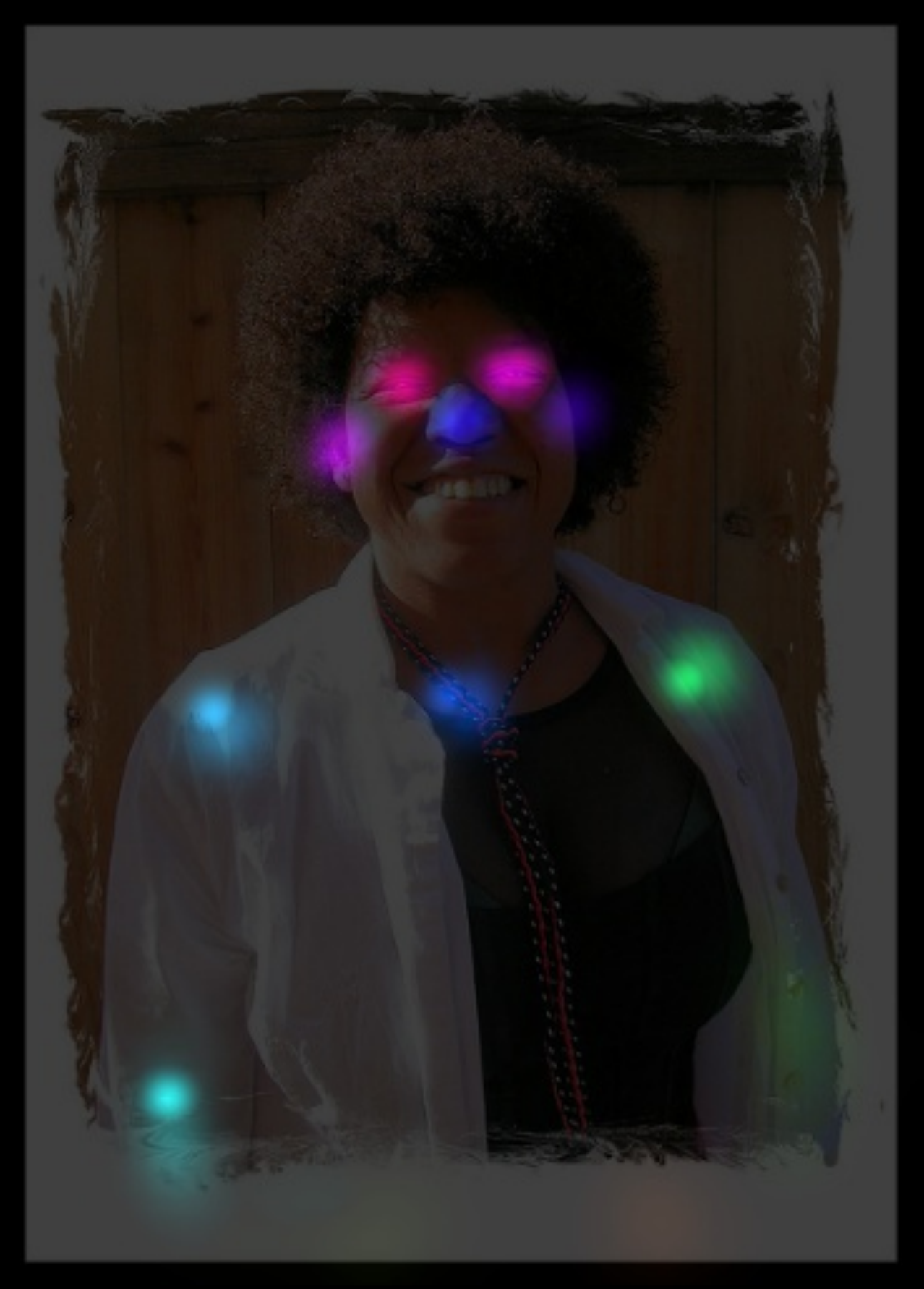} &
  \includegraphics[width=.133\hsize]{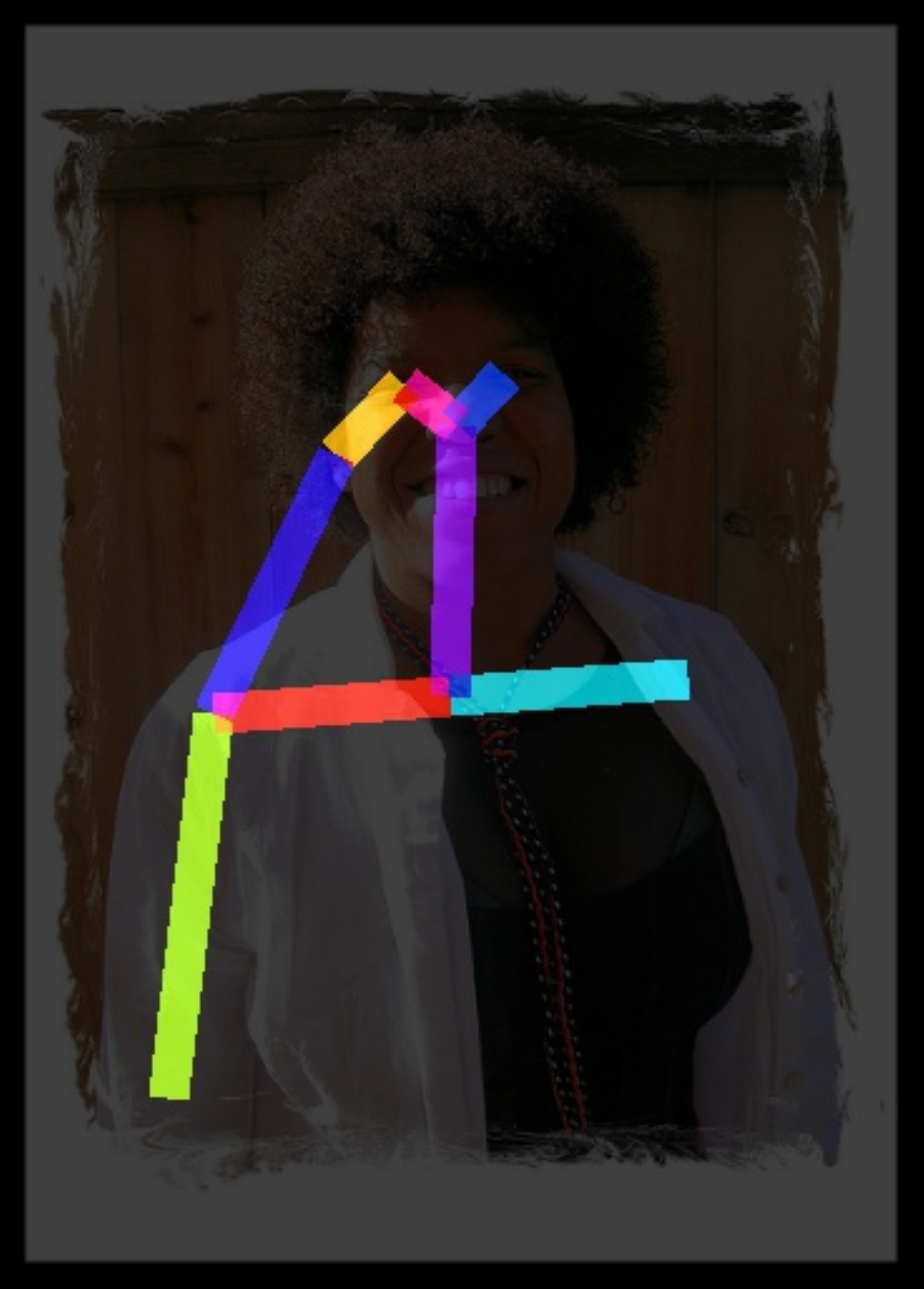} &
  \includegraphics[width=.133\hsize]{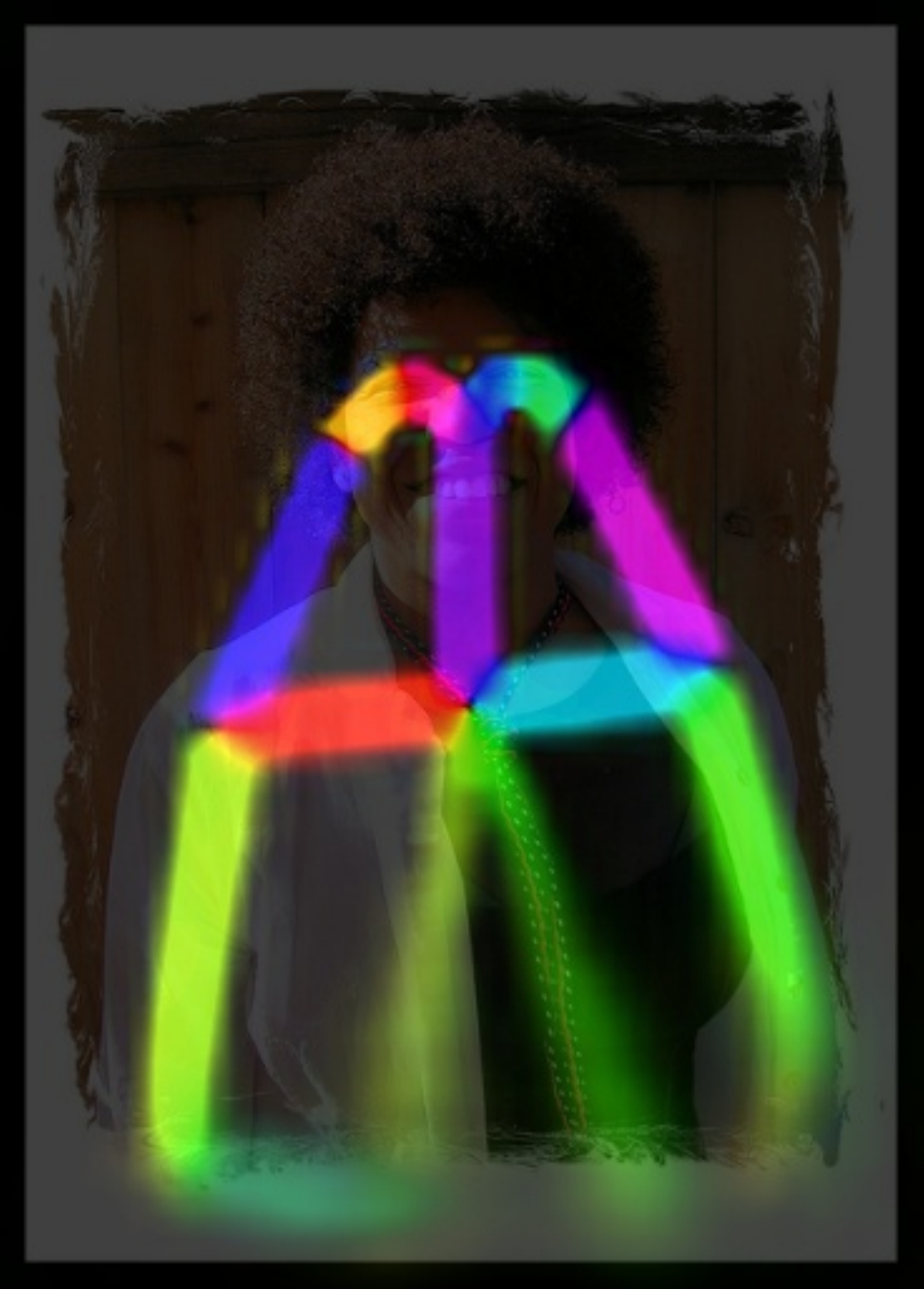} &
  \includegraphics[width=.133\hsize]{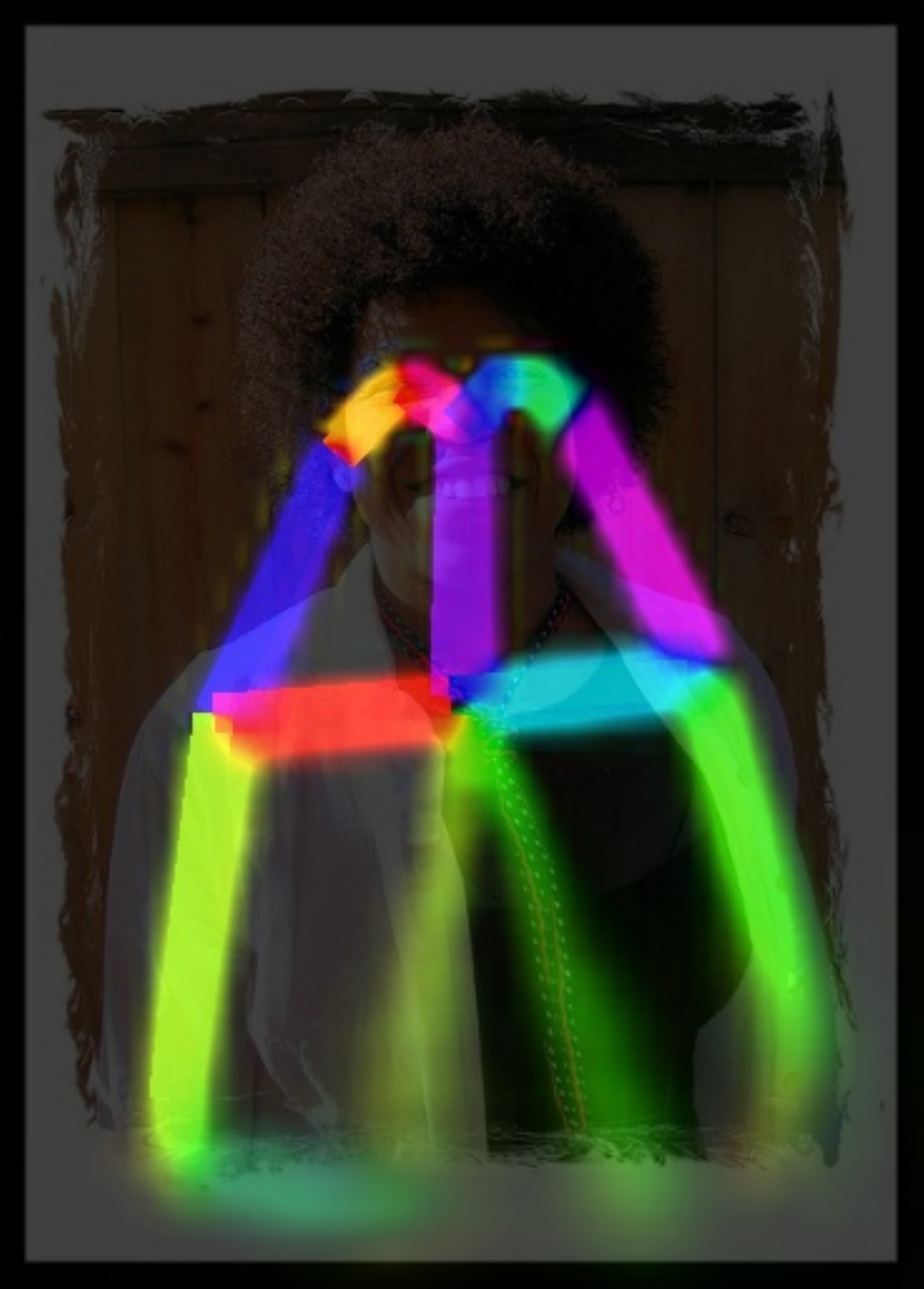} \\

  \raisebox{15mm}{(b)} &
  \includegraphics[width=.133\hsize]{result/00271997_gt_joints} &
  \includegraphics[width=.133\hsize]{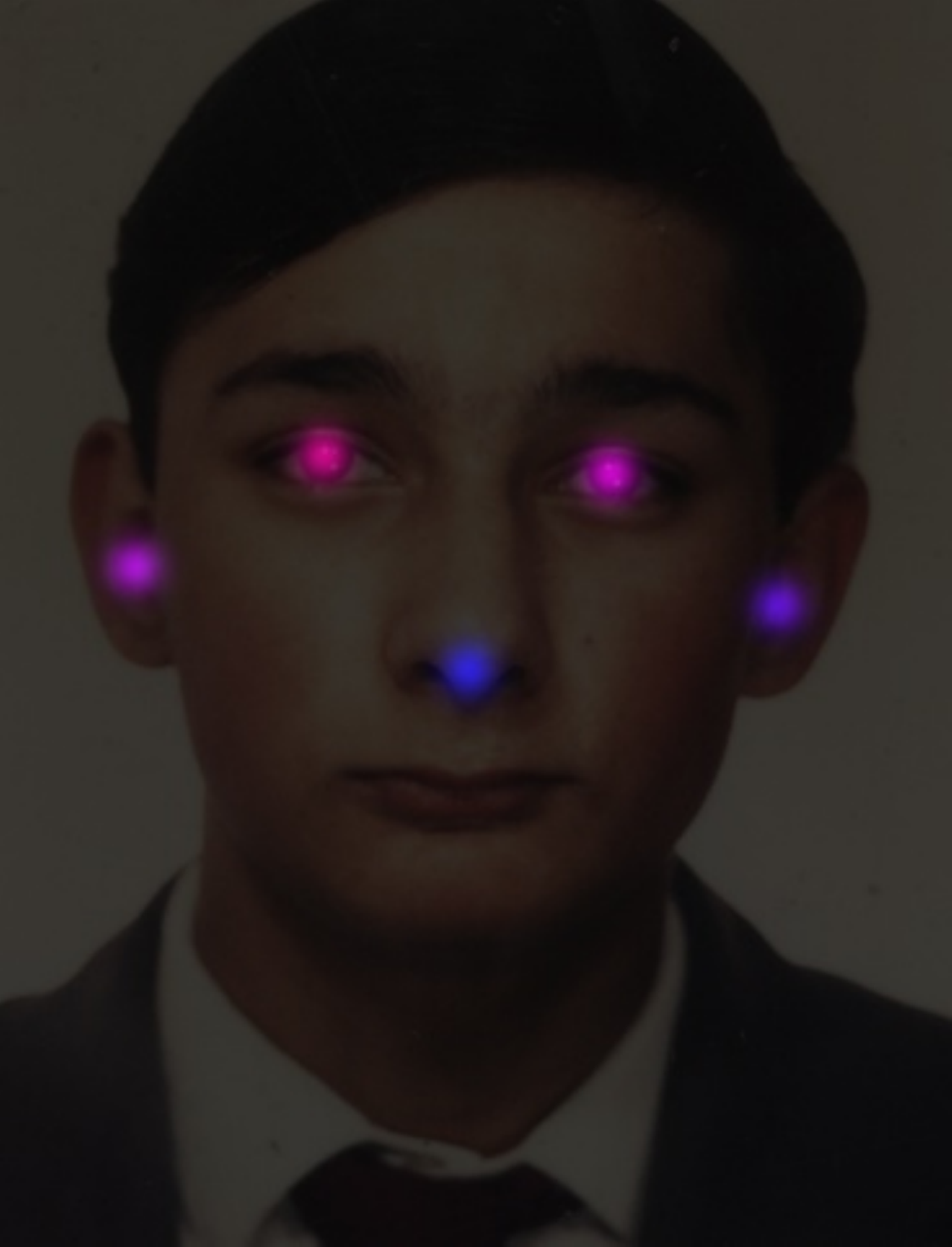} &
  \includegraphics[width=.133\hsize]{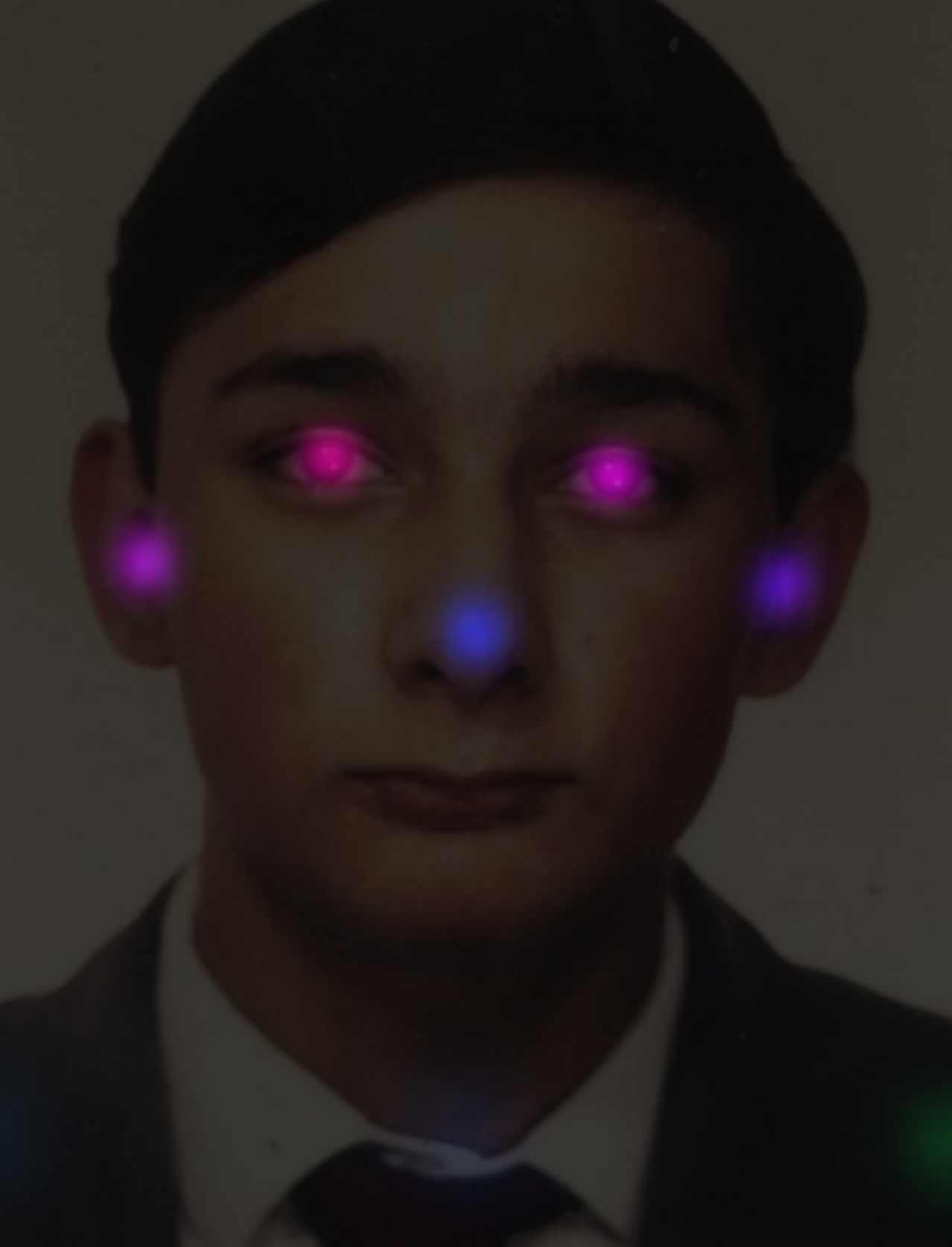} &
  \includegraphics[width=.133\hsize]{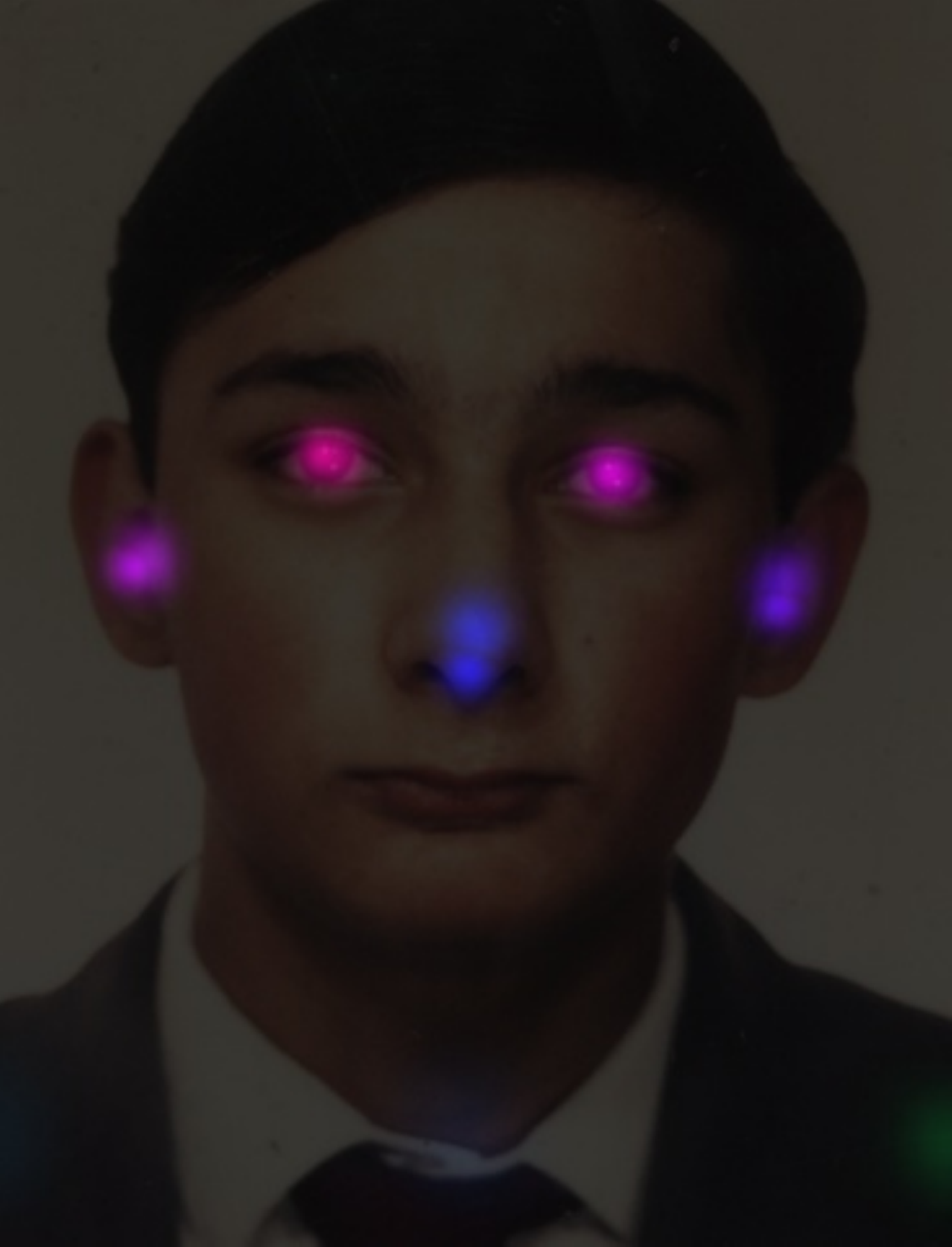} &
  \includegraphics[width=.133\hsize]{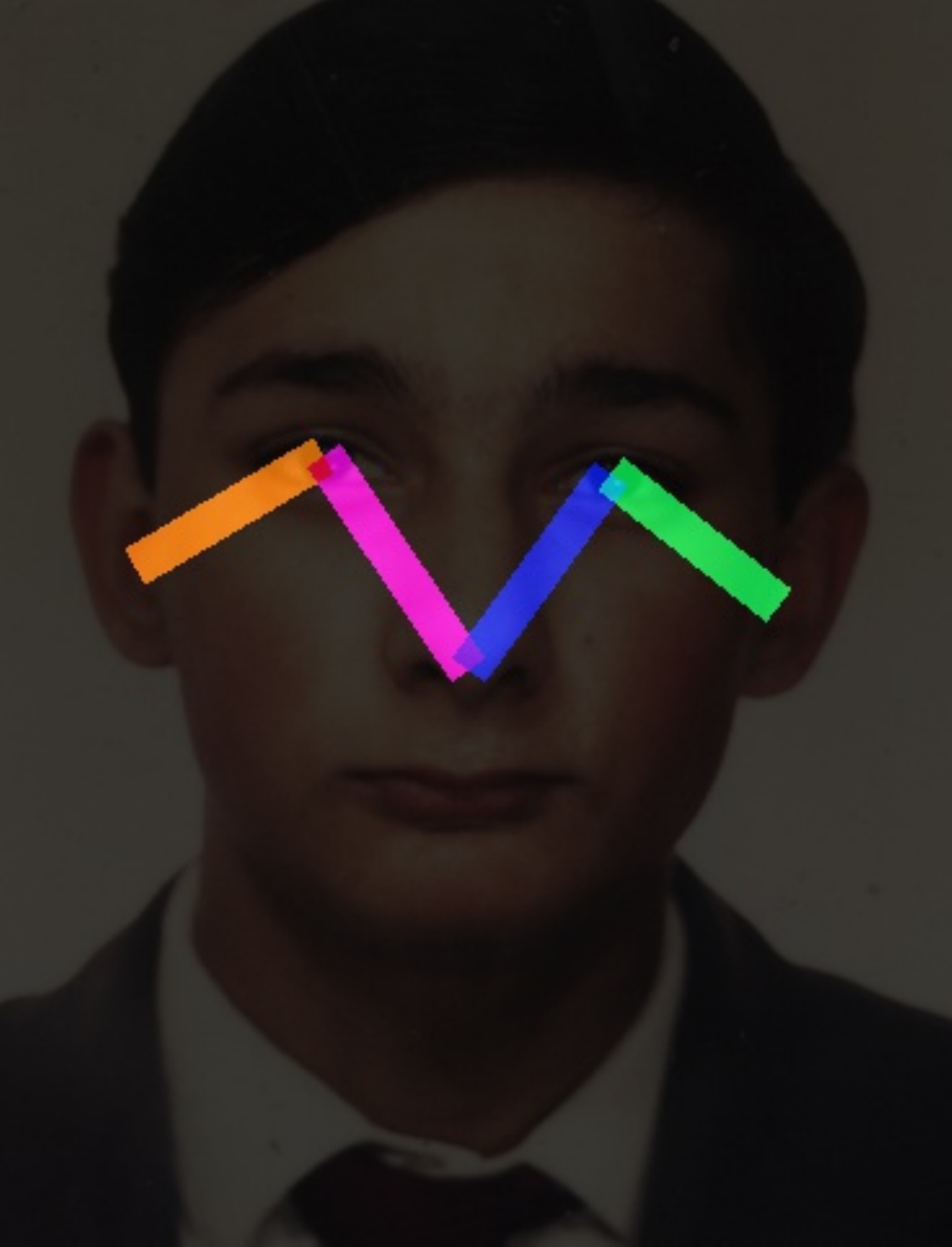} &
  \includegraphics[width=.133\hsize]{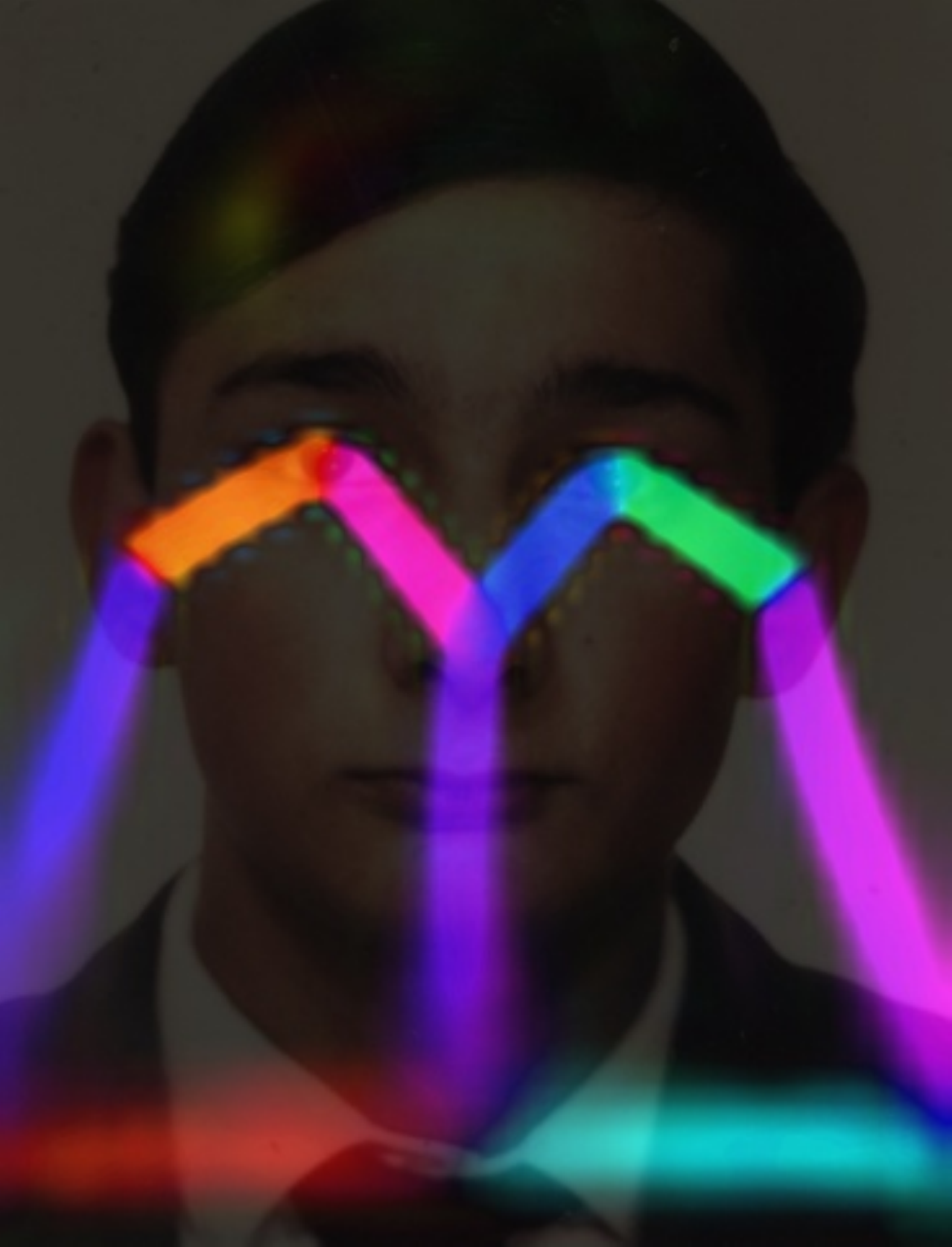} &
  \includegraphics[width=.133\hsize]{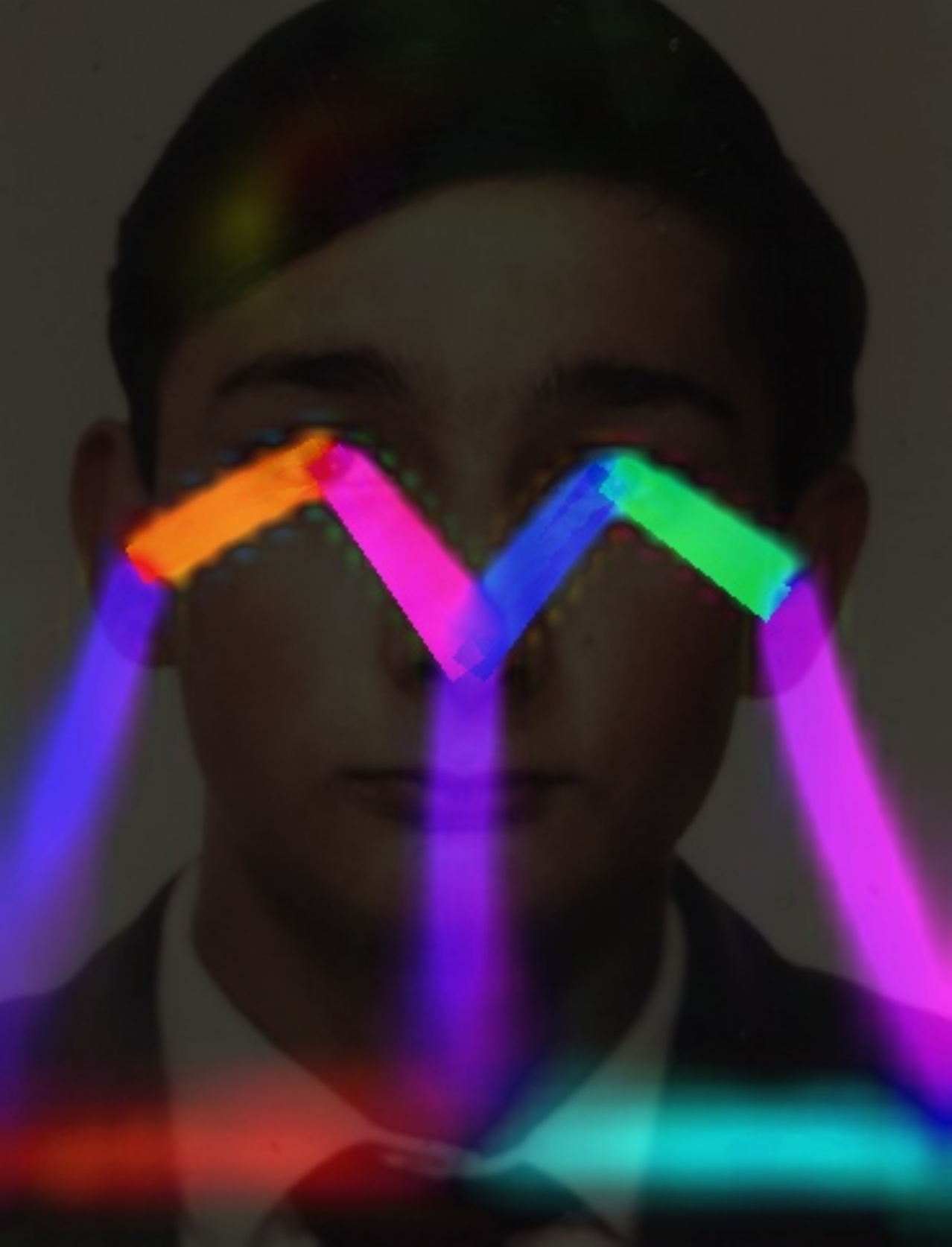} \\

  \raisebox{6mm}{(c)} &
  \includegraphics[width=.133\hsize]{result/00327592_gt_joints} &
  \includegraphics[width=.133\hsize]{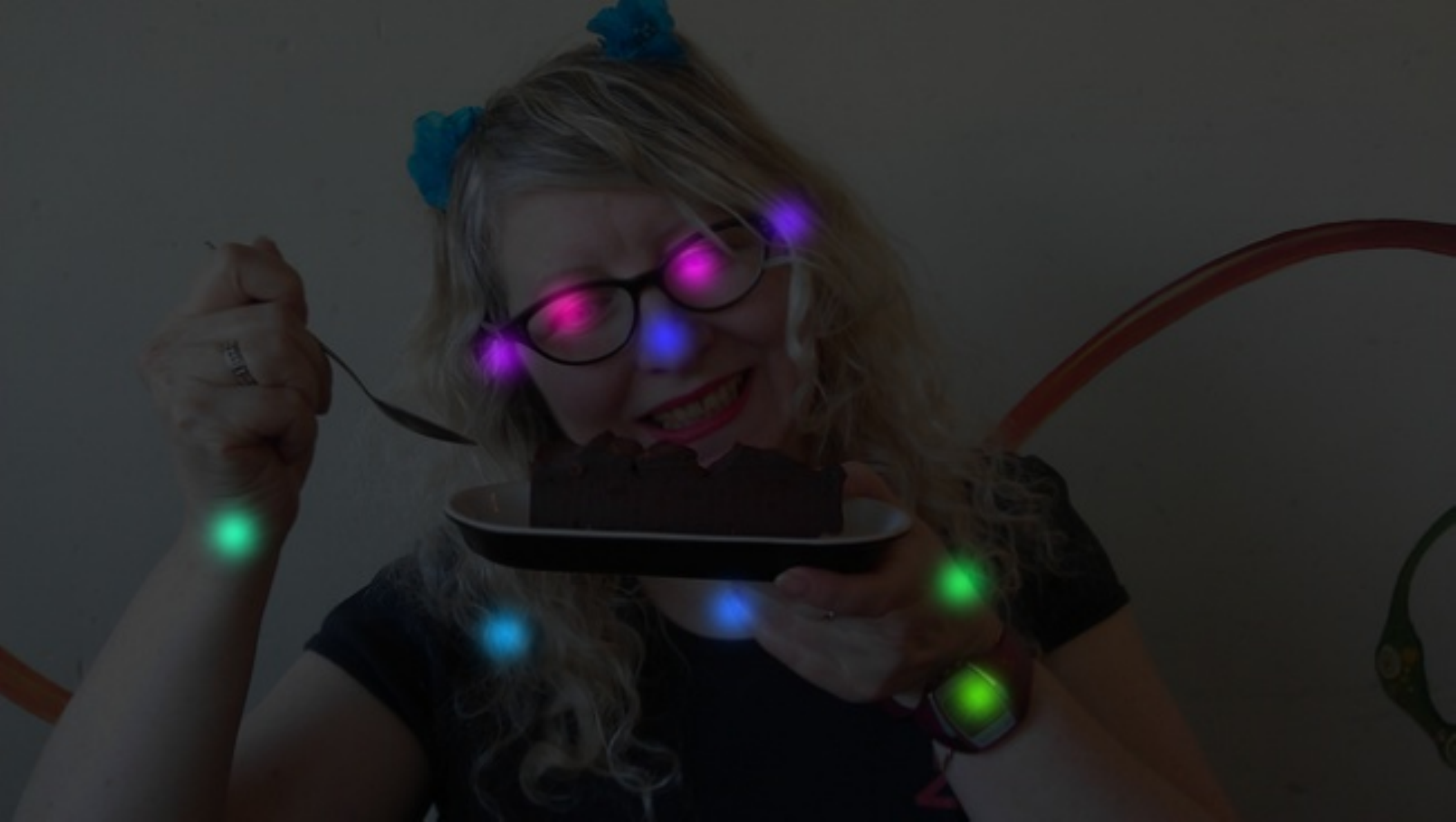} &
  \includegraphics[width=.133\hsize]{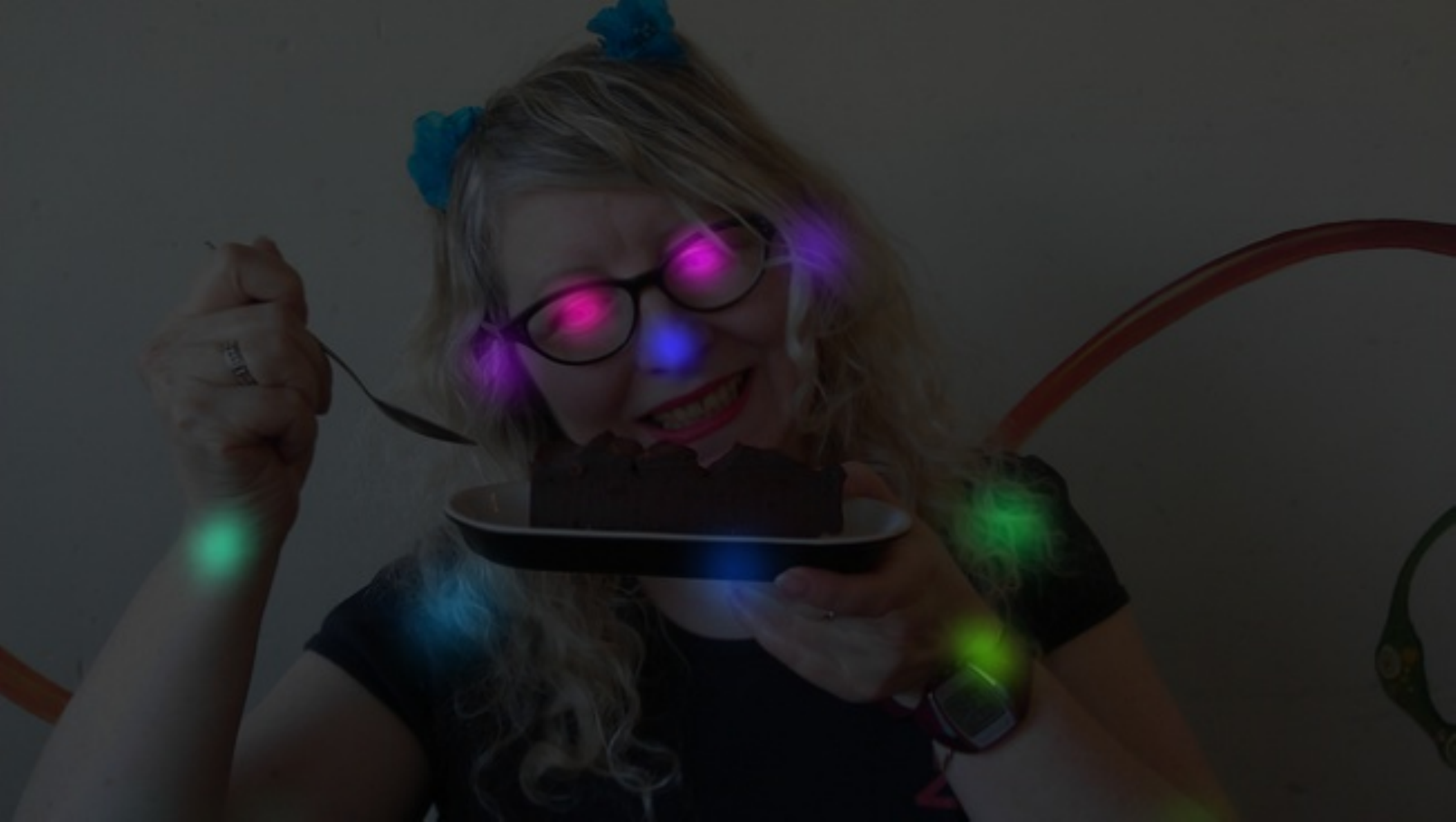} &
  \includegraphics[width=.133\hsize]{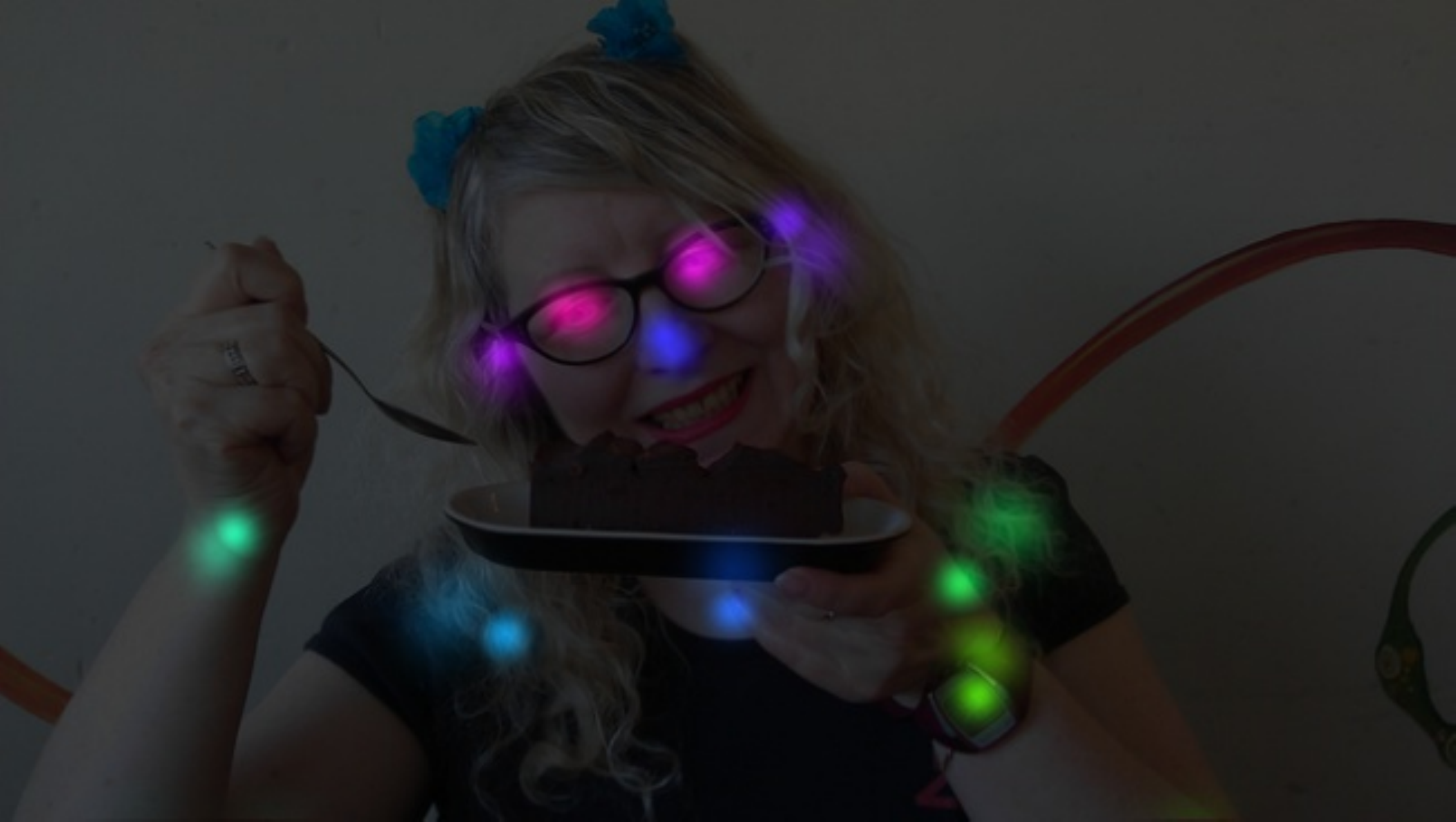} &
  \includegraphics[width=.133\hsize]{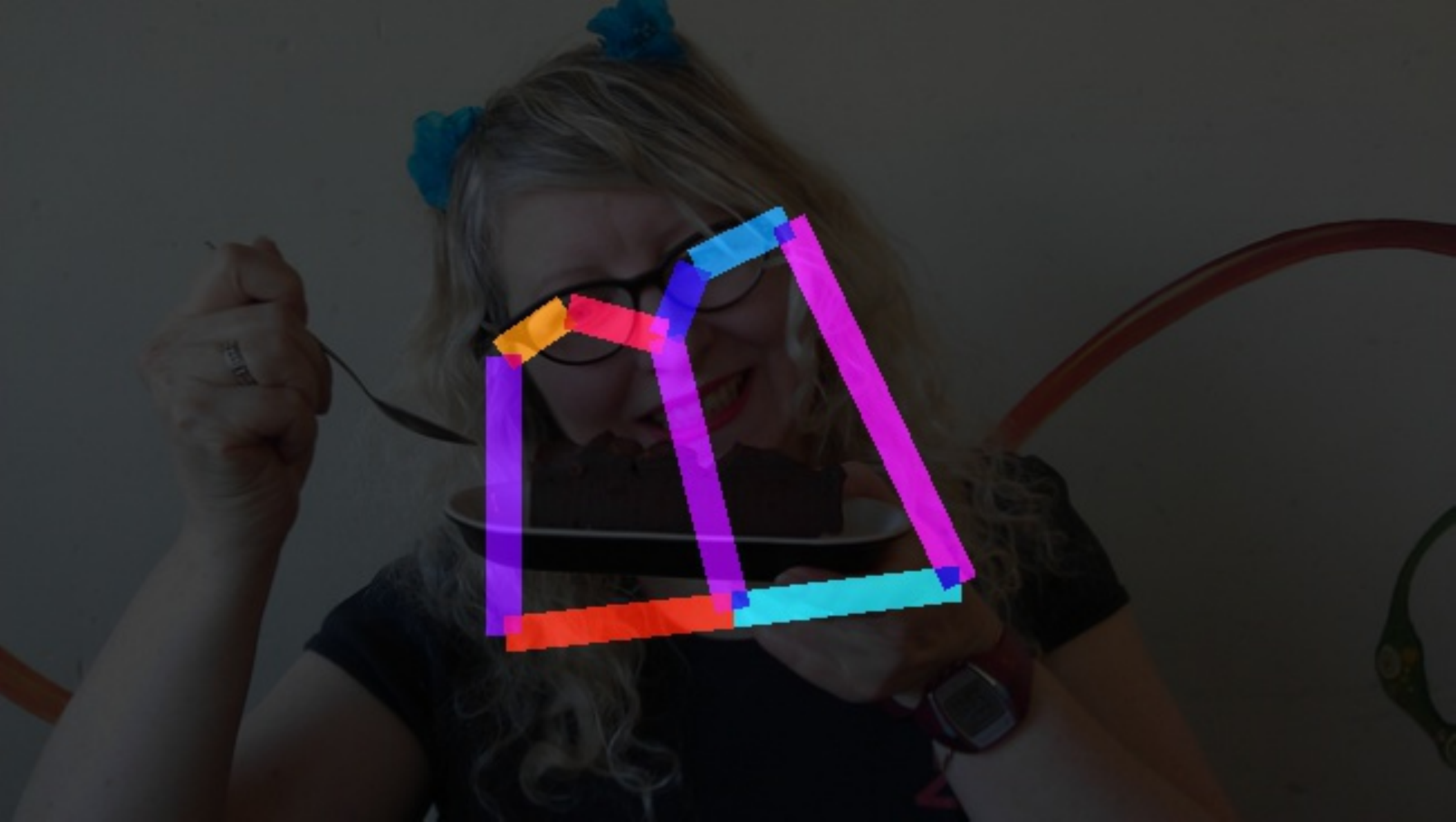} &
  \includegraphics[width=.133\hsize]{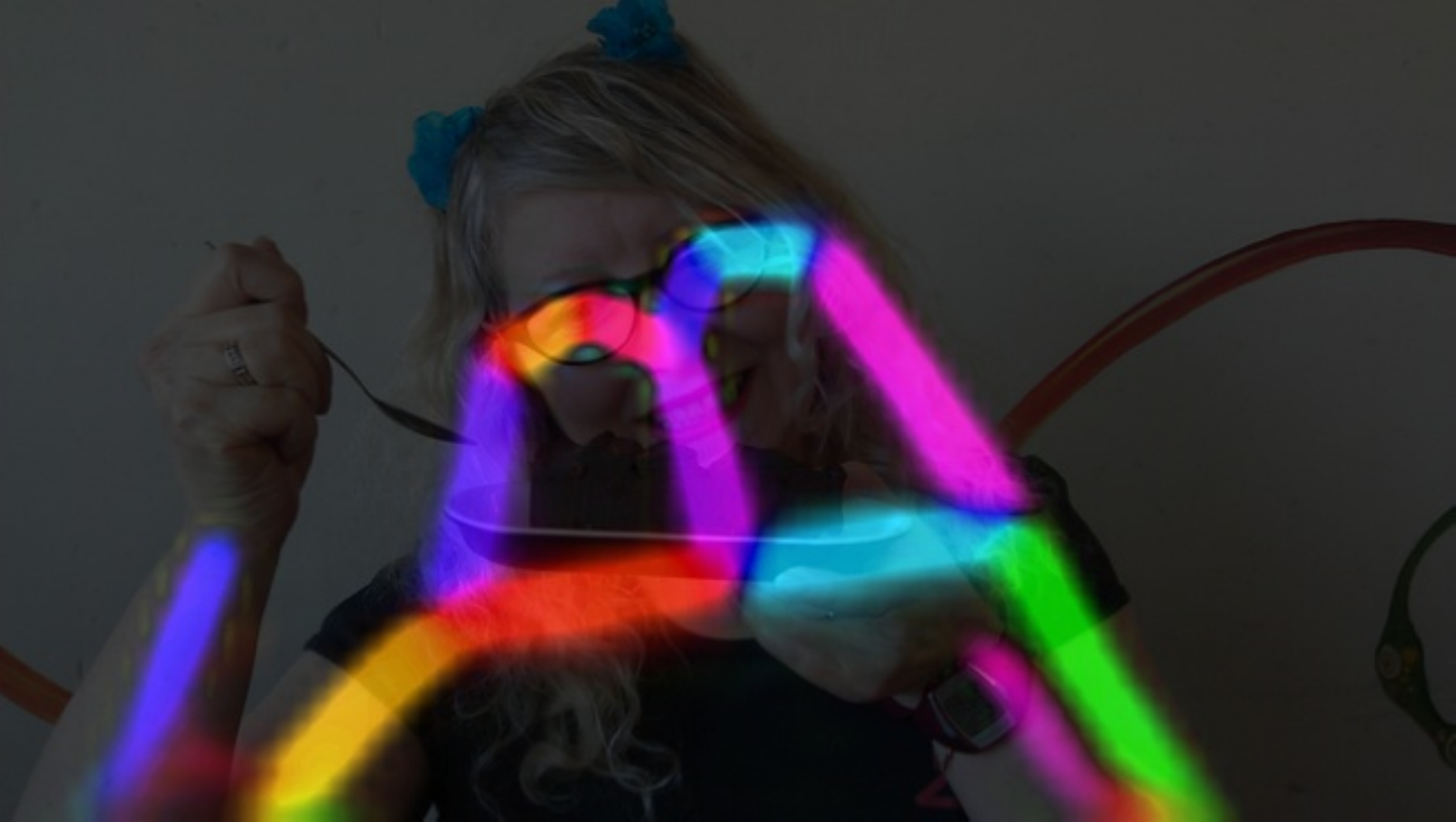} &
  \includegraphics[width=.133\hsize]{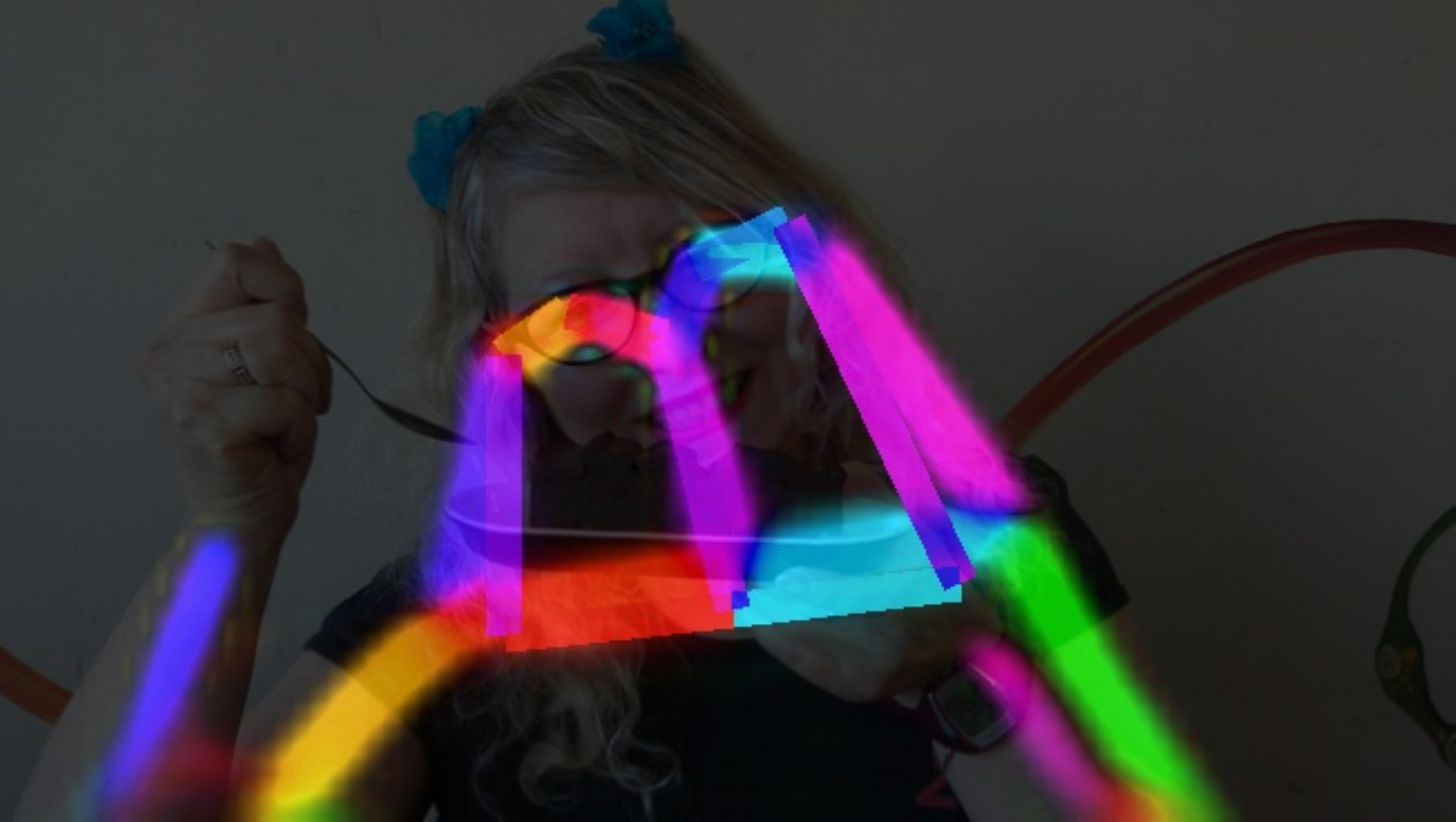} \\

  \addlinespace[1mm]

  \raisebox{10mm}{(d)} &
  \includegraphics[width=.133\hsize]{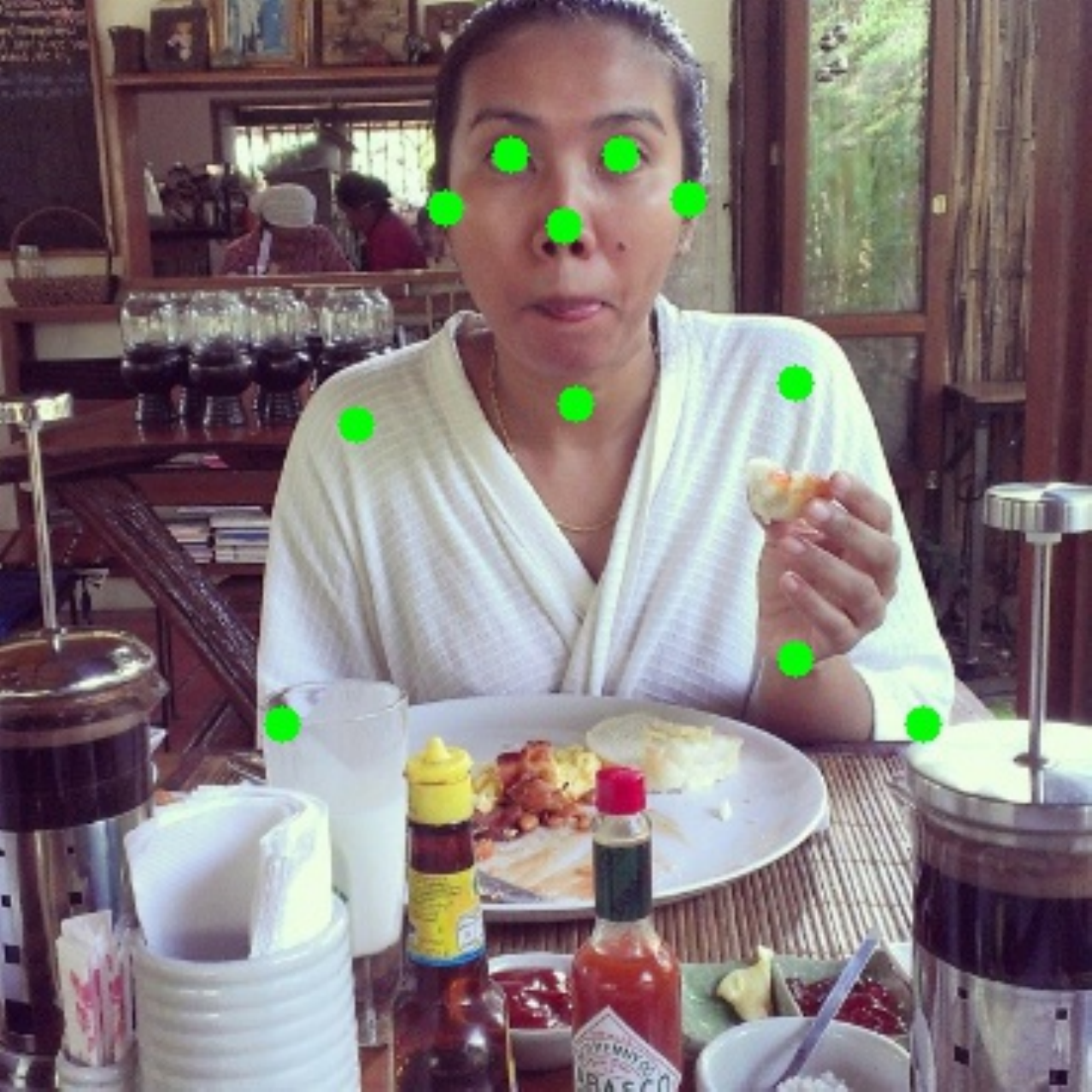} &
  \includegraphics[width=.133\hsize]{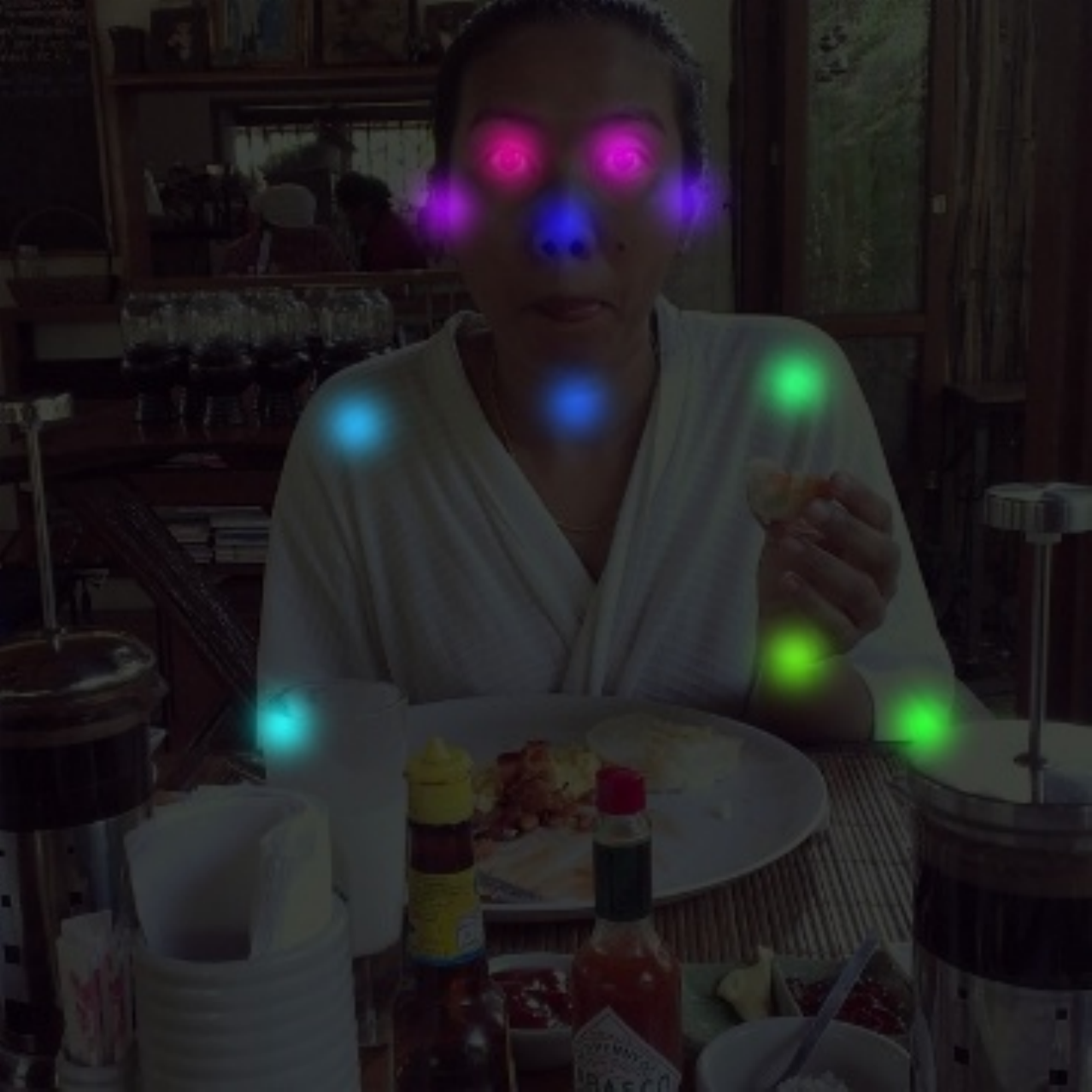} &
  \includegraphics[width=.133\hsize]{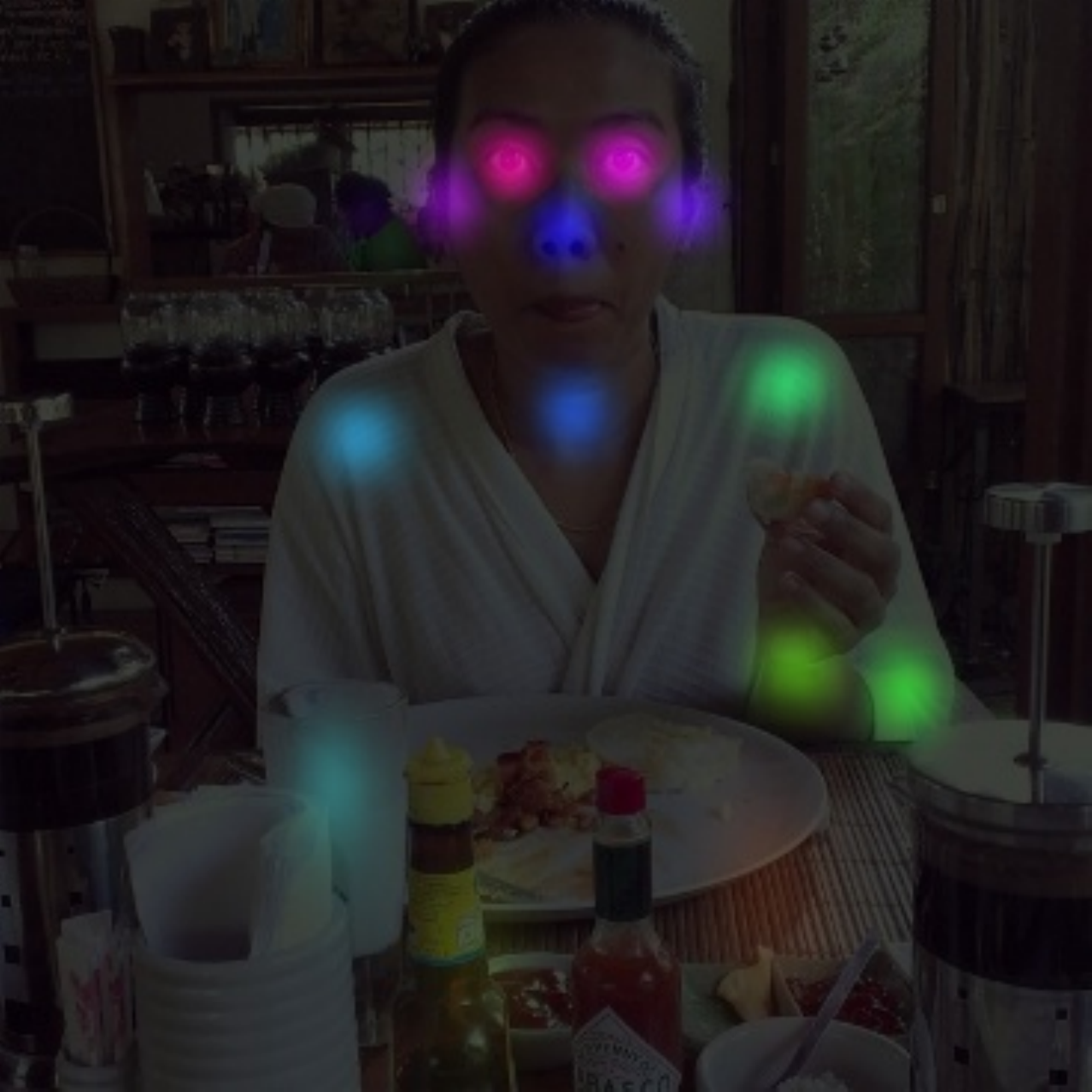} &
  \includegraphics[width=.133\hsize]{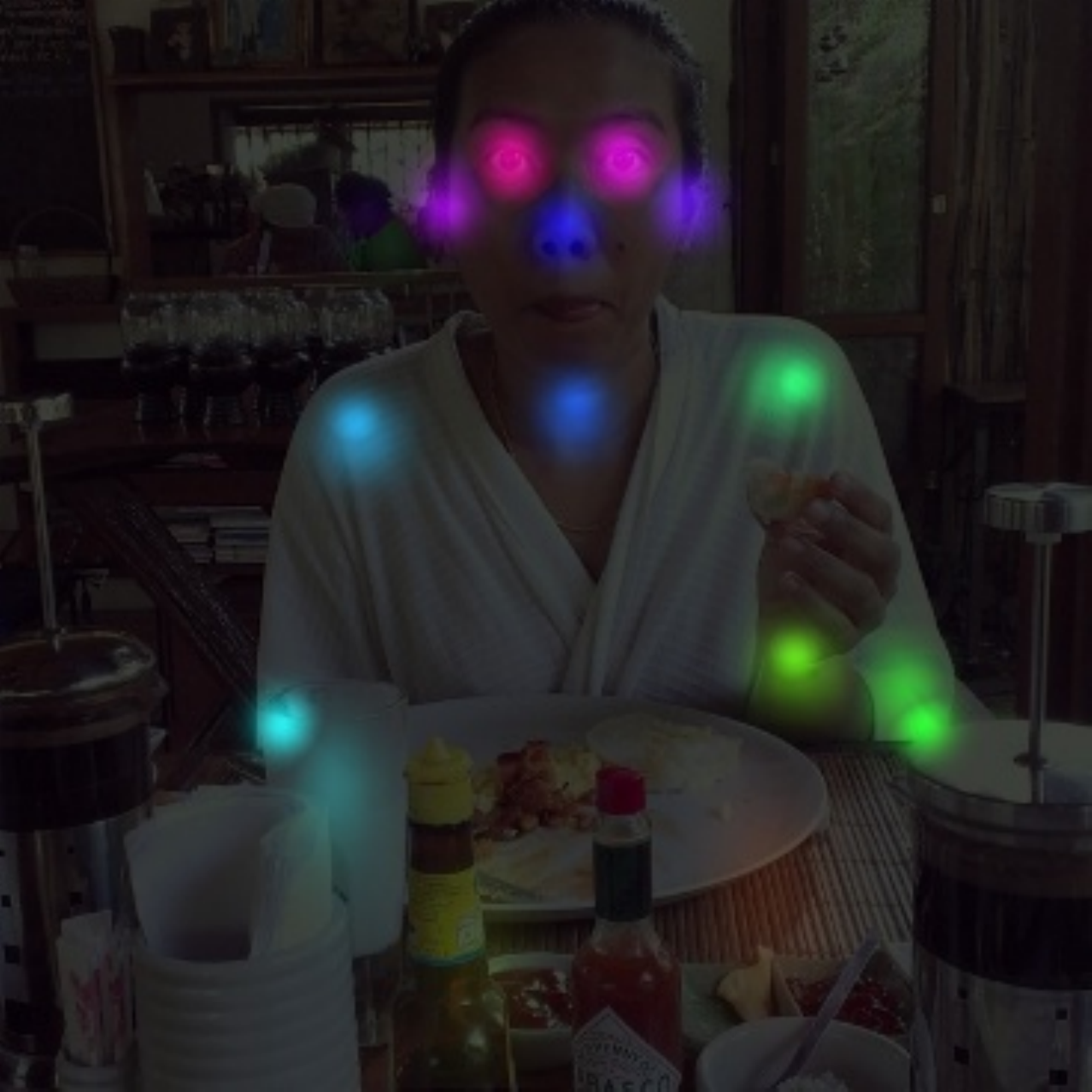} &
  \includegraphics[width=.133\hsize]{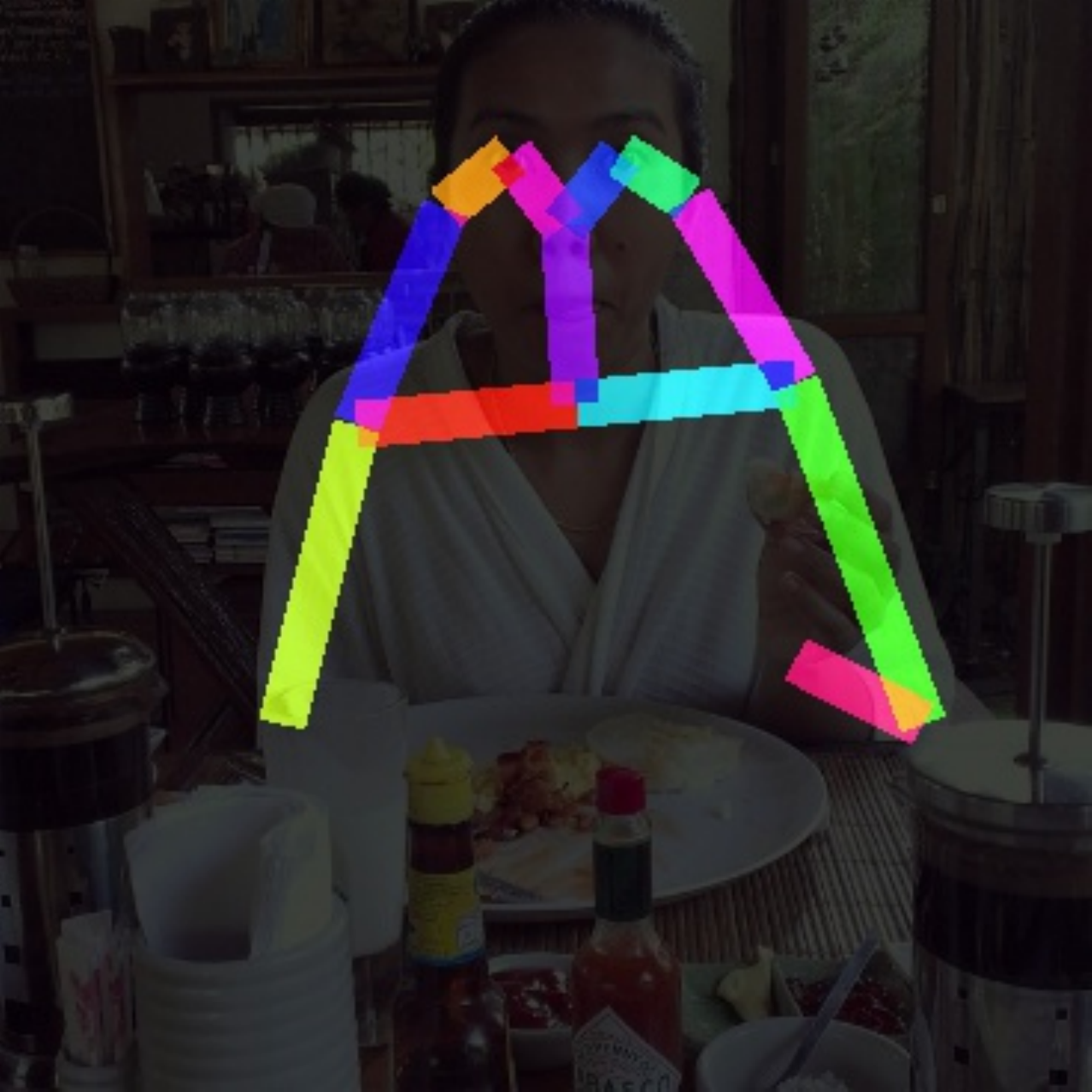} &
  \includegraphics[width=.133\hsize]{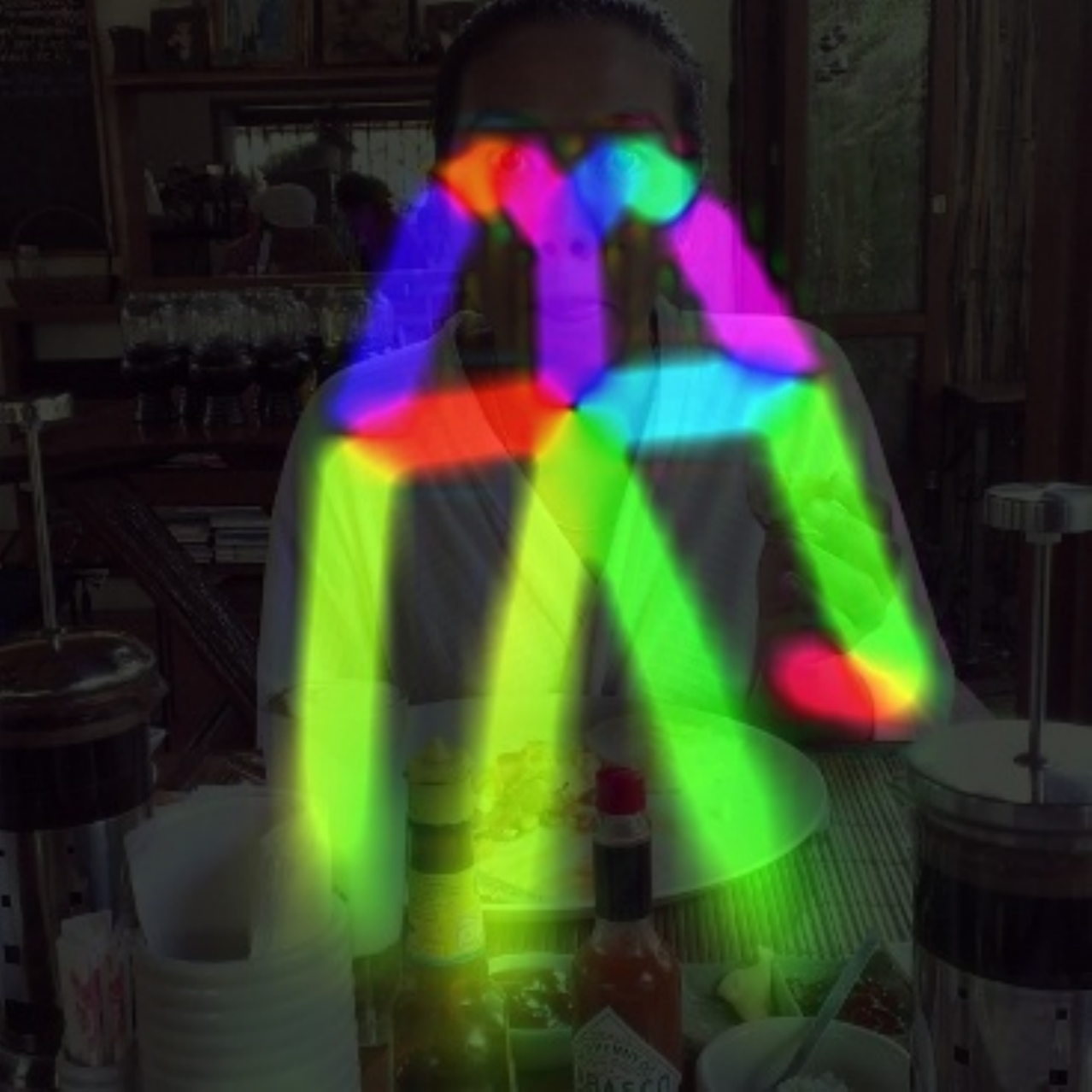} &
  \includegraphics[width=.133\hsize]{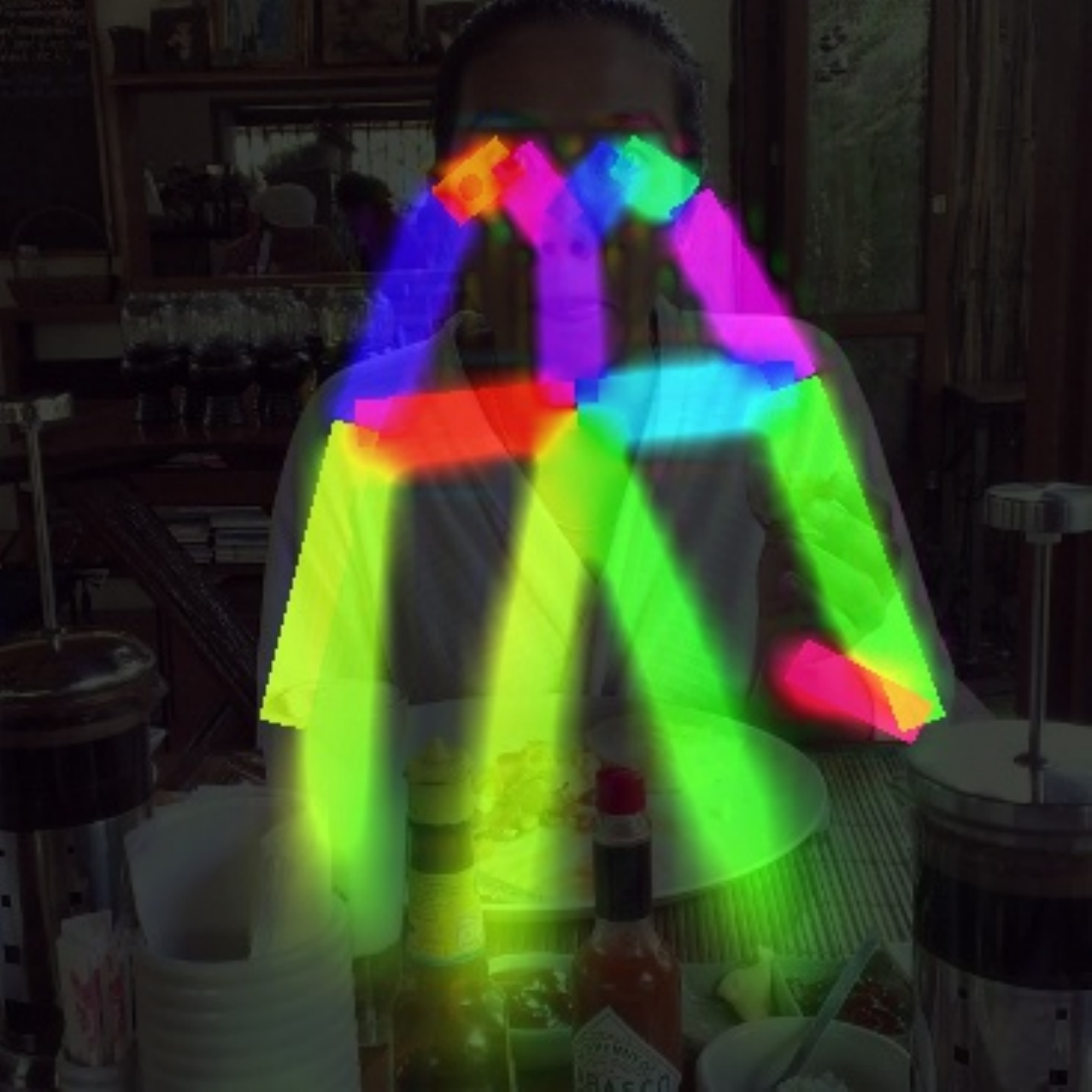} \\

  \raisebox{11mm}{(e)} &
  \includegraphics[width=.133\hsize]{result/00292024_gt_joints} &
  \includegraphics[width=.133\hsize]{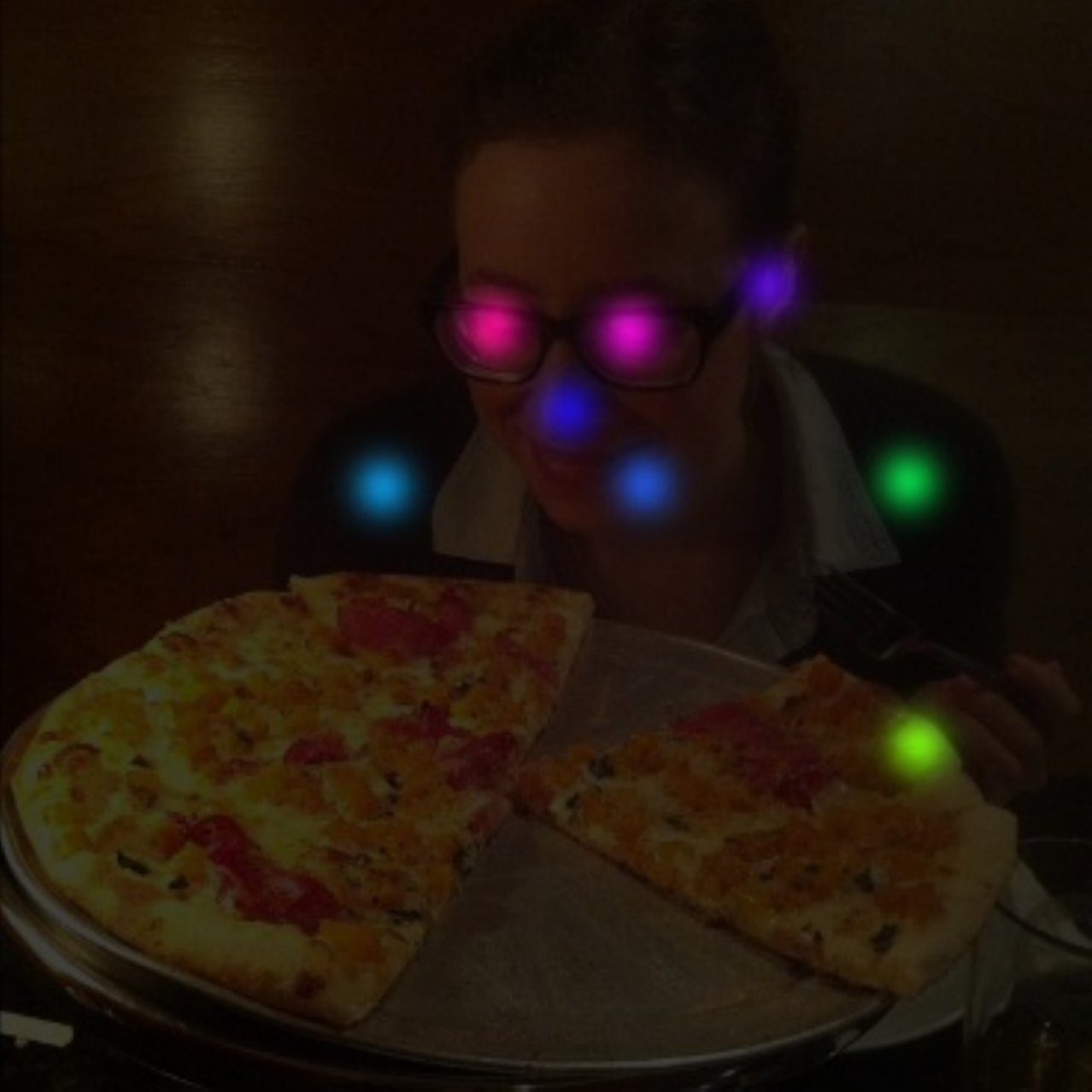} &
  \includegraphics[width=.133\hsize]{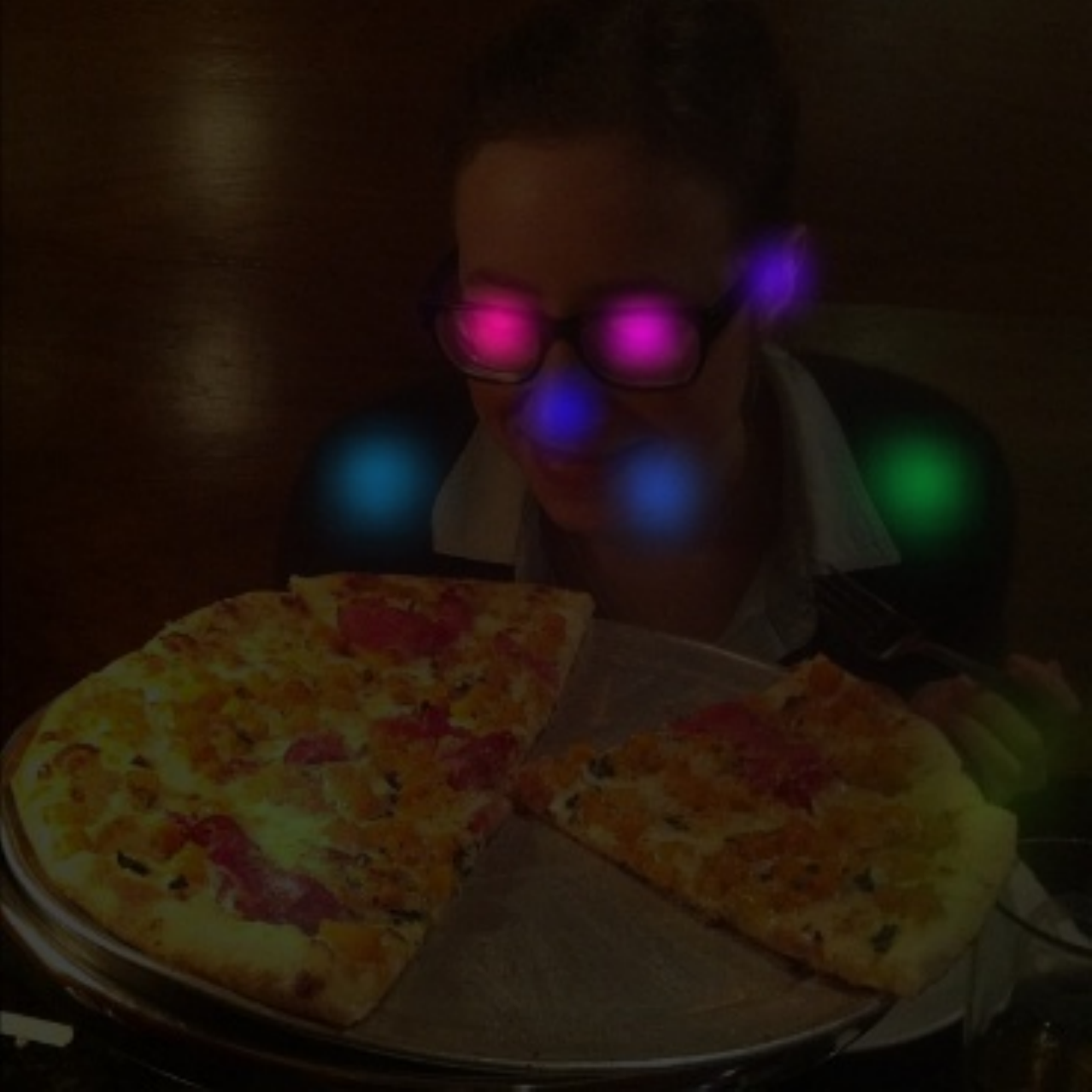} &
  \includegraphics[width=.133\hsize]{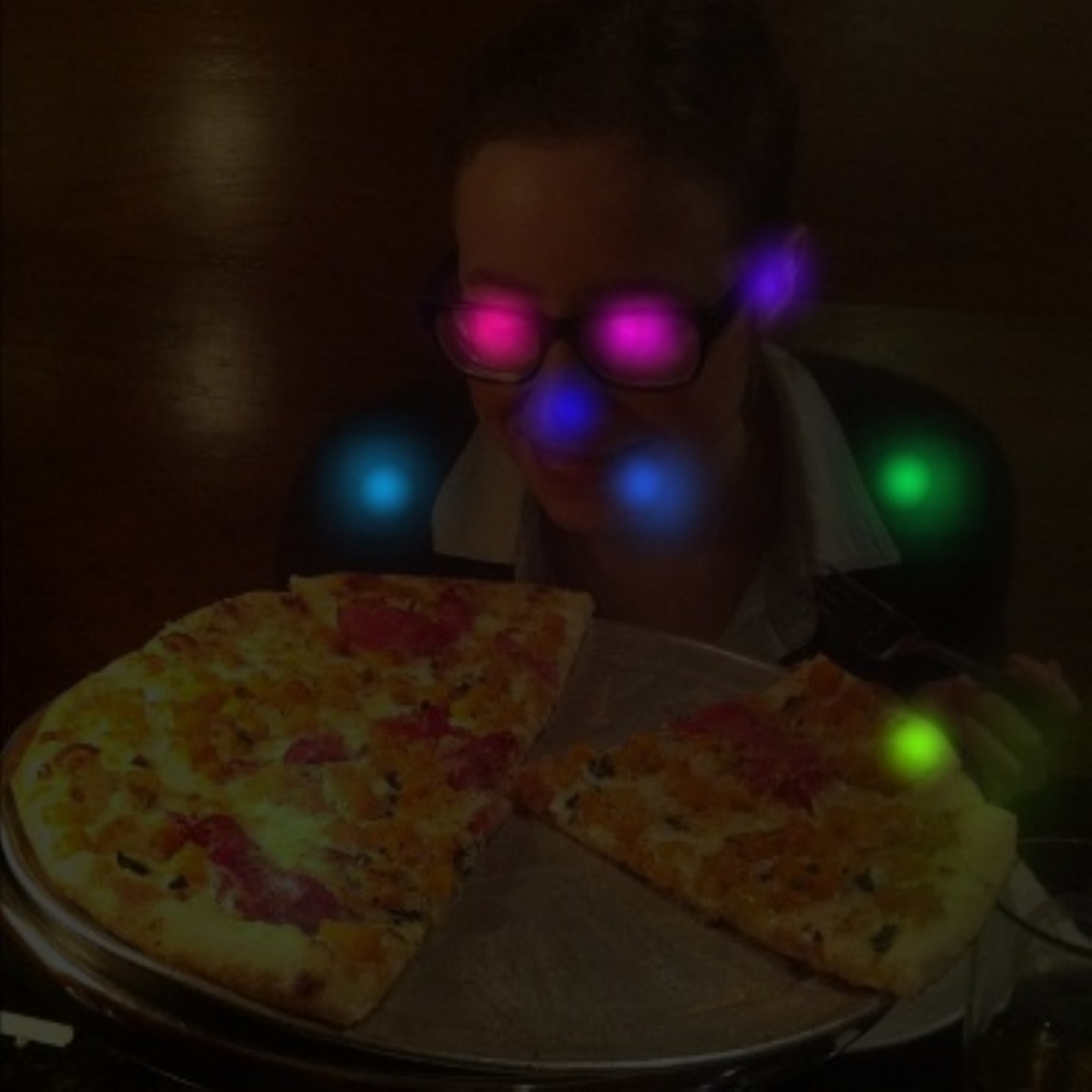} &
  \includegraphics[width=.133\hsize]{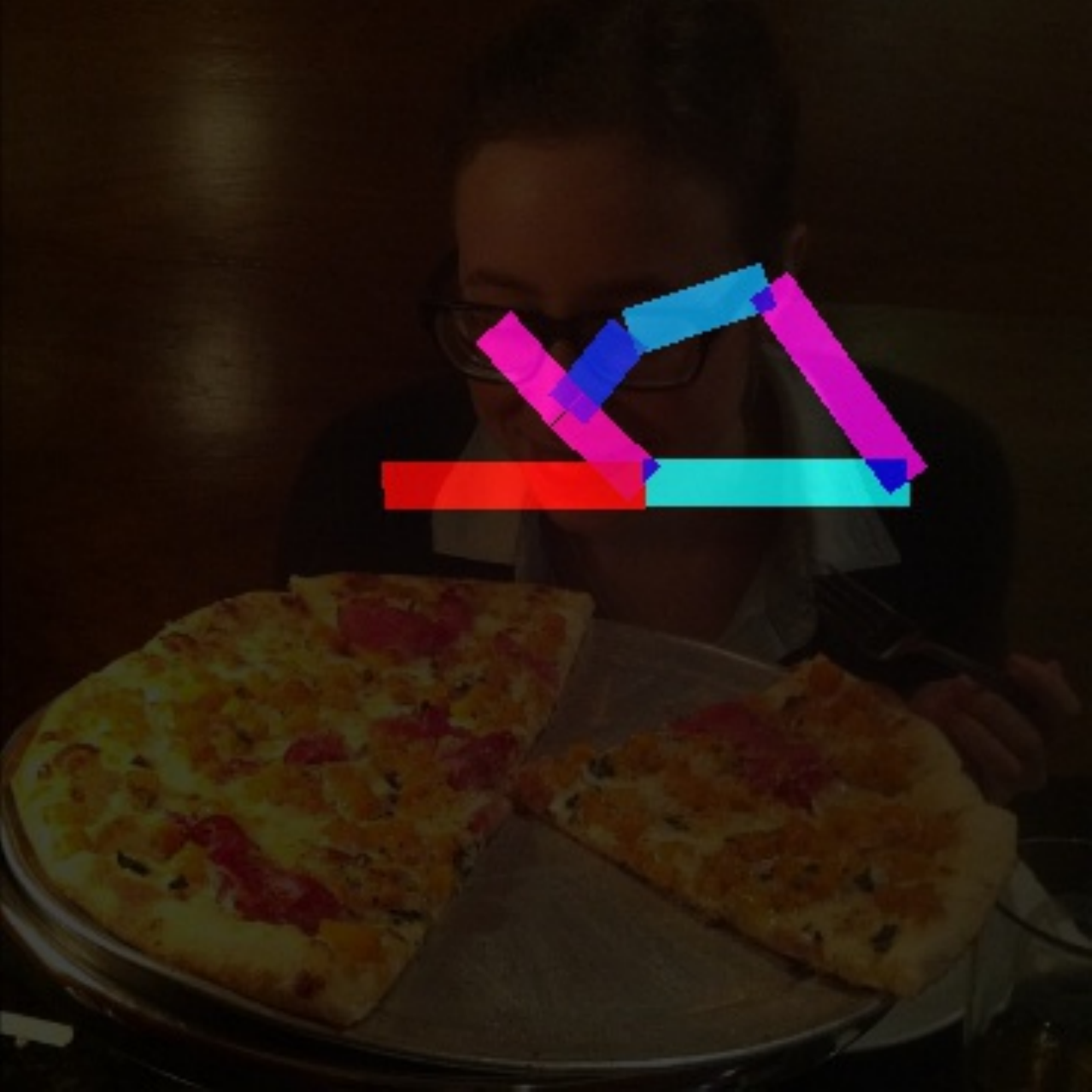} &
  \includegraphics[width=.133\hsize]{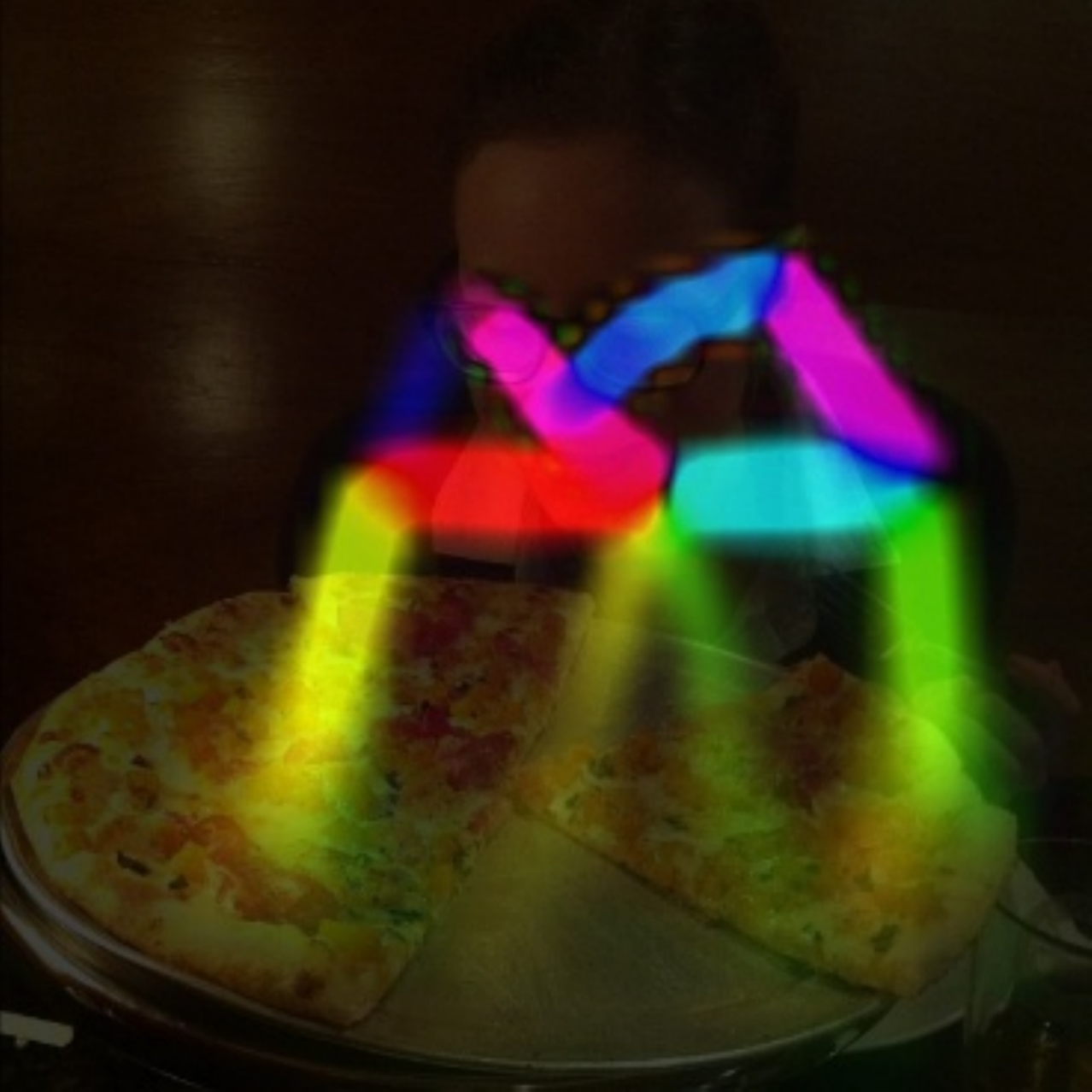} &
  \includegraphics[width=.133\hsize]{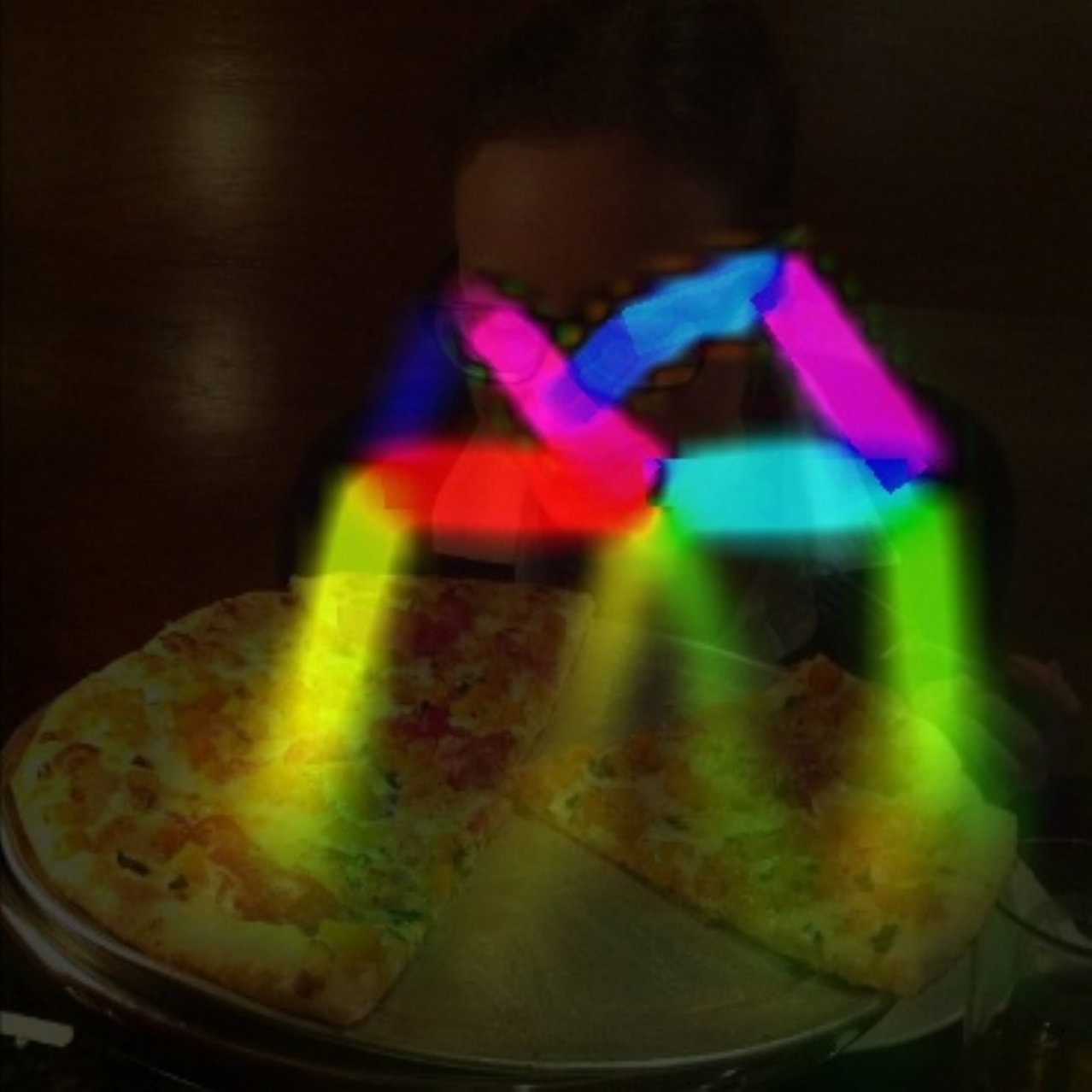} \\

  \addlinespace[1mm]

  \raisebox{12mm}{(f)} &
  \includegraphics[width=.133\hsize]{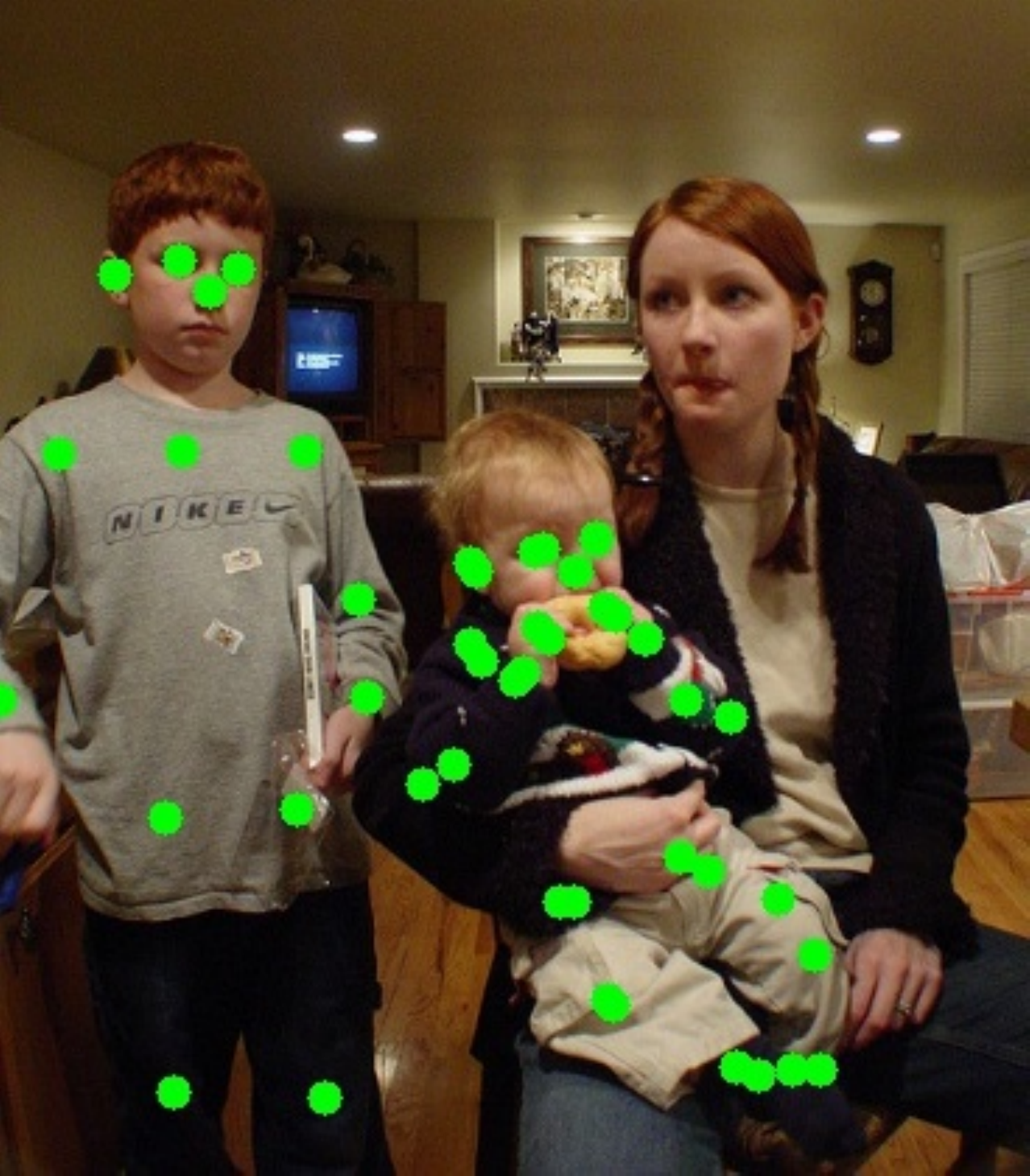} &
  \includegraphics[width=.133\hsize]{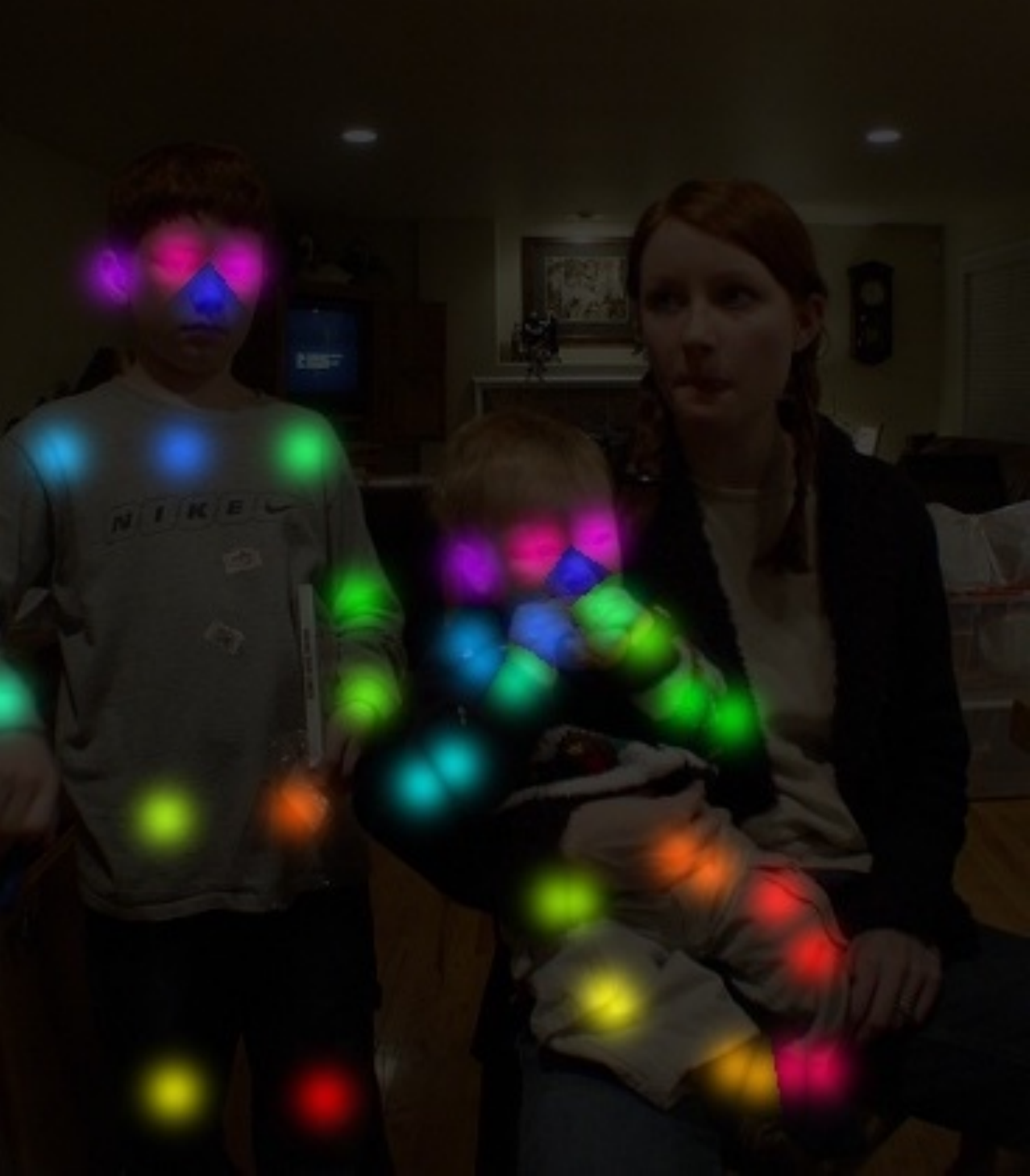} &
  \includegraphics[width=.133\hsize]{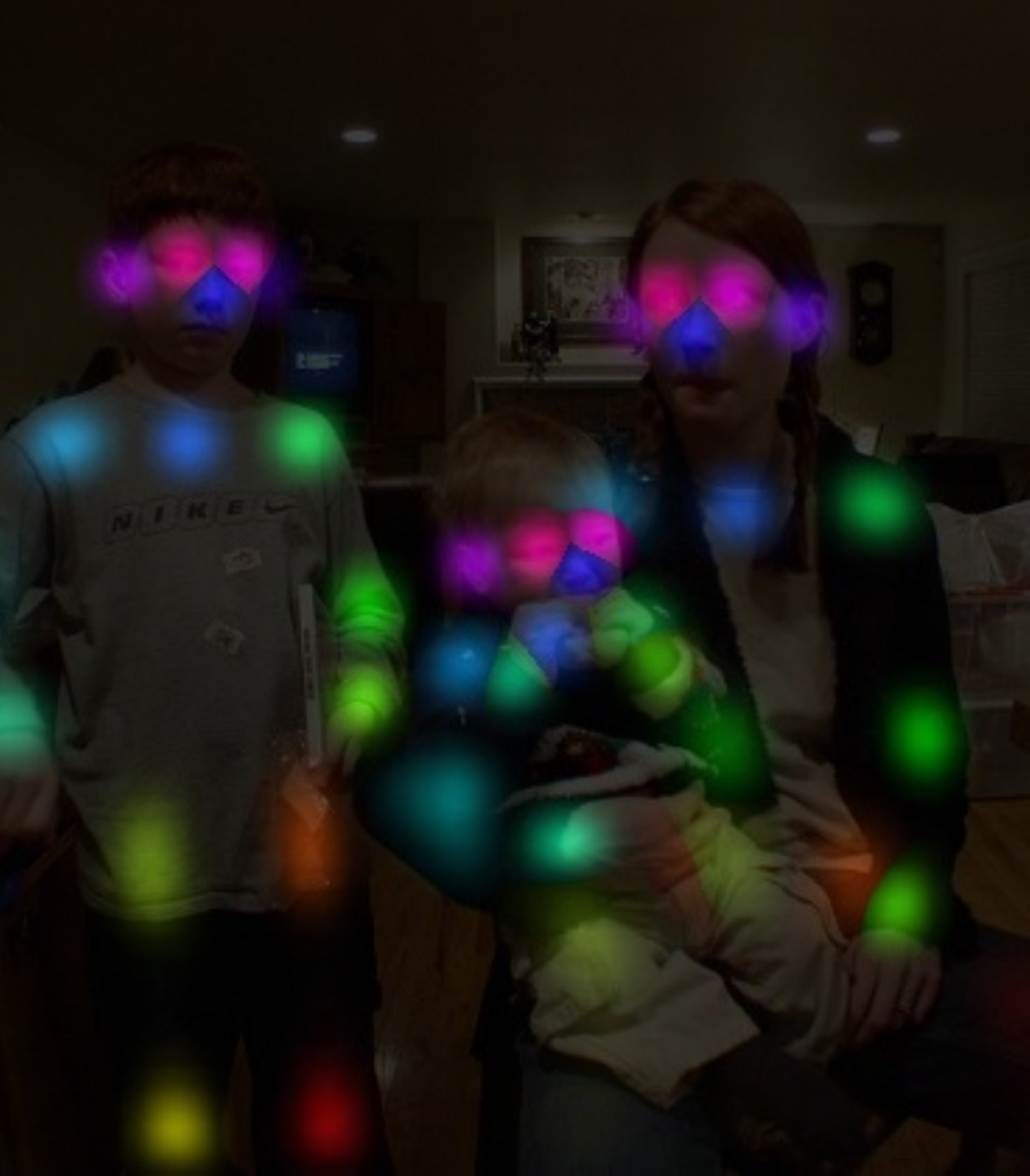} &
  \includegraphics[width=.133\hsize]{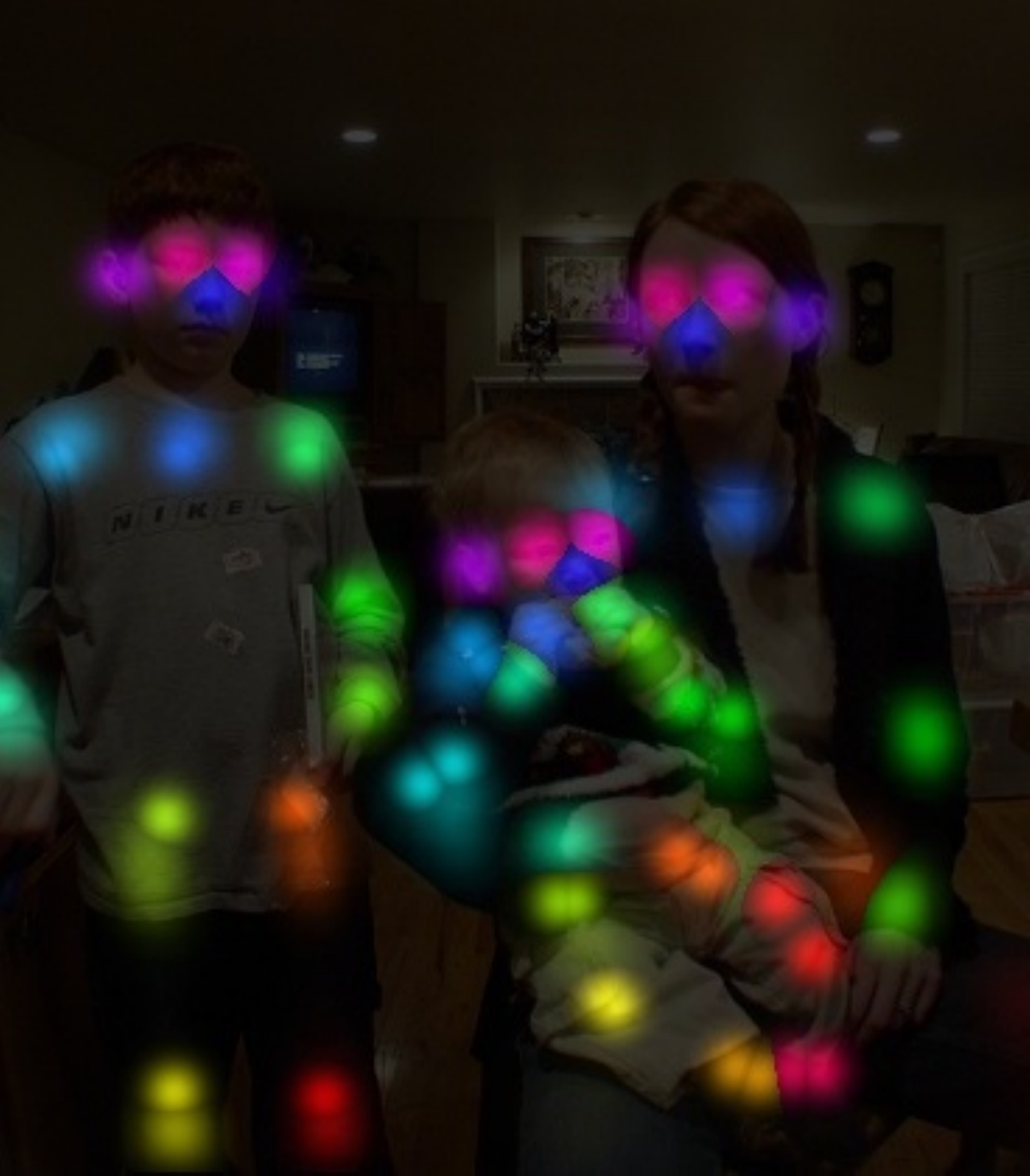} &
  \includegraphics[width=.133\hsize]{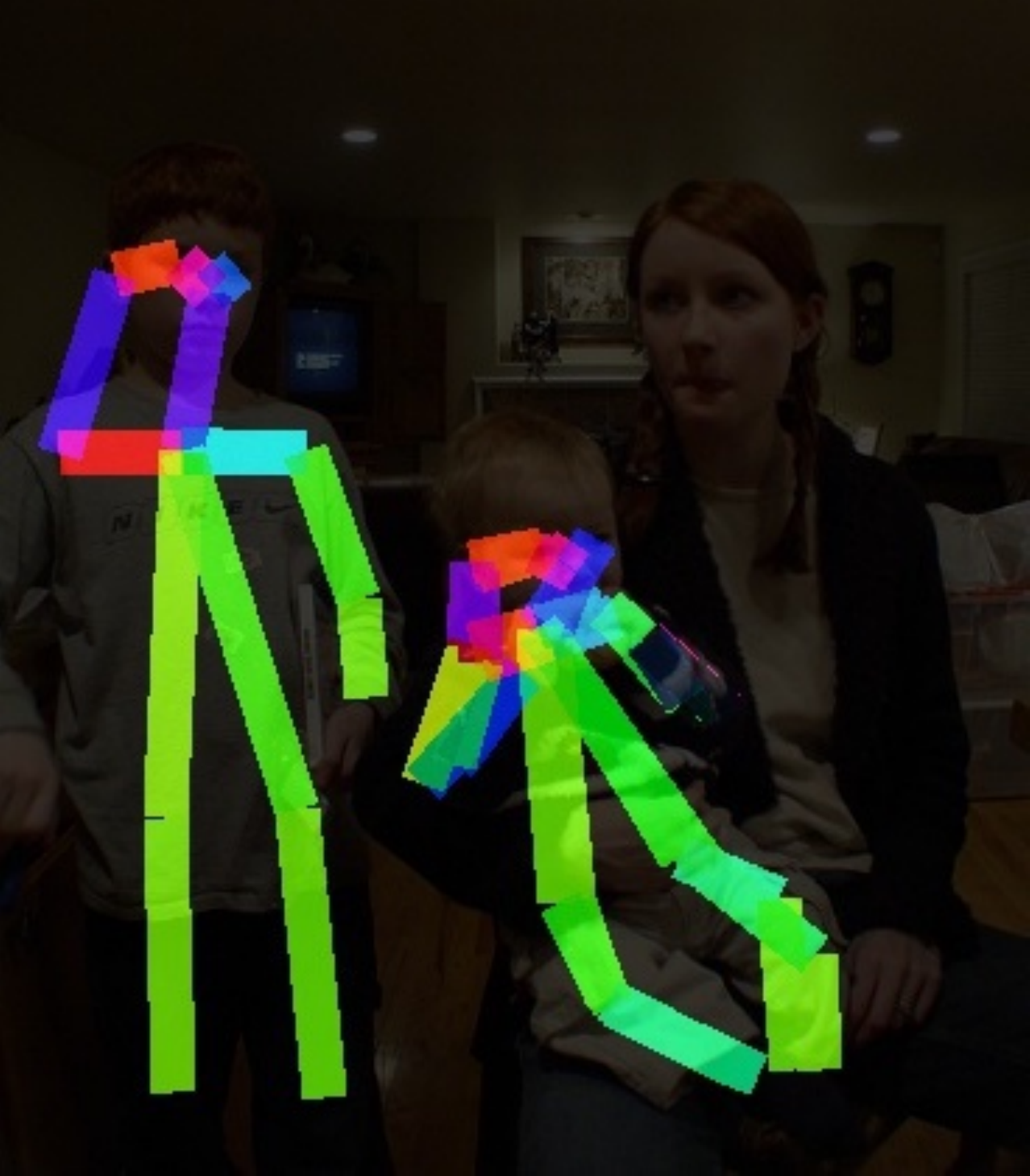} &
  \includegraphics[width=.133\hsize]{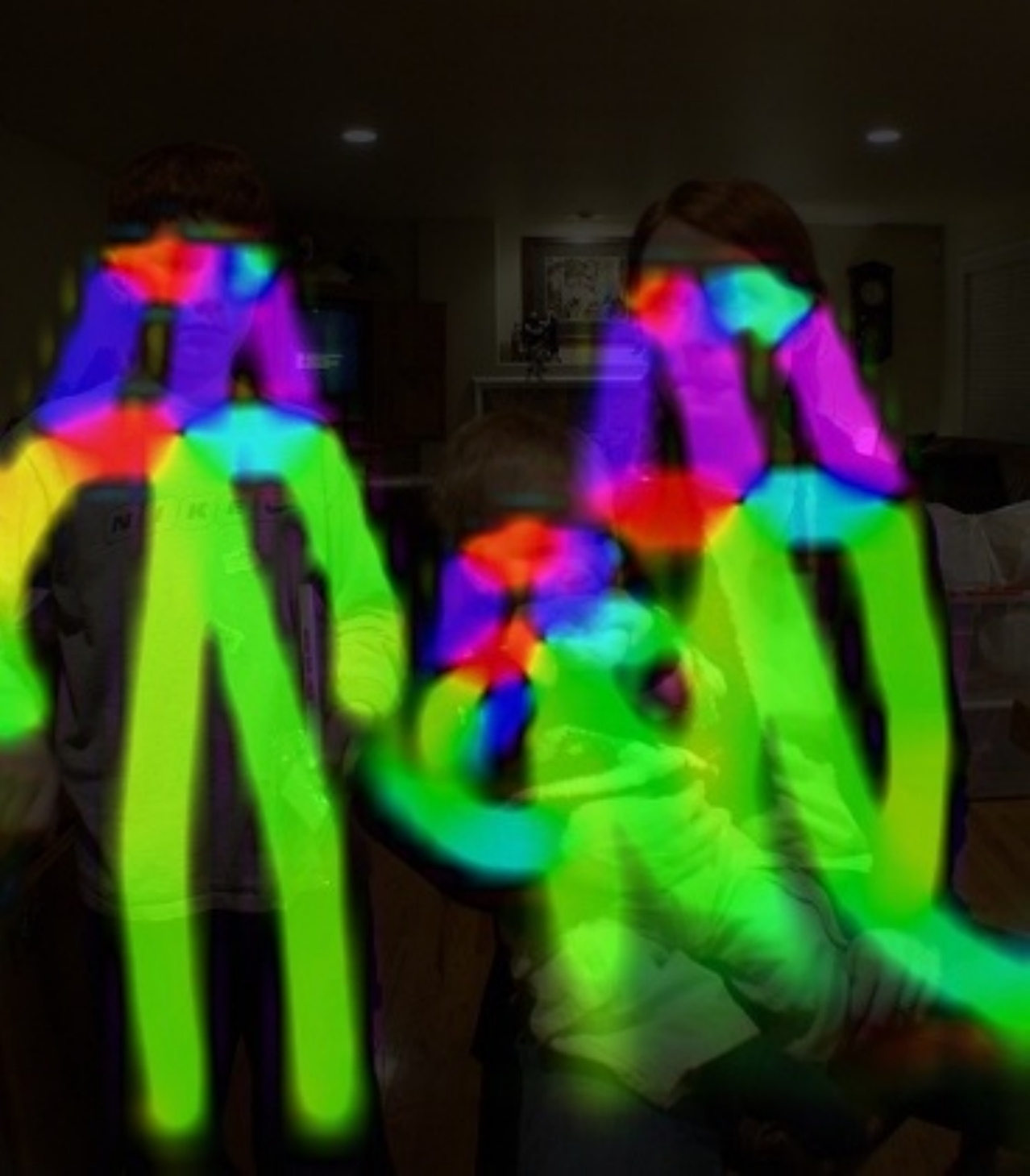} &
  \includegraphics[width=.133\hsize]{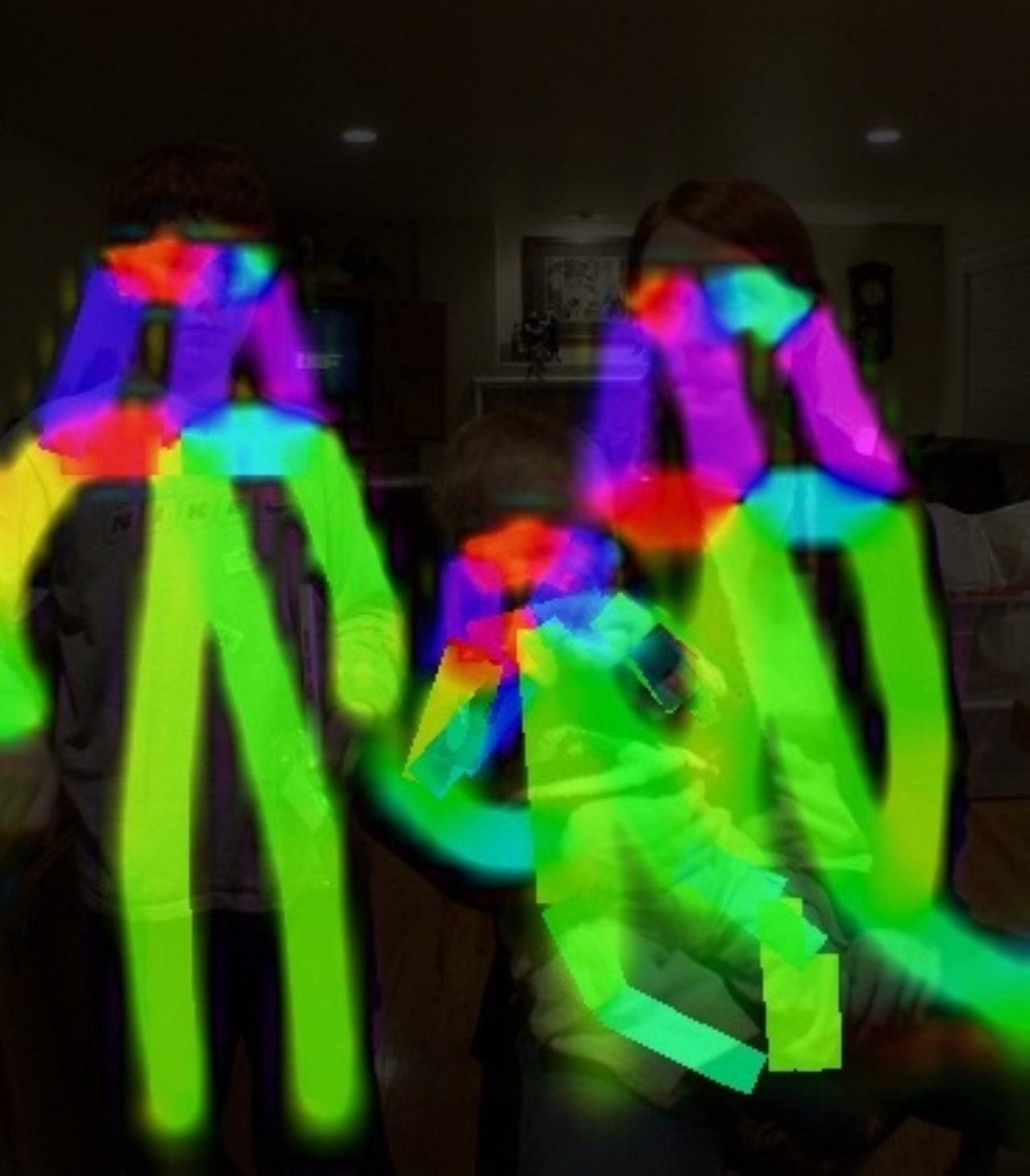} \\

  \addlinespace[1mm]

  \raisebox{8mm}{(g)} &
  \includegraphics[width=.133\hsize]{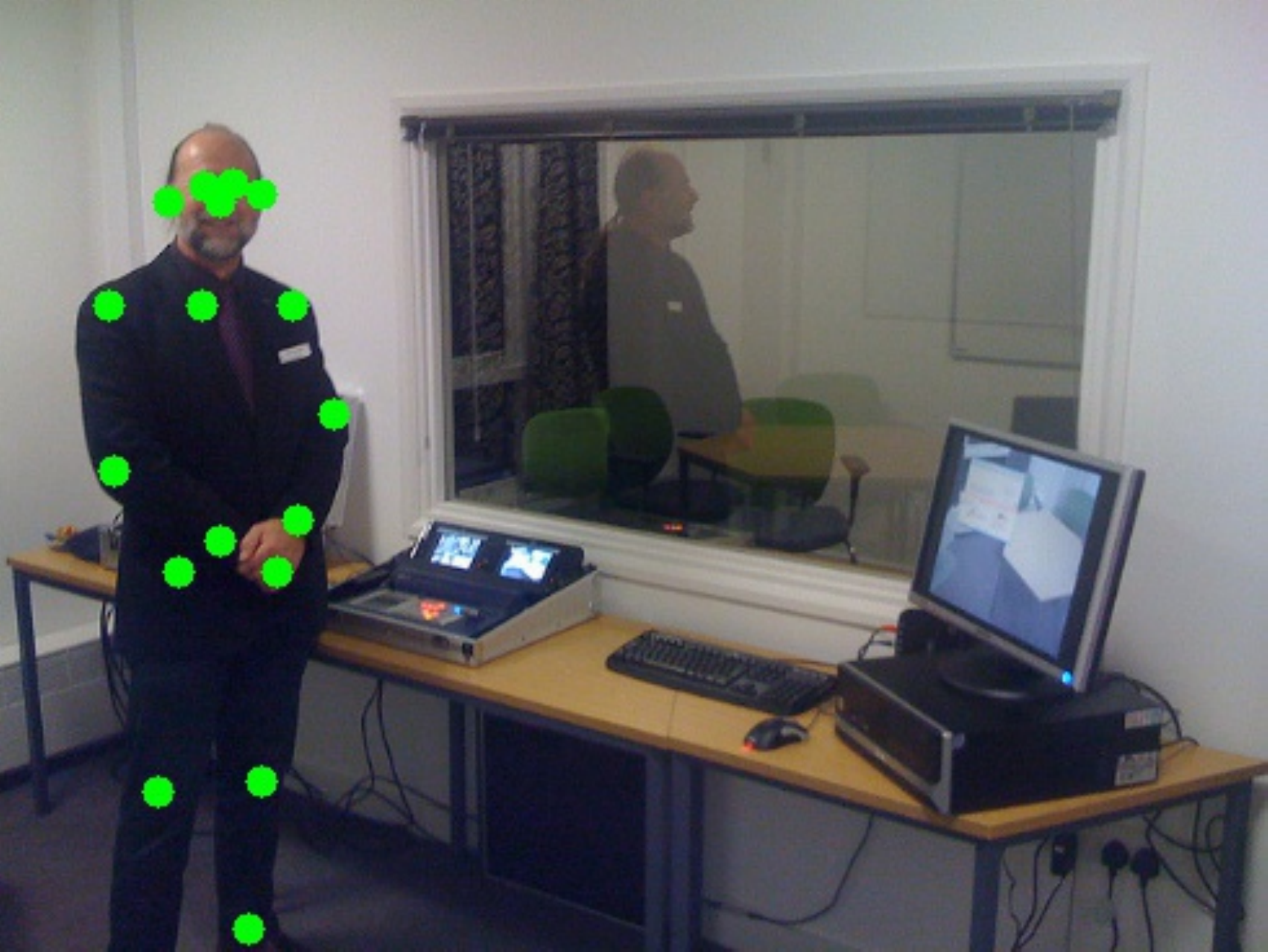} &
  \includegraphics[width=.133\hsize]{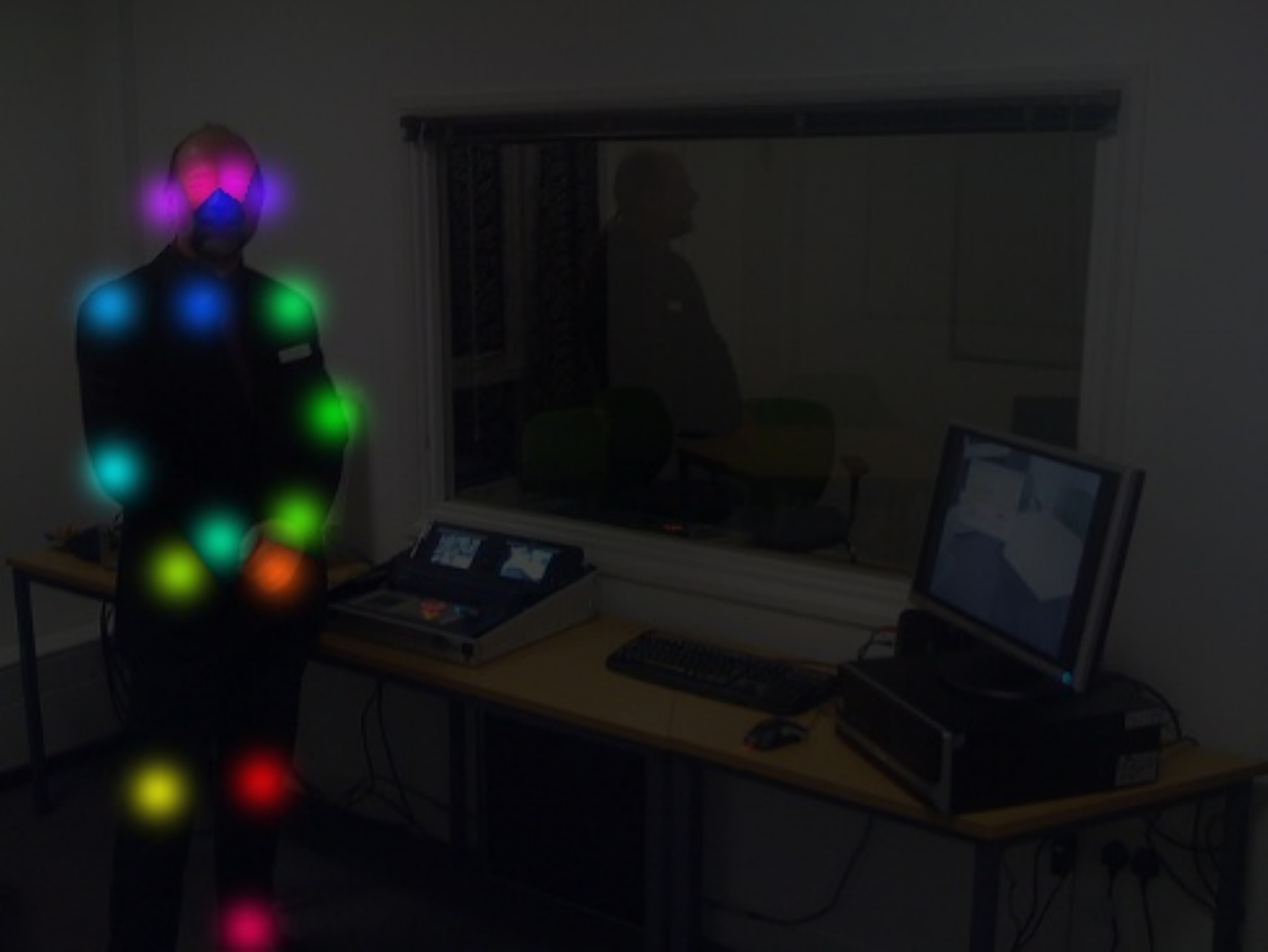} &
  \includegraphics[width=.133\hsize]{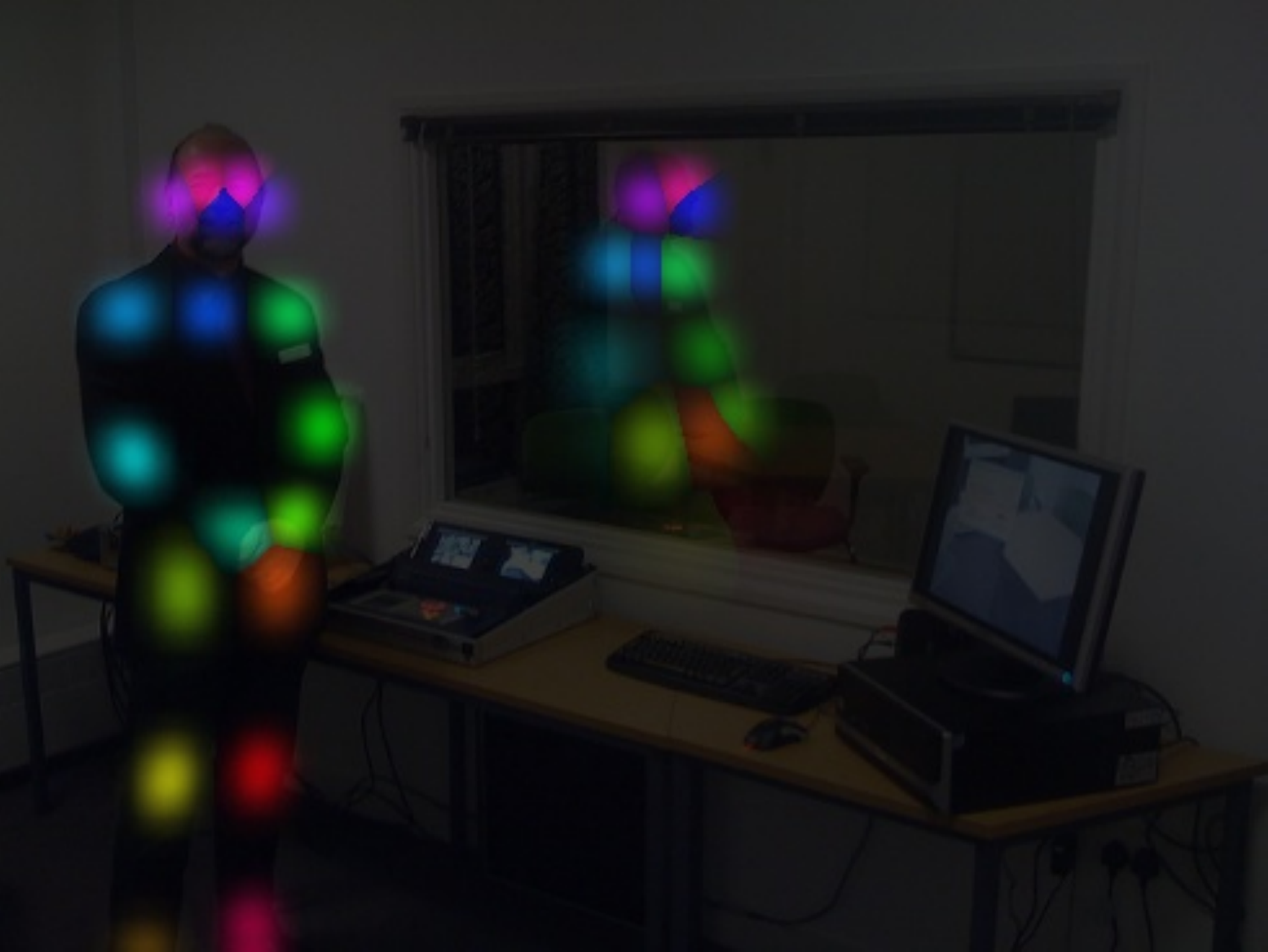} &
  \includegraphics[width=.133\hsize]{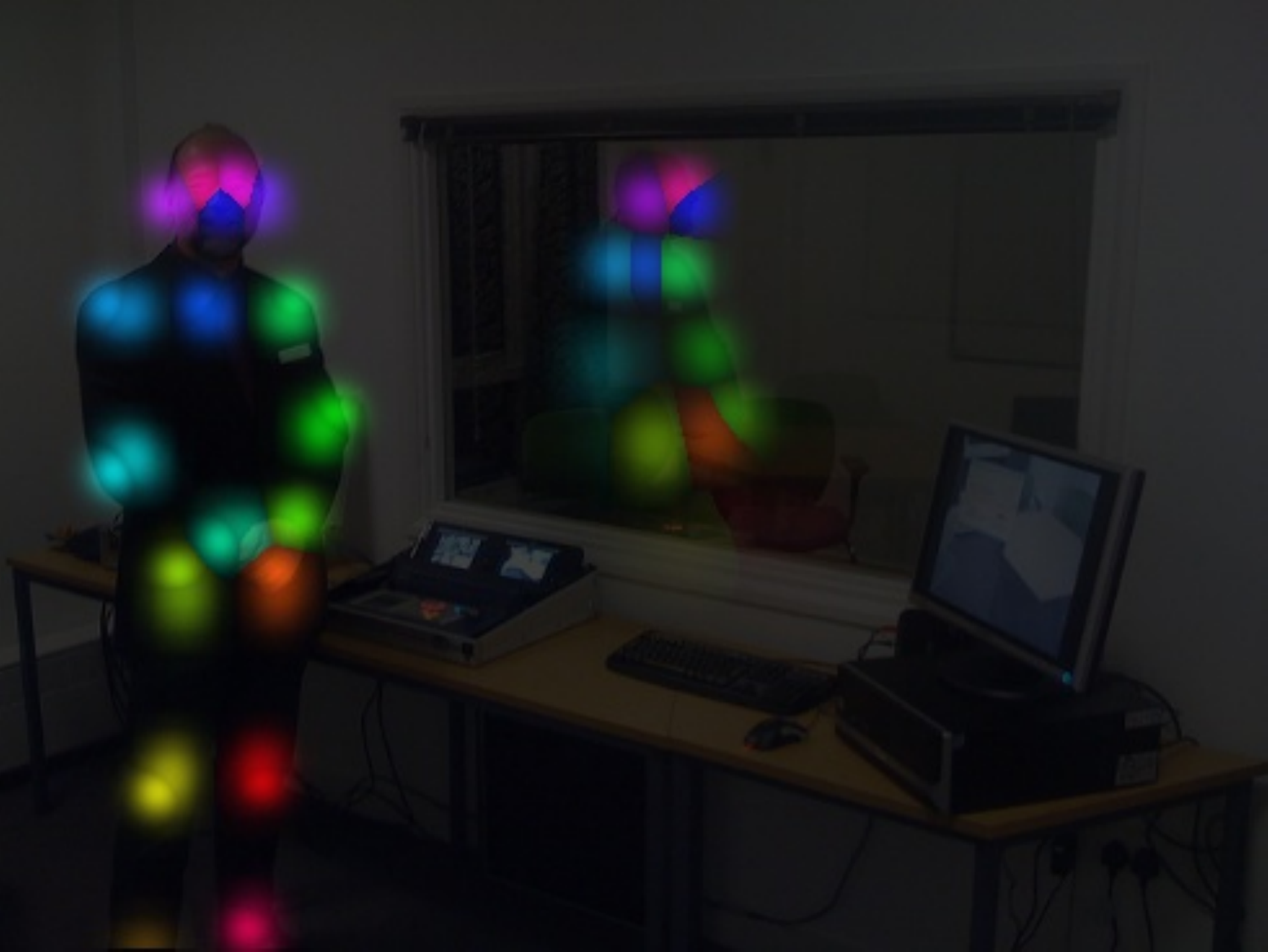} &
  \includegraphics[width=.133\hsize]{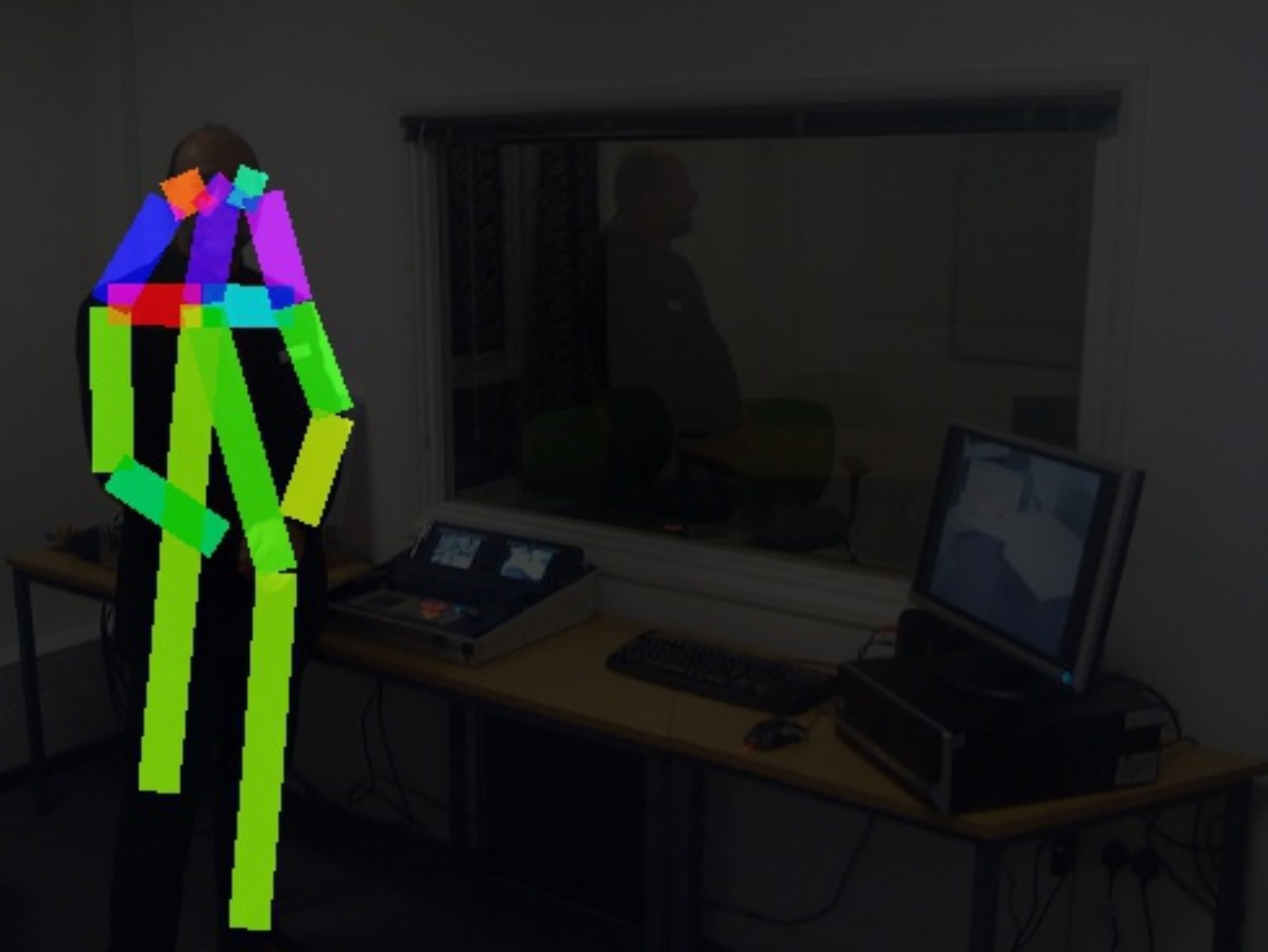} &
  \includegraphics[width=.133\hsize]{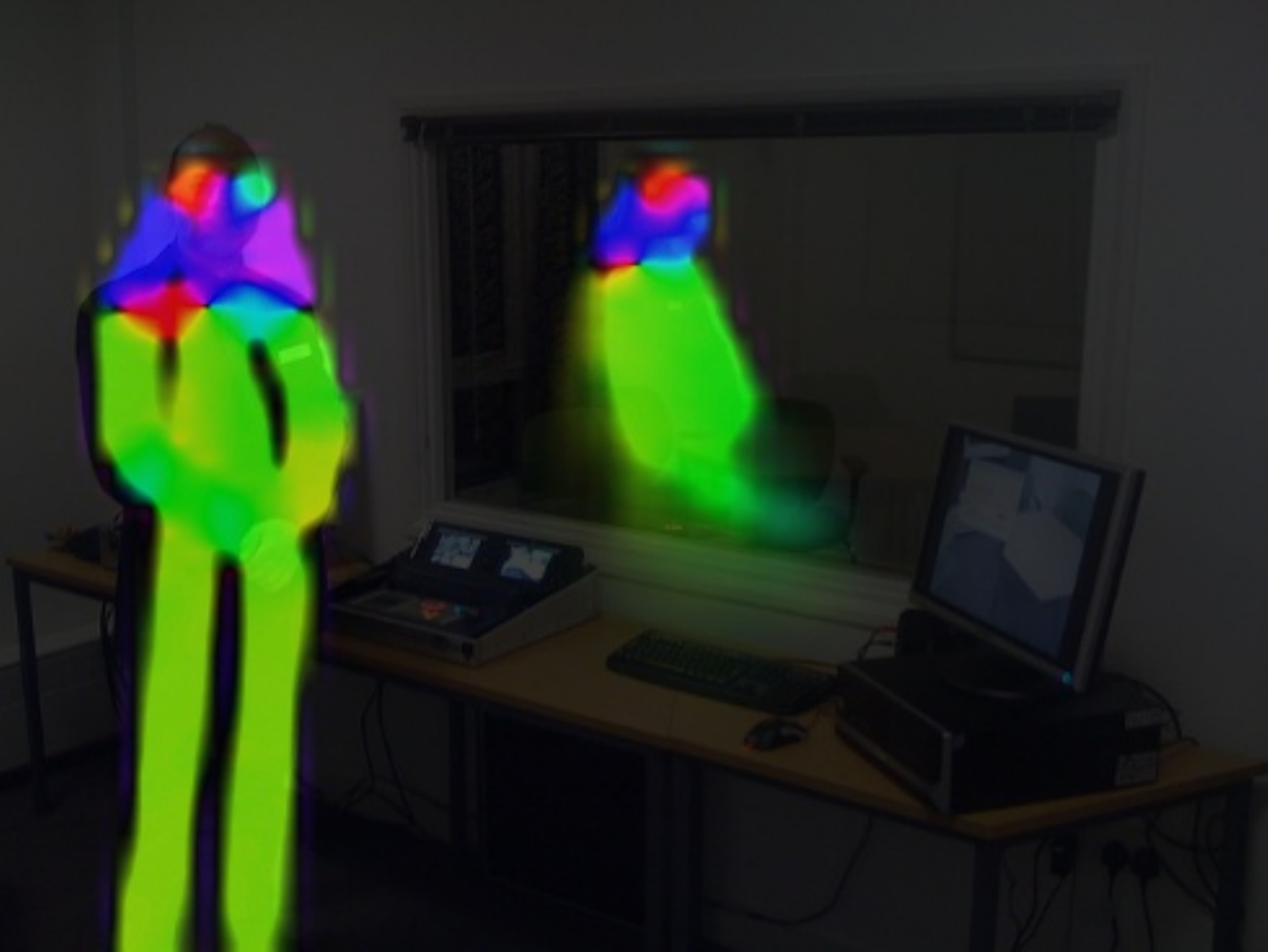} &
  \includegraphics[width=.133\hsize]{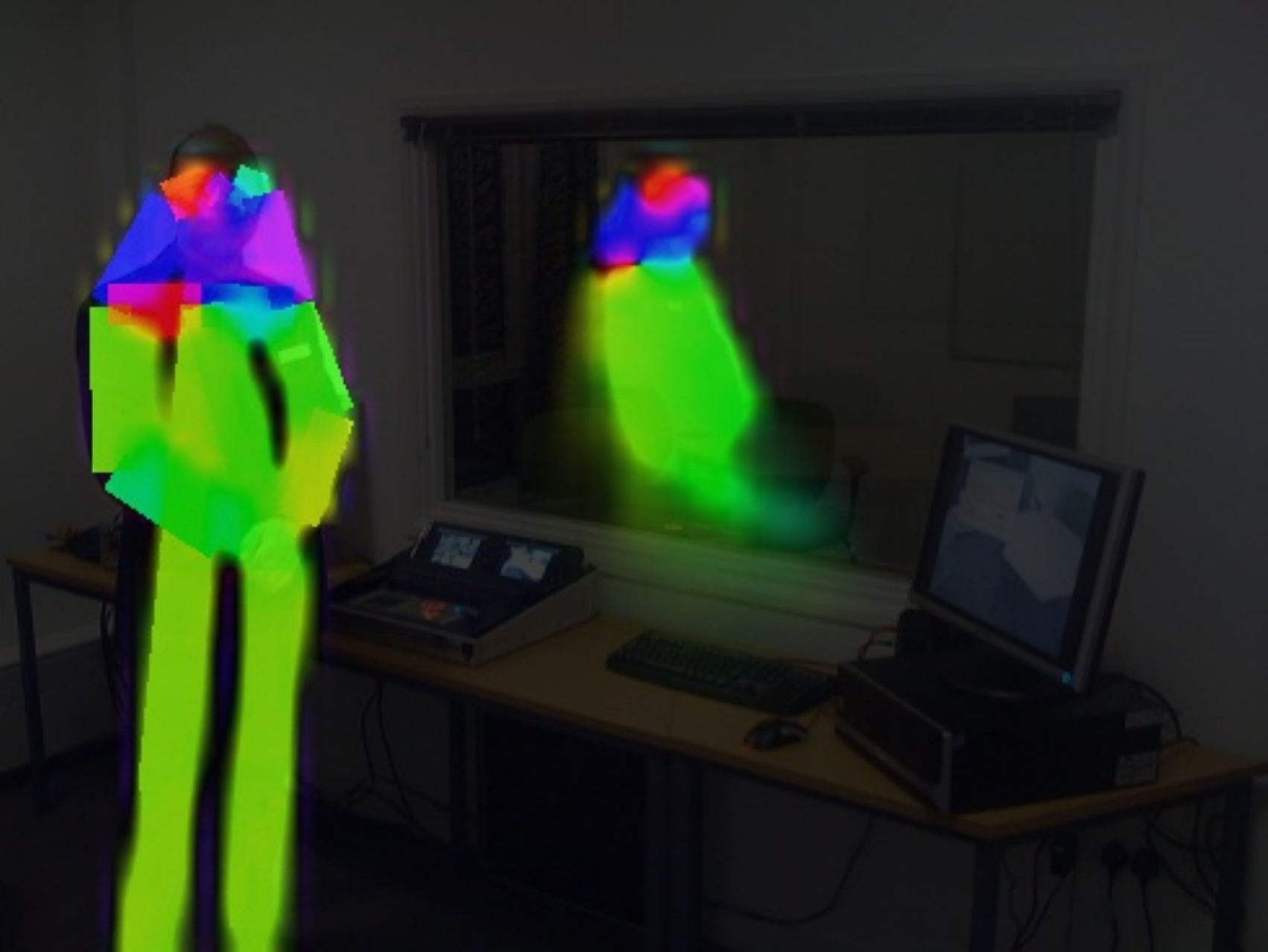} \\

  \raisebox{8mm}{(h)} &
  \includegraphics[width=.133\hsize]{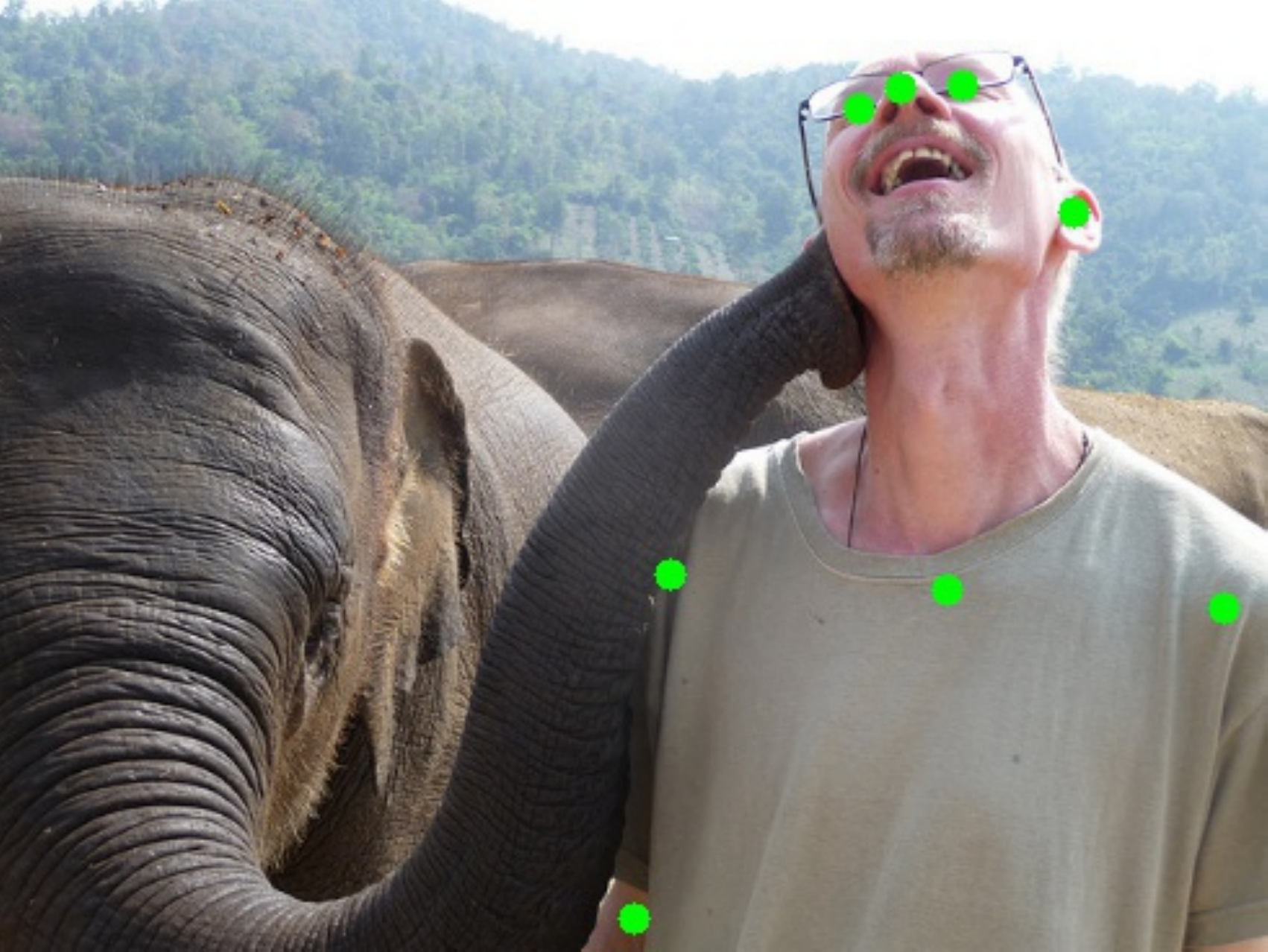} &
  \includegraphics[width=.133\hsize]{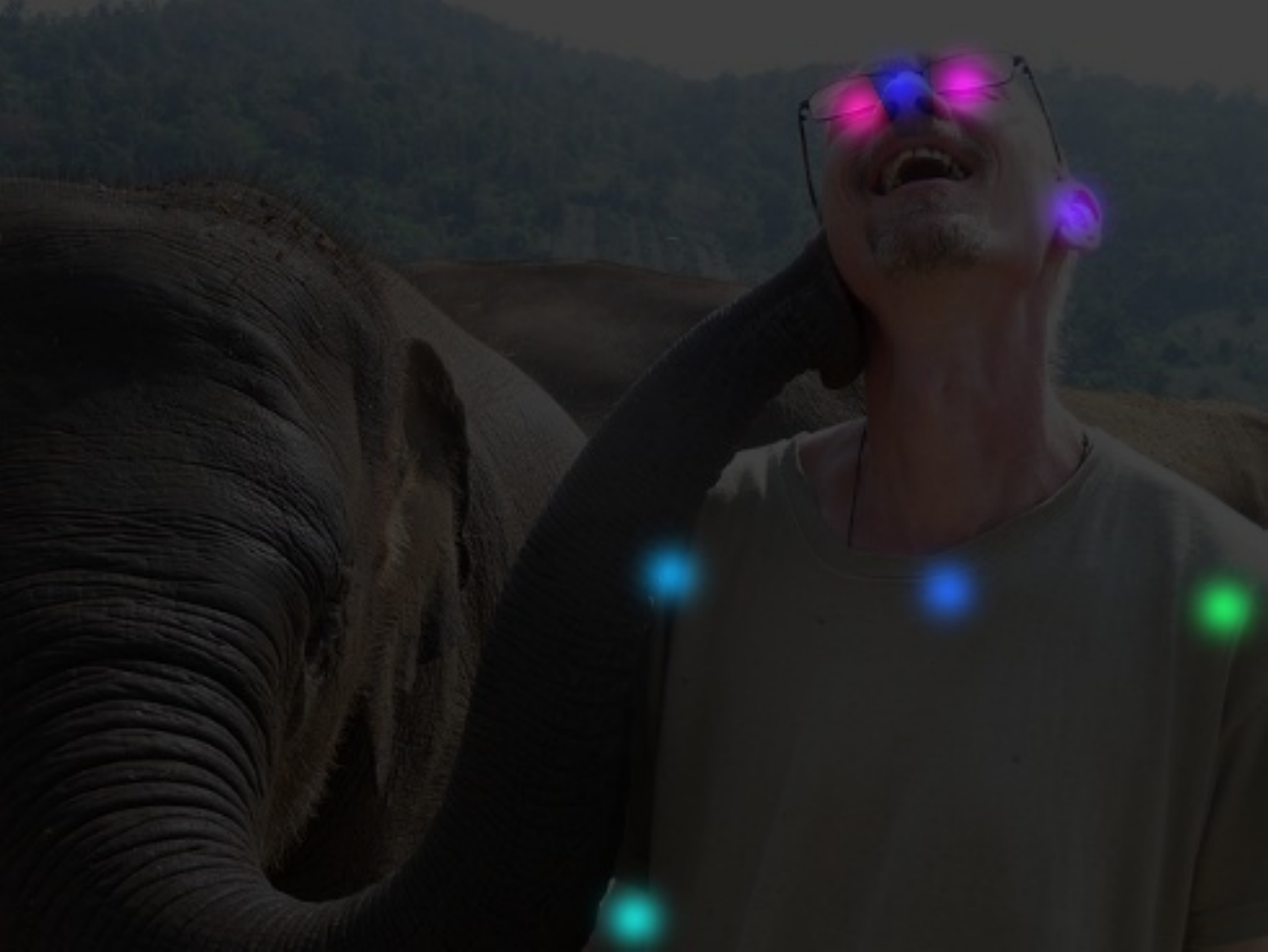} &
  \includegraphics[width=.133\hsize]{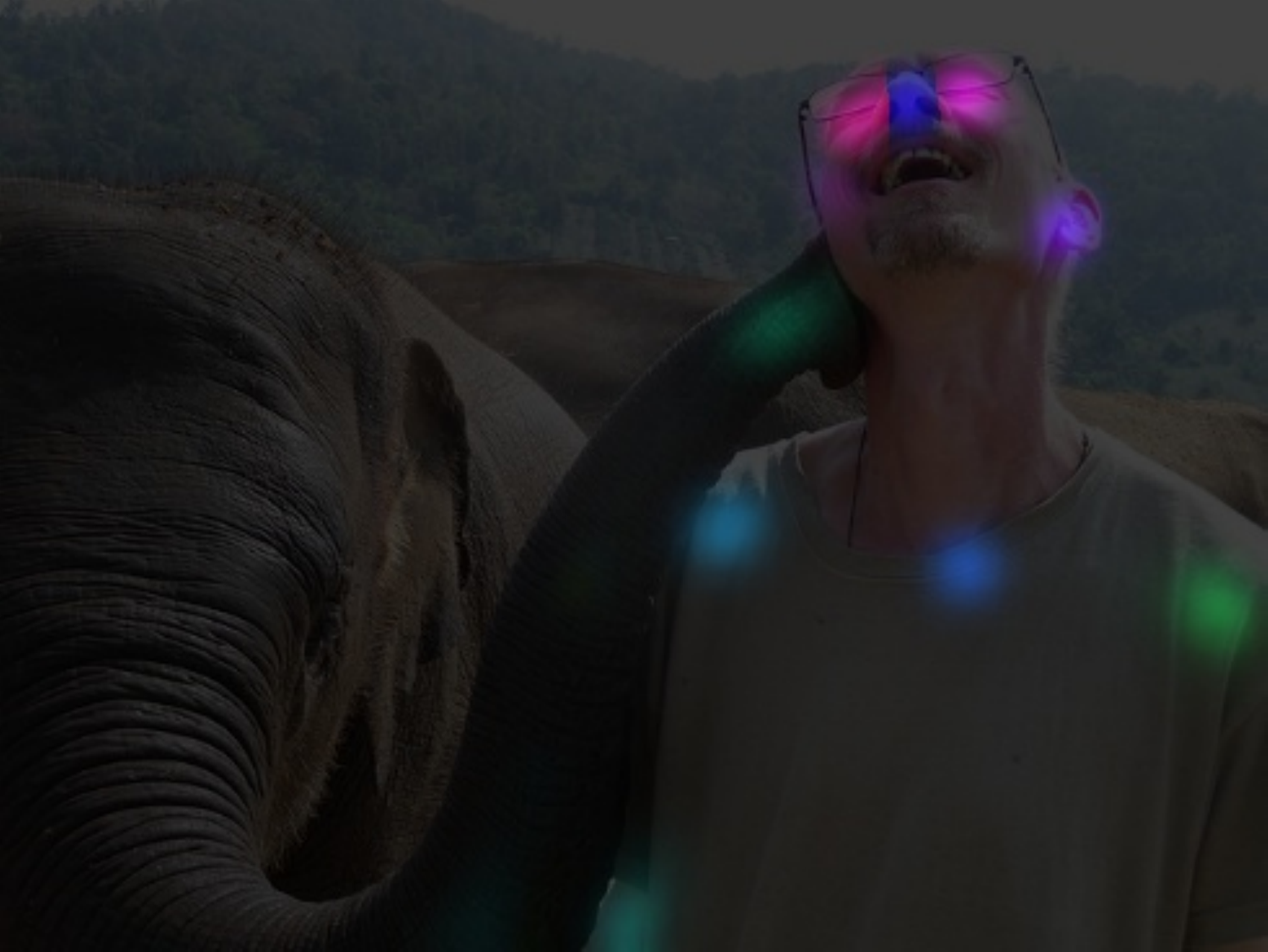} &
  \includegraphics[width=.133\hsize]{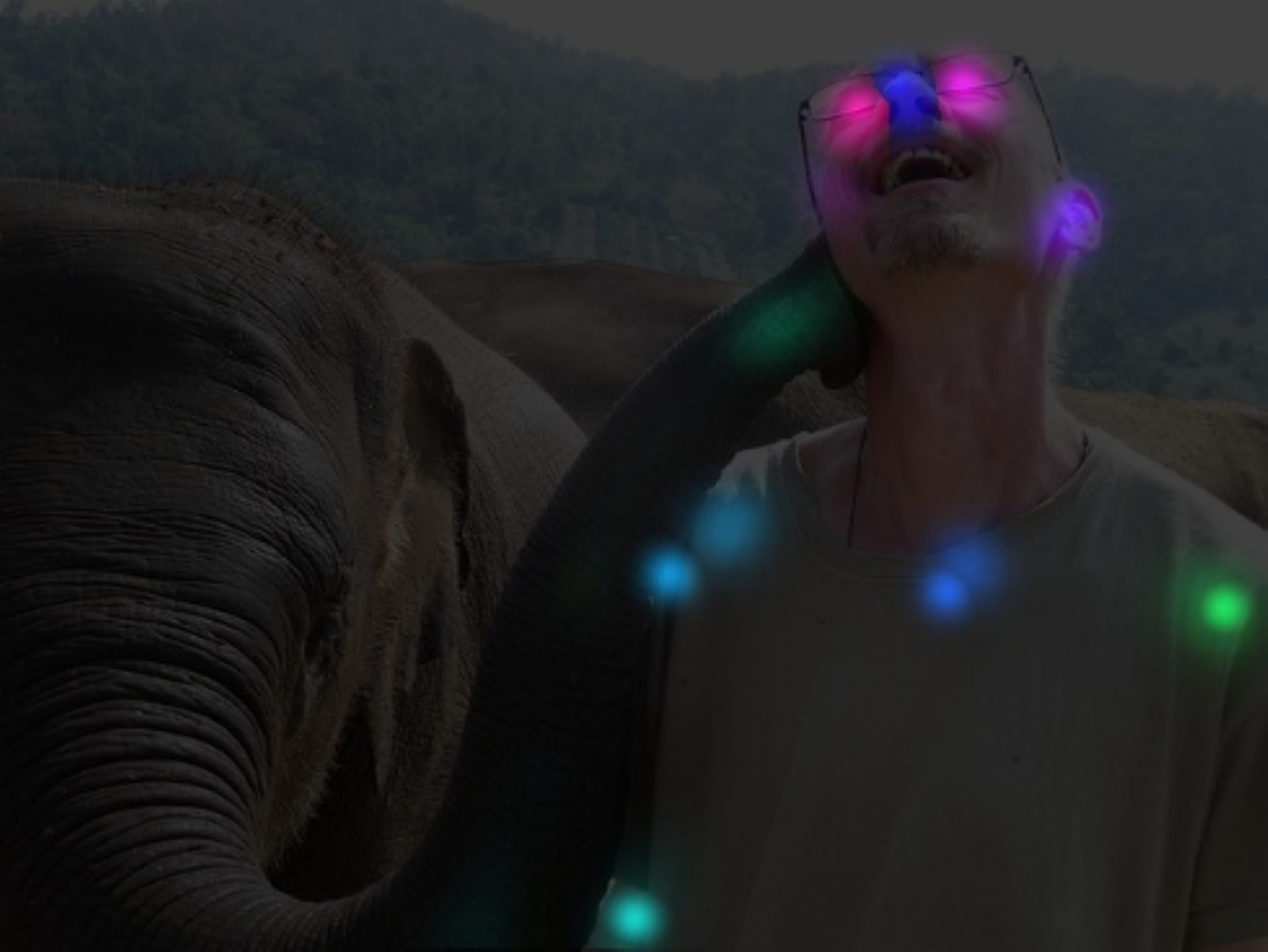} &
  \includegraphics[width=.133\hsize]{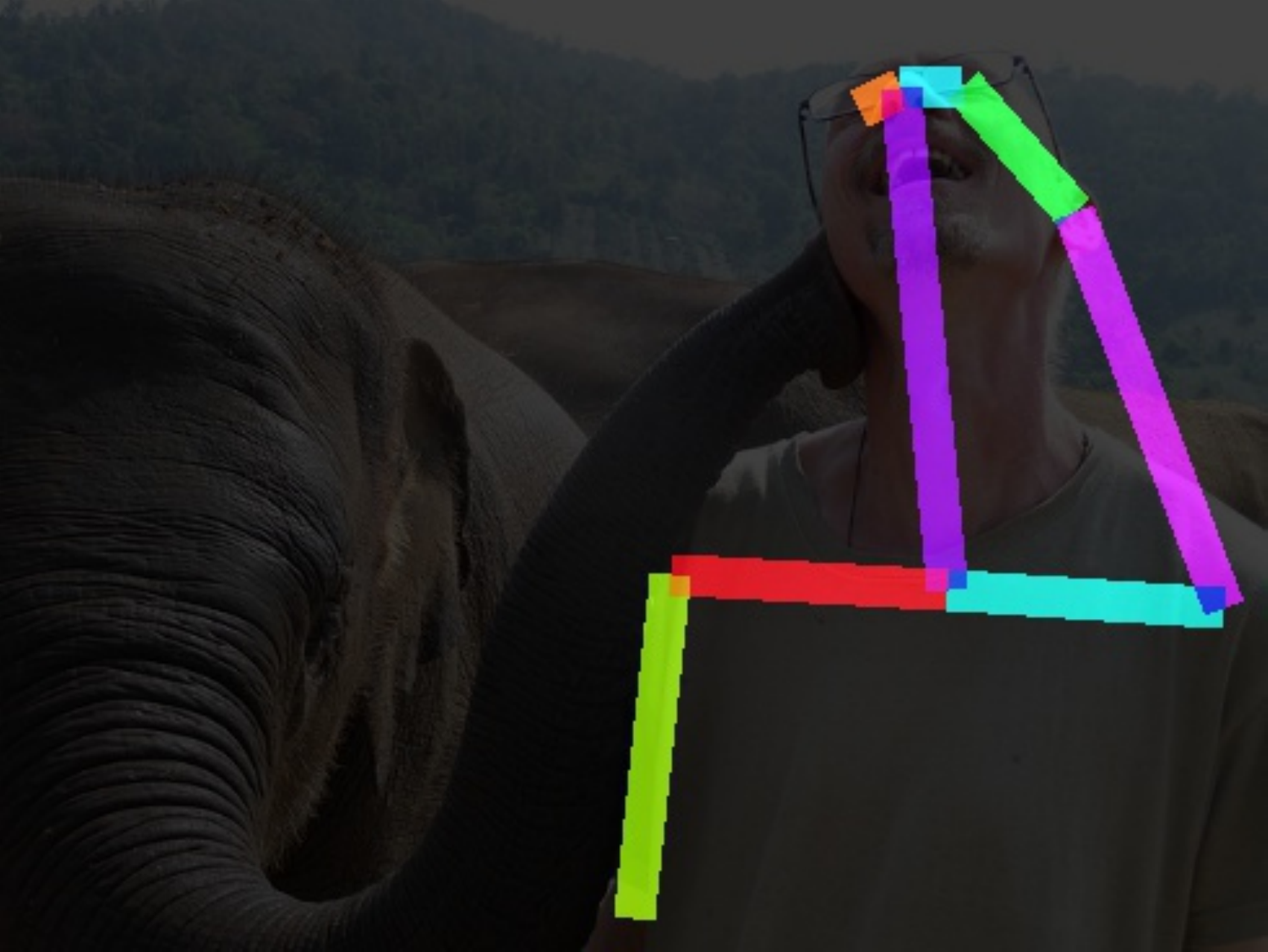} &
  \includegraphics[width=.133\hsize]{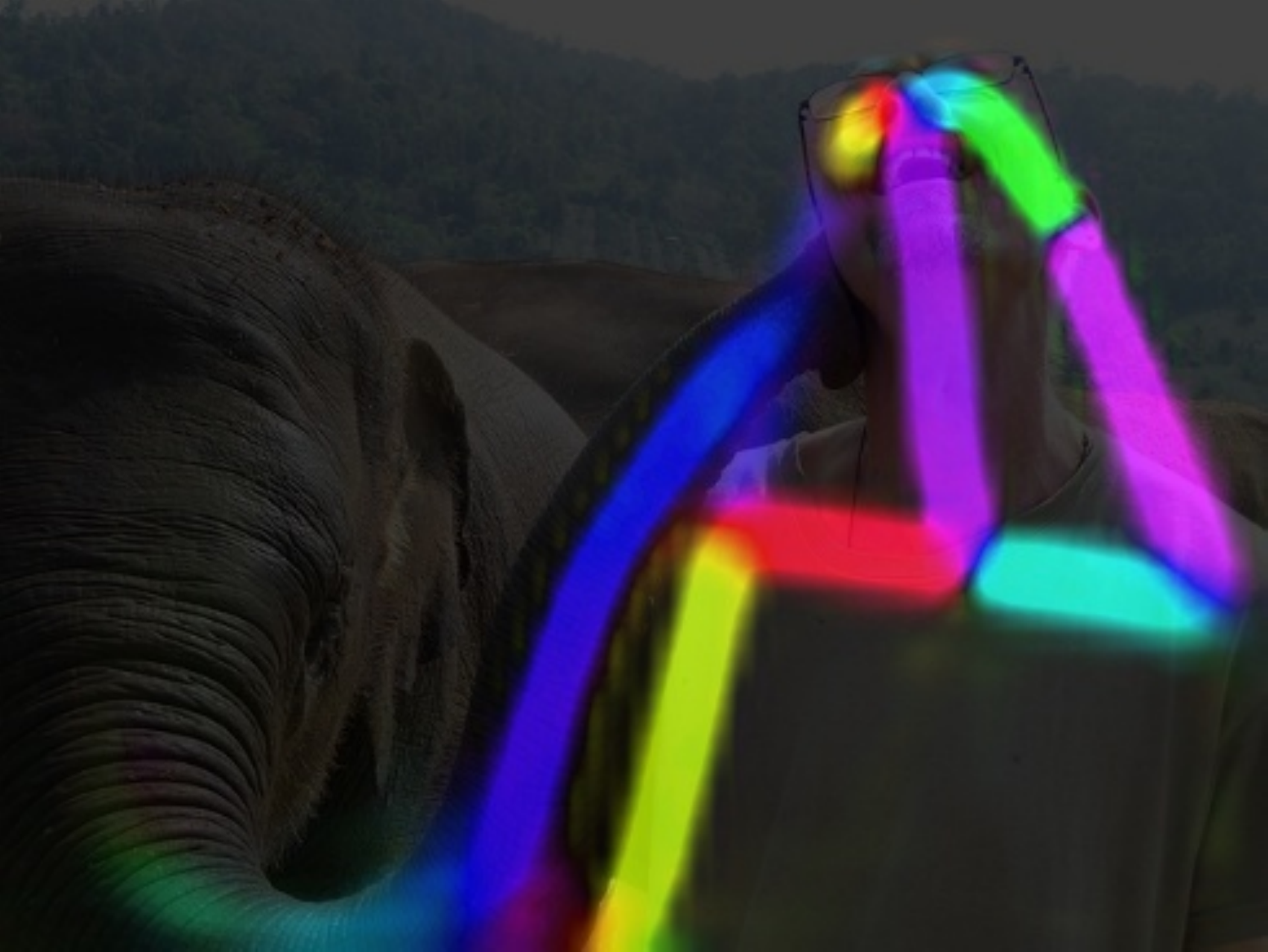} &
  \includegraphics[width=.133\hsize]{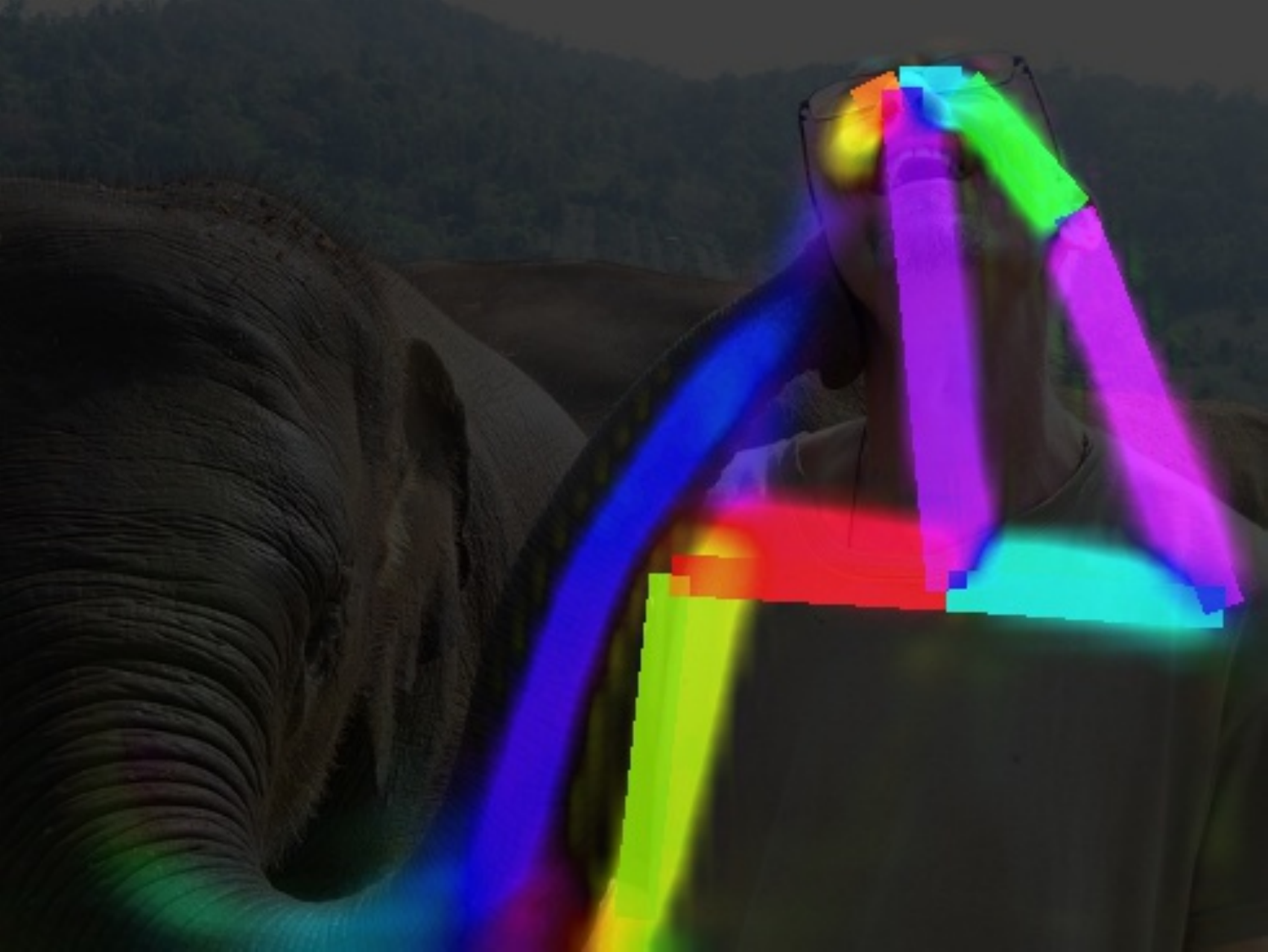} \\

  & \begin{tabular}{c}Keypoint\\annotations\end{tabular}
  & \begin{tabular}{c}Ground-truth\\confidence maps\end{tabular}
  & \begin{tabular}{c}Predicted\\confidence maps\end{tabular}
  & \begin{tabular}{c}Corrected\\confidence maps\end{tabular}
  & \begin{tabular}{c}Ground-truth\\PAFs\end{tabular}
  & \begin{tabular}{c}Predicted\\PAFs\end{tabular}
  & \begin{tabular}{c}Corrected\\PAFs\end{tabular}
\end{tabular}}
\end{center}
\caption{Some representative results of proposed label correction. (a)$\sim$(c) show results for images with keypoint protrusions, (d) and (e) with keypoint occlusion, and (f) with missed annotations, respectively. (g) and (h) show failure cases.}
\label{fig:qualitative_results}
\end{figure*}

\subsection{Ablation Study}
In order to confirm the performance gain was obtained by correcting both of confidence maps and PAFs, we train models correcting one of either of the labels. Table \ref{tab:ablation_study} shows the evaluation results of the models trained using unchanged labels, labels including only corrected confidence maps or PAFs, and labels that have both been corrected. The table shows, when we correct only confidence map or PAFs, the performance of the model is not good as the case where both labels are corrected. It is observed that both of corrected labels do contribute to improving the performance of the model.

% -------------------------------------------------------------

\subsection{Qualitative Results}
Figure \ref{fig:qualitative_results} shows some qualitative results of our proposed label correction. It can be seen that labels of data with keypoint protrusions, occlusion, and missed annotations are appropriately corrected.
Common failure cases include false estimations of the teacher model to objects similar to people and misalignment of the labels and the teacher output.

% -------------------------------------------------------------

\section{Conclusion}
In this paper, we first point out the existence of some patterns of inappropriate labels in a pose estimation method, especially the ones used for part confidence maps and part affinity fields.
We indicate that such labels result from the label map generation method based on keypoint annotations
and missed annotations in datasets. Then, in order to improve such labels, we propose a novel method using a teacher model trained with data including such inappropriate labels for correcting such labels.
In the experiments on the COCO dataset, we show the training using corrected labels by our proposed method improves the performance of the trained model. The model also converges faster as can be seen in training with knowledge distillation.
Label correction using a teacher model could also be applied to other pose estimation algorithms and other vision tasks such as object detection and semantic segmentation.

% -------------------------------------------------------------

{\small
\bibliographystyle{style/ieee}
\bibliography{ref}
}

\end{document}